%% file: main.tex
\pdftrailerid{}
\documentclass{article}
\usepackage{microtype}
\usepackage{graphicx}
\usepackage{subcaption}
\usepackage{booktabs}
\usepackage{array}
\usepackage{hyperref}

\usepackage[preprint]{icml2027}
\hypersetup{pdftitle={Fast to Start, Slow to Finish: Revisiting Position-Independent Caching for Reasoning Models},
  pdfauthor={Shuxiao Xie, Shuyang Xie, Dezhi Ran, Wei Yang, Tao Xie}, pdfsubject={Preprint}}
\usepackage{xurl}
\usepackage{amsmath}
\usepackage{amssymb}

\input{generated/evidence-macros.tex}

\makeatletter
\def\pc@sign#1#2\pc@end{#1}
\newcommand{\pc@box}[1]{\makebox[0.75em][l]{$^{#1}$}}
\newcommand{\pcblank}{\pc@box{}}
\newcommand{\pcmark}[2]{\edef\pc@lo{\expandafter\pc@sign#1\pc@end}\edef\pc@hi{\expandafter\pc@sign#2\pc@end}%
  \if+\pc@lo\pc@box{\uparrow}\else\if-\pc@hi\pc@box{\downarrow}\else\pcblank\fi\fi}
\newcommand{\pcmarkr}[2]{\edef\pc@lo{\expandafter\pc@sign#1\pc@end}\edef\pc@hi{\expandafter\pc@sign#2\pc@end}%
  \if+\pc@lo\pc@box{\downarrow}\else\if-\pc@hi\pc@box{\uparrow}\else\pcblank\fi\fi}
\makeatother

\makeatletter
\newcommand{\printAffiliationsPreprint}{\global\icml@noticeprintedtrue%
  \stepcounter{@affiliationcounter}%
  {\let\thefootnote\relax\footnotetext{\hspace*{-\footnotesep}{\scriptsize%
      \forloop{@affilnum}{1}{\value{@affilnum} < \value{@affiliationcounter}}{
        \textsuperscript{\arabic{@affilnum}}\csname @affilname\the@affilnum\endcsname
      }.%
      { }Correspondence to: \icmlcorrespondingauthor@text. \textit{Preprint.}}%
    }
  }
}
\makeatother

\icmltitlerunning{Revisiting Position-Independent Caching for Reasoning Models}

\begin{document}
\twocolumn[
  \icmltitle{Fast to Start, Slow to Finish: Revisiting Position-Independent Caching for Reasoning Models}
  \begin{icmlauthorlist}
    \icmlauthor{Shuxiao Xie}{tlaic,isac}
    \icmlauthor{Shuyang Xie}{hit}
    \icmlauthor{Dezhi Ran}{tlaic}
    \icmlauthor{Wei Yang}{isac}
    \icmlauthor{Tao Xie}{tlaic,isac,pku,sisoc}
  \end{icmlauthorlist}
  \icmlaffiliation{tlaic}{Beijing Tongming Lake Information Technology Application Innovation Center (TLAIC), China}
  \icmlaffiliation{isac}{Fudan University Institute of Systems for Advanced Computing, China}
  \icmlaffiliation{hit}{Harbin Institute of Technology, China}
  \icmlaffiliation{pku}{Key Lab of HCST (PKU), MOE; SCS, Peking University, Beijing, China}
  \icmlaffiliation{sisoc}{Shanghai Institute of Systems for Open Computing, China}
  \icmlcorrespondingauthor{Dezhi Ran}{dezhiran@pku.edu.cn}
  \icmlcorrespondingauthor{Tao Xie}{taoxie@pku.edu.cn}
  \vskip 0.3in
]
\printAffiliationsPreprint

\begin{abstract}
Prefix caching rarely hits when prompts share text chunks in varying orders. Position-independent caching (PIC) instead prefills each chunk separately offline, reuses these compiled KV caches at any position, and recomputes a small share of tokens online to restore cross-chunk attention. PIC cuts first-token latency on short answers, but with long reasoning chains, that latency no longer reflects the time saved. We therefore revisit eight PIC methods in a unified reimplementation on four reasoning models and three QA benchmarks. At 20\% recompute or less, the published methods lose accuracy against full prefill and generate more tokens in most of the 12 model--benchmark cells. In serving, two of them can even finish requests later than full prefill. To trace how a PIC cache damages reasoning, we replay full prefill's output over the cache. The damage begins with standalone compilation, whose displacement is largely shared across prompts. Compiling behind one fixed text therefore offsets much of it offline, leaving the rest to online recomputation. Since the recomputed tokens reveal their chunk's shift, propagating their mean change reduces the damage further. We assemble these three steps into a training-free recipe that outscores each published method at the same 20\% recompute in \NAggFOneTwoMTwoZeroVsQZTwoZeroAccGainCells{}--\NAggGOneTwoMTwoZeroVsCBAccGainCells{} of the 12 cells and generates fewer tokens in \NAggFOneTwoMTwoZeroVsQZTwoZeroGenShorterCells{}--\NAggDarDAPTwoZeroVsMTwoZeroGenPosCells{}. In serving on an H20 GPU, the recipe reaches \NAggCapPTwoZeroRatioMin{}--\NAggCapOneFourPTwoZeroRatioMax{}$\times$ full prefill's peak completion rate over the loads tested.
\end{abstract}

\input{sections/01-introduction}
\input{sections/02-background}
\input{sections/03-revisiting}
\input{sections/04-diagnosis}
\input{sections/05-recipe}
\input{sections/06-serving}
\input{sections/07-conclusion}

\bibliographystyle{icml2027}
\bibliography{main}

\input{sections/A-ports}

\input{sections/D-estimate}

\input{sections/E-fixedtext}
\input{sections/B-cells}

\input{sections/A-appendix}

\end{document}

%% file: generated/evidence-macros.tex
\newcommand{\NFcQEightrCBCon}{0.230}
\newcommand{\NFcQEightrCBGen}{761}
\newcommand{\NFcQEightrPKpTwoZeroCon}{0.427}
\newcommand{\NFcQEightrPKpTwoZeroGen}{1112}
\newcommand{\NFcQEightrMTwoZeroCon}{0.555}
\newcommand{\NFcQEightrMTwoZeroGen}{692}
\newcommand{\NFcQEightrMTwoZeroiCon}{0.578}
\newcommand{\NFcQEightrMTwoZeroiGen}{684}
\newcommand{\NFcQEightrFRCon}{0.320}
\newcommand{\NFcQEightrFRGen}{1719}
\newcommand{\NFcQEightrFCon}{0.630}
\newcommand{\NFcQEightrFGen}{522}
\newcommand{\NFcQEightrFNq}{100}
\newcommand{\NFcQEightrMTwoZeroVsFConLo}{-0.130}
\newcommand{\NFcQEightrMTwoZeroVsFConHi}{-0.028}
\newcommand{\NFcQEightrMTwoZeroVsFGenLo}{+77}
\newcommand{\NFcQEightrMTwoZeroVsFGenHi}{+259}
\newcommand{\NFcQEightrMTwoZeroiVsFConLo}{-0.100}
\newcommand{\NFcQEightrMTwoZeroiVsFConHi}{-0.010}
\newcommand{\NFcQEightrMTwoZeroiVsFGenLo}{+83}
\newcommand{\NFcQEightrMTwoZeroiVsFGenHi}{+244}
\newcommand{\NFcQEightrMTwoZeroVsPKpTwoZeroConLo}{+0.055}
\newcommand{\NFcQEightrMTwoZeroVsPKpTwoZeroConHi}{+0.200}
\newcommand{\NFcQEightrMTwoZeroVsPKpTwoZeroGenLo}{-581}
\newcommand{\NFcQEightrMTwoZeroVsPKpTwoZeroGenHi}{-260}
\newcommand{\NFcQEightrMTwoZeroVsCBConLo}{+0.240}
\newcommand{\NFcQEightrMTwoZeroVsCBConHi}{+0.410}
\newcommand{\NFcQEightrMTwoZeroVsCBGenLo}{-310}
\newcommand{\NFcQEightrMTwoZeroVsCBGenHi}{+152}
\newcommand{\NFcQEightrMTwoZeroVsFRConLo}{+0.158}
\newcommand{\NFcQEightrMTwoZeroVsFRConHi}{+0.318}
\newcommand{\NFcQEightrMTwoZeroVsFRGenLo}{-1264}
\newcommand{\NFcQEightrMTwoZeroVsFRGenHi}{-808}
\newcommand{\NFcQOneFourrCBCon}{0.280}
\newcommand{\NFcQOneFourrCBGen}{1138}
\newcommand{\NFcQOneFourrPKpTwoZeroCon}{0.530}
\newcommand{\NFcQOneFourrPKpTwoZeroGen}{992}
\newcommand{\NFcQOneFourrMTwoZeroCon}{0.620}
\newcommand{\NFcQOneFourrMTwoZeroGen}{455}
\newcommand{\NFcQOneFourrMTwoZeroiCon}{0.635}
\newcommand{\NFcQOneFourrMTwoZeroiGen}{462}
\newcommand{\NFcQOneFourrFRCon}{0.440}
\newcommand{\NFcQOneFourrFRGen}{1307}
\newcommand{\NFcQOneFourrFCon}{0.645}
\newcommand{\NFcQOneFourrFGen}{436}
\newcommand{\NFcQOneFourrMTwoZeroVsFConLo}{-0.072}
\newcommand{\NFcQOneFourrMTwoZeroVsFConHi}{+0.018}
\newcommand{\NFcQOneFourrMTwoZeroVsFGenLo}{-21}
\newcommand{\NFcQOneFourrMTwoZeroVsFGenHi}{+60}
\newcommand{\NFcQOneFourrMTwoZeroiVsFConLo}{-0.052}
\newcommand{\NFcQOneFourrMTwoZeroiVsFConHi}{+0.030}
\newcommand{\NFcQOneFourrMTwoZeroiVsFGenLo}{-37}
\newcommand{\NFcQOneFourrMTwoZeroiVsFGenHi}{+89}
\newcommand{\NFcQOneFourrMTwoZeroVsPKpTwoZeroConLo}{+0.010}
\newcommand{\NFcQOneFourrMTwoZeroVsPKpTwoZeroConHi}{+0.172}
\newcommand{\NFcQOneFourrMTwoZeroVsPKpTwoZeroGenLo}{-704}
\newcommand{\NFcQOneFourrMTwoZeroVsPKpTwoZeroGenHi}{-378}
\newcommand{\NFcQOneFourrMTwoZeroVsCBConLo}{+0.245}
\newcommand{\NFcQOneFourrMTwoZeroVsCBConHi}{+0.435}
\newcommand{\NFcQOneFourrMTwoZeroVsCBGenLo}{-878}
\newcommand{\NFcQOneFourrMTwoZeroVsCBGenHi}{-491}
\newcommand{\NFcQOneFourrMTwoZeroVsFRConLo}{+0.102}
\newcommand{\NFcQOneFourrMTwoZeroVsFRConHi}{+0.258}
\newcommand{\NFcQOneFourrMTwoZeroVsFRGenLo}{-1064}
\newcommand{\NFcQOneFourrMTwoZeroVsFRGenHi}{-642}
\newcommand{\NFcROnerMTwoZeroCon}{0.535}
\newcommand{\NFcROnerMTwoZeroGen}{803}
\newcommand{\NFcROnerMTwoZeroiCon}{0.555}
\newcommand{\NFcROnerMTwoZeroiGen}{632}
\newcommand{\NFcROnerFRCon}{0.333}
\newcommand{\NFcROnerFRGen}{1047}
\newcommand{\NFcROnerCBCon}{0.260}
\newcommand{\NFcROnerCBGen}{1678}
\newcommand{\NFcROnerPKpTwoZeroCon}{0.490}
\newcommand{\NFcROnerPKpTwoZeroGen}{963}
\newcommand{\NFcROnerFCon}{0.580}
\newcommand{\NFcROnerFGen}{496}
\newcommand{\NFcROnerMTwoZeroVsFConLo}{-0.092}
\newcommand{\NFcROnerMTwoZeroVsFConHi}{+0.000}
\newcommand{\NFcROnerMTwoZeroVsFGenLo}{+158}
\newcommand{\NFcROnerMTwoZeroVsFGenHi}{+460}
\newcommand{\NFcROnerMTwoZeroiVsFConLo}{-0.068}
\newcommand{\NFcROnerMTwoZeroiVsFConHi}{+0.018}
\newcommand{\NFcROnerMTwoZeroiVsFGenLo}{+46}
\newcommand{\NFcROnerMTwoZeroiVsFGenHi}{+250}
\newcommand{\NFcROnerMTwoZeroVsPKpTwoZeroConLo}{-0.013}
\newcommand{\NFcROnerMTwoZeroVsPKpTwoZeroConHi}{+0.102}
\newcommand{\NFcROnerMTwoZeroVsPKpTwoZeroGenLo}{-314}
\newcommand{\NFcROnerMTwoZeroVsPKpTwoZeroGenHi}{+2}
\newcommand{\NFcROnerMTwoZeroVsCBConLo}{+0.195}
\newcommand{\NFcROnerMTwoZeroVsCBConHi}{+0.355}
\newcommand{\NFcROnerMTwoZeroVsCBGenLo}{-1177}
\newcommand{\NFcROnerMTwoZeroVsCBGenHi}{-573}
\newcommand{\NFcROnerMTwoZeroVsFRConLo}{+0.133}
\newcommand{\NFcROnerMTwoZeroVsFRConHi}{+0.278}
\newcommand{\NFcROnerMTwoZeroVsFRGenLo}{-480}
\newcommand{\NFcROnerMTwoZeroVsFRGenHi}{-10}
\newcommand{\NFcQEightwCBFOne}{0.692}
\newcommand{\NFcQEightwCBGen}{721}
\newcommand{\NFcQEightwPKpTwoZeroFOne}{0.714}
\newcommand{\NFcQEightwPKpTwoZeroGen}{607}
\newcommand{\NFcQEightwMTwoZeroFOne}{0.735}
\newcommand{\NFcQEightwMTwoZeroGen}{538}
\newcommand{\NFcQEightwMTwoZeroiFOne}{0.731}
\newcommand{\NFcQEightwMTwoZeroiGen}{540}
\newcommand{\NFcQEightwFRFOne}{0.690}
\newcommand{\NFcQEightwFRGen}{769}
\newcommand{\NFcQEightwFFOne}{0.753}
\newcommand{\NFcQEightwFGen}{450}
\newcommand{\NFcQEightwFNq}{294}
\newcommand{\NFcQEightwMTwoZeroVsFFOneLo}{-0.036}
\newcommand{\NFcQEightwMTwoZeroVsFFOneHi}{-0.003}
\newcommand{\NFcQEightwMTwoZeroVsFGenLo}{+46}
\newcommand{\NFcQEightwMTwoZeroVsFGenHi}{+133}
\newcommand{\NFcQEightwMTwoZeroiVsFFOneLo}{-0.042}
\newcommand{\NFcQEightwMTwoZeroiVsFFOneHi}{-0.006}
\newcommand{\NFcQEightwMTwoZeroiVsFGenLo}{+47}
\newcommand{\NFcQEightwMTwoZeroiVsFGenHi}{+135}
\newcommand{\NFcQEightwMTwoZeroVsPKpTwoZeroFOneLo}{+0.000}
\newcommand{\NFcQEightwMTwoZeroVsPKpTwoZeroFOneHi}{+0.042}
\newcommand{\NFcQEightwMTwoZeroVsPKpTwoZeroGenLo}{-124}
\newcommand{\NFcQEightwMTwoZeroVsPKpTwoZeroGenHi}{-17}
\newcommand{\NFcQEightwMTwoZeroVsCBFOneLo}{+0.017}
\newcommand{\NFcQEightwMTwoZeroVsCBFOneHi}{+0.070}
\newcommand{\NFcQEightwMTwoZeroVsCBGenLo}{-249}
\newcommand{\NFcQEightwMTwoZeroVsCBGenHi}{-116}
\newcommand{\NFcQEightwMTwoZeroVsFRFOneLo}{+0.021}
\newcommand{\NFcQEightwMTwoZeroVsFRFOneHi}{+0.069}
\newcommand{\NFcQEightwMTwoZeroVsFRGenLo}{-323}
\newcommand{\NFcQEightwMTwoZeroVsFRGenHi}{-145}
\newcommand{\NFcQOneFourwCBFOne}{0.676}
\newcommand{\NFcQOneFourwCBGen}{507}
\newcommand{\NFcQOneFourwPKpTwoZeroFOne}{0.720}
\newcommand{\NFcQOneFourwPKpTwoZeroGen}{460}
\newcommand{\NFcQOneFourwMTwoZeroFOne}{0.728}
\newcommand{\NFcQOneFourwMTwoZeroGen}{395}
\newcommand{\NFcQOneFourwMTwoZeroiFOne}{0.734}
\newcommand{\NFcQOneFourwMTwoZeroiGen}{376}
\newcommand{\NFcQOneFourwFRFOne}{0.725}
\newcommand{\NFcQOneFourwFRGen}{489}
\newcommand{\NFcQOneFourwFFOne}{0.740}
\newcommand{\NFcQOneFourwFGen}{360}
\newcommand{\NFcQOneFourwMTwoZeroVsFFOneLo}{-0.023}
\newcommand{\NFcQOneFourwMTwoZeroVsFFOneHi}{-0.001}
\newcommand{\NFcQOneFourwMTwoZeroVsFGenLo}{+11}
\newcommand{\NFcQOneFourwMTwoZeroVsFGenHi}{+62}
\newcommand{\NFcQOneFourwMTwoZeroiVsFFOneLo}{-0.017}
\newcommand{\NFcQOneFourwMTwoZeroiVsFFOneHi}{+0.005}
\newcommand{\NFcQOneFourwMTwoZeroiVsFGenLo}{-4}
\newcommand{\NFcQOneFourwMTwoZeroiVsFGenHi}{+34}
\newcommand{\NFcQOneFourwMTwoZeroVsPKpTwoZeroFOneLo}{-0.008}
\newcommand{\NFcQOneFourwMTwoZeroVsPKpTwoZeroFOneHi}{+0.025}
\newcommand{\NFcQOneFourwMTwoZeroVsPKpTwoZeroGenLo}{-106}
\newcommand{\NFcQOneFourwMTwoZeroVsPKpTwoZeroGenHi}{-28}
\newcommand{\NFcQOneFourwMTwoZeroVsCBFOneLo}{+0.026}
\newcommand{\NFcQOneFourwMTwoZeroVsCBFOneHi}{+0.082}
\newcommand{\NFcQOneFourwMTwoZeroVsCBGenLo}{-176}
\newcommand{\NFcQOneFourwMTwoZeroVsCBGenHi}{-53}
\newcommand{\NFcQOneFourwMTwoZeroVsFRFOneLo}{-0.016}
\newcommand{\NFcQOneFourwMTwoZeroVsFRFOneHi}{+0.024}
\newcommand{\NFcQOneFourwMTwoZeroVsFRGenLo}{-156}
\newcommand{\NFcQOneFourwMTwoZeroVsFRGenHi}{-39}
\newcommand{\NFcNwMTwoZeroBox}{0.599}
\newcommand{\NFcNwMTwoZeroGen}{976}
\newcommand{\NFcNwMTwoZeroiBox}{0.591}
\newcommand{\NFcNwMTwoZeroiGen}{961}
\newcommand{\NFcNwFRBox}{0.473}
\newcommand{\NFcNwFRGen}{1496}
\newcommand{\NFcNwCBBox}{0.456}
\newcommand{\NFcNwCBGen}{1529}
\newcommand{\NFcNwPKpTwoZeroBox}{0.577}
\newcommand{\NFcNwPKpTwoZeroGen}{1139}
\newcommand{\NFcNwFBox}{0.612}
\newcommand{\NFcNwFGen}{800}
\newcommand{\NFcNwMTwoZeroVsFBoxLo}{-0.039}
\newcommand{\NFcNwMTwoZeroVsFBoxHi}{+0.013}
\newcommand{\NFcNwMTwoZeroVsFGenLo}{+80}
\newcommand{\NFcNwMTwoZeroVsFGenHi}{+274}
\newcommand{\NFcNwMTwoZeroiVsFBoxLo}{-0.046}
\newcommand{\NFcNwMTwoZeroiVsFBoxHi}{+0.005}
\newcommand{\NFcNwMTwoZeroiVsFGenLo}{+66}
\newcommand{\NFcNwMTwoZeroiVsFGenHi}{+261}
\newcommand{\NFcNwMTwoZeroVsPKpTwoZeroBoxLo}{-0.003}
\newcommand{\NFcNwMTwoZeroVsPKpTwoZeroBoxHi}{+0.049}
\newcommand{\NFcNwMTwoZeroVsPKpTwoZeroGenLo}{-287}
\newcommand{\NFcNwMTwoZeroVsPKpTwoZeroGenHi}{-38}
\newcommand{\NFcNwMTwoZeroVsCBBoxLo}{+0.109}
\newcommand{\NFcNwMTwoZeroVsCBBoxHi}{+0.179}
\newcommand{\NFcNwMTwoZeroVsCBGenLo}{-701}
\newcommand{\NFcNwMTwoZeroVsCBGenHi}{-403}
\newcommand{\NFcNwMTwoZeroVsFRBoxLo}{+0.094}
\newcommand{\NFcNwMTwoZeroVsFRBoxHi}{+0.161}
\newcommand{\NFcNwMTwoZeroVsFRGenLo}{-659}
\newcommand{\NFcNwMTwoZeroVsFRGenHi}{-387}
\newcommand{\NFcQEighthCBFOne}{0.698}
\newcommand{\NFcQEighthCBGen}{850}
\newcommand{\NFcQEighthPKpTwoZeroFOne}{0.707}
\newcommand{\NFcQEighthPKpTwoZeroGen}{677}
\newcommand{\NFcQEighthMTwoZeroFOne}{0.726}
\newcommand{\NFcQEighthMTwoZeroGen}{613}
\newcommand{\NFcQEighthMTwoZeroiFOne}{0.727}
\newcommand{\NFcQEighthMTwoZeroiGen}{619}
\newcommand{\NFcQEighthFRFOne}{0.691}
\newcommand{\NFcQEighthFRGen}{738}
\newcommand{\NFcQEighthFFOne}{0.722}
\newcommand{\NFcQEighthFGen}{590}
\newcommand{\NFcQEighthFNq}{288}
\newcommand{\NFcQEighthMTwoZeroVsFFOneLo}{-0.012}
\newcommand{\NFcQEighthMTwoZeroVsFFOneHi}{+0.019}
\newcommand{\NFcQEighthMTwoZeroVsFGenLo}{-14}
\newcommand{\NFcQEighthMTwoZeroVsFGenHi}{+63}
\newcommand{\NFcQEighthMTwoZeroiVsFFOneLo}{-0.010}
\newcommand{\NFcQEighthMTwoZeroiVsFFOneHi}{+0.020}
\newcommand{\NFcQEighthMTwoZeroiVsFGenLo}{-16}
\newcommand{\NFcQEighthMTwoZeroiVsFGenHi}{+76}
\newcommand{\NFcQEighthMTwoZeroVsPKpTwoZeroFOneLo}{-0.001}
\newcommand{\NFcQEighthMTwoZeroVsPKpTwoZeroFOneHi}{+0.039}
\newcommand{\NFcQEighthMTwoZeroVsPKpTwoZeroGenLo}{-115}
\newcommand{\NFcQEighthMTwoZeroVsPKpTwoZeroGenHi}{-15}
\newcommand{\NFcQEighthMTwoZeroVsCBFOneLo}{+0.004}
\newcommand{\NFcQEighthMTwoZeroVsCBFOneHi}{+0.053}
\newcommand{\NFcQEighthMTwoZeroVsCBGenLo}{-315}
\newcommand{\NFcQEighthMTwoZeroVsCBGenHi}{-169}
\newcommand{\NFcQEighthMTwoZeroVsFRFOneLo}{+0.013}
\newcommand{\NFcQEighthMTwoZeroVsFRFOneHi}{+0.060}
\newcommand{\NFcQEighthMTwoZeroVsFRGenLo}{-185}
\newcommand{\NFcQEighthMTwoZeroVsFRGenHi}{-72}
\newcommand{\NFcQOneFourhCBFOne}{0.698}
\newcommand{\NFcQOneFourhCBGen}{604}
\newcommand{\NFcQOneFourhPKpTwoZeroFOne}{0.722}
\newcommand{\NFcQOneFourhPKpTwoZeroGen}{540}
\newcommand{\NFcQOneFourhMTwoZeroFOne}{0.734}
\newcommand{\NFcQOneFourhMTwoZeroGen}{486}
\newcommand{\NFcQOneFourhMTwoZeroiFOne}{0.729}
\newcommand{\NFcQOneFourhMTwoZeroiGen}{501}
\newcommand{\NFcQOneFourhFRFOne}{0.707}
\newcommand{\NFcQOneFourhFRGen}{556}
\newcommand{\NFcQOneFourhFFOne}{0.738}
\newcommand{\NFcQOneFourhFGen}{482}
\newcommand{\NFcQOneFourhMTwoZeroVsFFOneLo}{-0.019}
\newcommand{\NFcQOneFourhMTwoZeroVsFFOneHi}{+0.008}
\newcommand{\NFcQOneFourhMTwoZeroVsFGenLo}{-17}
\newcommand{\NFcQOneFourhMTwoZeroVsFGenHi}{+27}
\newcommand{\NFcQOneFourhMTwoZeroiVsFFOneLo}{-0.023}
\newcommand{\NFcQOneFourhMTwoZeroiVsFFOneHi}{+0.004}
\newcommand{\NFcQOneFourhMTwoZeroiVsFGenLo}{+0}
\newcommand{\NFcQOneFourhMTwoZeroiVsFGenHi}{+40}
\newcommand{\NFcQOneFourhMTwoZeroVsPKpTwoZeroFOneLo}{-0.003}
\newcommand{\NFcQOneFourhMTwoZeroVsPKpTwoZeroFOneHi}{+0.028}
\newcommand{\NFcQOneFourhMTwoZeroVsPKpTwoZeroGenLo}{-88}
\newcommand{\NFcQOneFourhMTwoZeroVsPKpTwoZeroGenHi}{-25}
\newcommand{\NFcQOneFourhMTwoZeroVsCBFOneLo}{+0.016}
\newcommand{\NFcQOneFourhMTwoZeroVsCBFOneHi}{+0.058}
\newcommand{\NFcQOneFourhMTwoZeroVsCBGenLo}{-172}
\newcommand{\NFcQOneFourhMTwoZeroVsCBGenHi}{-69}
\newcommand{\NFcQOneFourhMTwoZeroVsFRFOneLo}{+0.010}
\newcommand{\NFcQOneFourhMTwoZeroVsFRFOneHi}{+0.045}
\newcommand{\NFcQOneFourhMTwoZeroVsFRGenLo}{-105}
\newcommand{\NFcQOneFourhMTwoZeroVsFRGenHi}{-36}
\newcommand{\NFcNhMTwoZeroBox}{0.563}
\newcommand{\NFcNhMTwoZeroGen}{977}
\newcommand{\NFcNhMTwoZeroiBox}{0.559}
\newcommand{\NFcNhMTwoZeroiGen}{951}
\newcommand{\NFcNhFRBox}{0.463}
\newcommand{\NFcNhFRGen}{1238}
\newcommand{\NFcNhCBBox}{0.382}
\newcommand{\NFcNhCBGen}{1466}
\newcommand{\NFcNhPKpTwoZeroBox}{0.497}
\newcommand{\NFcNhPKpTwoZeroGen}{1142}
\newcommand{\NFcNhFBox}{0.583}
\newcommand{\NFcNhFGen}{806}
\newcommand{\NFcNhMTwoZeroVsFBoxLo}{-0.042}
\newcommand{\NFcNhMTwoZeroVsFBoxHi}{+0.002}
\newcommand{\NFcNhMTwoZeroVsFGenLo}{+93}
\newcommand{\NFcNhMTwoZeroVsFGenHi}{+252}
\newcommand{\NFcNhMTwoZeroiVsFBoxLo}{-0.045}
\newcommand{\NFcNhMTwoZeroiVsFBoxHi}{-0.002}
\newcommand{\NFcNhMTwoZeroiVsFGenLo}{+65}
\newcommand{\NFcNhMTwoZeroiVsFGenHi}{+230}
\newcommand{\NFcNhMTwoZeroVsPKpTwoZeroBoxLo}{+0.037}
\newcommand{\NFcNhMTwoZeroVsPKpTwoZeroBoxHi}{+0.093}
\newcommand{\NFcNhMTwoZeroVsPKpTwoZeroGenLo}{-264}
\newcommand{\NFcNhMTwoZeroVsPKpTwoZeroGenHi}{-69}
\newcommand{\NFcNhMTwoZeroVsCBBoxLo}{+0.147}
\newcommand{\NFcNhMTwoZeroVsCBBoxHi}{+0.216}
\newcommand{\NFcNhMTwoZeroVsCBGenLo}{-620}
\newcommand{\NFcNhMTwoZeroVsCBGenHi}{-361}
\newcommand{\NFcNhMTwoZeroVsFRBoxLo}{+0.070}
\newcommand{\NFcNhMTwoZeroVsFRBoxHi}{+0.131}
\newcommand{\NFcNhMTwoZeroVsFRGenLo}{-375}
\newcommand{\NFcNhMTwoZeroVsFRGenHi}{-152}
\newcommand{\NPubQEightrAKCon}{0.265}
\newcommand{\NPubQEightrAKGen}{1404}
\newcommand{\NPubQEightrPKCon}{0.370}
\newcommand{\NPubQEightrPKGen}{1342}
\newcommand{\NPubQEightrMTwoZeroVsMnpGenPt}{-107}
\newcommand{\NPubQEightrMTwoZeroVsPKConLo}{+0.098}
\newcommand{\NPubQEightrMTwoZeroVsPKConHi}{+0.270}
\newcommand{\NPubQEightrMTwoZeroVsPKGenLo}{-882}
\newcommand{\NPubQEightrMTwoZeroVsPKGenHi}{-437}
\newcommand{\NPubQEightrMTwoZeroVsAKConLo}{+0.212}
\newcommand{\NPubQEightrMTwoZeroVsAKConHi}{+0.370}
\newcommand{\NPubQEightrMTwoZeroVsAKGenLo}{-863}
\newcommand{\NPubQEightrMTwoZeroVsAKGenHi}{-570}
\newcommand{\NPubNwAKBox}{0.483}
\newcommand{\NPubNwAKGen}{1653}
\newcommand{\NPubNwPKBox}{0.498}
\newcommand{\NPubNwPKGen}{1319}
\newcommand{\NPubNwMTwoZeroVsMnpBoxPt}{+0.035}
\newcommand{\NPubNwMTwoZeroVsPKBoxLo}{+0.072}
\newcommand{\NPubNwMTwoZeroVsPKBoxHi}{+0.131}
\newcommand{\NPubNwMTwoZeroVsPKGenLo}{-470}
\newcommand{\NPubNwMTwoZeroVsPKGenHi}{-208}
\newcommand{\NPubNwMTwoZeroVsAKBoxLo}{+0.083}
\newcommand{\NPubNwMTwoZeroVsAKBoxHi}{+0.151}
\newcommand{\NPubNwMTwoZeroVsAKGenLo}{-810}
\newcommand{\NPubNwMTwoZeroVsAKGenHi}{-545}
\newcommand{\NQzQEightrQZTwoZeroCon}{0.330}
\newcommand{\NQzQEightrQZTwoZeroGen}{2436}
\newcommand{\NQzQEightrQZFiveZeroCon}{0.443}
\newcommand{\NQzQEightrQZFiveZeroGen}{1201}
\newcommand{\NQzQEightrMTwoZeroVsQZTwoZeroConLo}{+0.130}
\newcommand{\NQzQEightrMTwoZeroVsQZTwoZeroConHi}{+0.318}
\newcommand{\NQzQEightrMTwoZeroVsQZTwoZeroGenLo}{-2095}
\newcommand{\NQzQEightrMTwoZeroVsQZTwoZeroGenHi}{-1395}
\newcommand{\NQzQEightrMTwoZeroVsQZFiveZeroConLo}{+0.045}
\newcommand{\NQzQEightrMTwoZeroVsQZFiveZeroConHi}{+0.180}
\newcommand{\NQzQEightrMTwoZeroVsQZFiveZeroGenLo}{-734}
\newcommand{\NQzQEightrMTwoZeroVsQZFiveZeroGenHi}{-316}
\newcommand{\NQzNwQZTwoZeroBox}{0.503}
\newcommand{\NQzNwQZTwoZeroGen}{1271}
\newcommand{\NQzNwQZFiveZeroBox}{0.589}
\newcommand{\NQzNwQZFiveZeroGen}{870}
\newcommand{\NQzNwMTwoZeroVsQZTwoZeroBoxLo}{+0.066}
\newcommand{\NQzNwMTwoZeroVsQZTwoZeroBoxHi}{+0.127}
\newcommand{\NQzNwMTwoZeroVsQZTwoZeroGenLo}{-424}
\newcommand{\NQzNwMTwoZeroVsQZTwoZeroGenHi}{-170}
\newcommand{\NQzNwMTwoZeroVsQZFiveZeroBoxLo}{-0.018}
\newcommand{\NQzNwMTwoZeroVsQZFiveZeroBoxHi}{+0.038}
\newcommand{\NQzNwMTwoZeroVsQZFiveZeroGenLo}{+19}
\newcommand{\NQzNwMTwoZeroVsQZFiveZeroGenHi}{+206}
\newcommand{\NEpQEightrLTwoCon}{0.037}
\newcommand{\NEpQEightrLTwoGen}{5786}
\newcommand{\NEpQEightrLThreeTwoCon}{0.198}
\newcommand{\NEpQEightrLThreeTwoGen}{1572}
\newcommand{\NEpQEightrMTwoZeroVsLTwoConLo}{+0.432}
\newcommand{\NEpQEightrMTwoZeroVsLTwoConHi}{+0.600}
\newcommand{\NEpQEightrMTwoZeroVsLTwoGenLo}{-5265}
\newcommand{\NEpQEightrMTwoZeroVsLTwoGenHi}{-4914}
\newcommand{\NEpQEightrMTwoZeroVsLThreeTwoConLo}{+0.268}
\newcommand{\NEpQEightrMTwoZeroVsLThreeTwoConHi}{+0.445}
\newcommand{\NEpQEightrMTwoZeroVsLThreeTwoGenLo}{-1133}
\newcommand{\NEpQEightrMTwoZeroVsLThreeTwoGenHi}{-649}
\newcommand{\NEpNwLTwoBox}{0.264}
\newcommand{\NEpNwLTwoGen}{2439}
\newcommand{\NEpNwLThreeTwoBox}{0.300}
\newcommand{\NEpNwLThreeTwoGen}{2224}
\newcommand{\NEpNwMTwoZeroVsLTwoBoxLo}{+0.296}
\newcommand{\NEpNwMTwoZeroVsLTwoBoxHi}{+0.378}
\newcommand{\NEpNwMTwoZeroVsLTwoGenLo}{-1677}
\newcommand{\NEpNwMTwoZeroVsLTwoGenHi}{-1256}
\newcommand{\NEpNwMTwoZeroVsLThreeTwoBoxLo}{+0.261}
\newcommand{\NEpNwMTwoZeroVsLThreeTwoBoxHi}{+0.338}
\newcommand{\NEpNwMTwoZeroVsLThreeTwoGenLo}{-1429}
\newcommand{\NEpNwMTwoZeroVsLThreeTwoGenHi}{-1068}
\newcommand{\NPFourZeroQEightrMFourZeroCon}{0.603}
\newcommand{\NPFourZeroQEightrMFourZeroGen}{602}
\newcommand{\NPFourZeroQEightrMFourZeroiCon}{0.608}
\newcommand{\NPFourZeroQEightrMFourZeroiGen}{568}
\newcommand{\NPFourZeroQEightrMFourZeroVsFConLo}{-0.062}
\newcommand{\NPFourZeroQEightrMFourZeroVsFConHi}{+0.003}
\newcommand{\NPFourZeroQEightrMFourZeroVsFGenLo}{+9}
\newcommand{\NPFourZeroQEightrMFourZeroVsFGenHi}{+151}
\newcommand{\NPFourZeroQEightrMFourZeroiVsFConLo}{-0.055}
\newcommand{\NPFourZeroQEightrMFourZeroiVsFConHi}{+0.007}
\newcommand{\NPFourZeroQEightrMFourZeroiVsFGenLo}{-22}
\newcommand{\NPFourZeroQEightrMFourZeroiVsFGenHi}{+112}
\newcommand{\NPFourZeroQEightwMFourZeroFOne}{0.747}
\newcommand{\NPFourZeroQEightwMFourZeroGen}{488}
\newcommand{\NPFourZeroQEightwMFourZeroiFOne}{0.750}
\newcommand{\NPFourZeroQEightwMFourZeroiGen}{475}
\newcommand{\NPFourZeroQEightwMFourZeroVsFFOneLo}{-0.021}
\newcommand{\NPFourZeroQEightwMFourZeroVsFFOneHi}{+0.007}
\newcommand{\NPFourZeroQEightwMFourZeroVsFGenLo}{+8}
\newcommand{\NPFourZeroQEightwMFourZeroVsFGenHi}{+70}
\newcommand{\NPFourZeroQEightwMFourZeroiVsFFOneLo}{-0.019}
\newcommand{\NPFourZeroQEightwMFourZeroiVsFFOneHi}{+0.010}
\newcommand{\NPFourZeroQEightwMFourZeroiVsFGenLo}{-10}
\newcommand{\NPFourZeroQEightwMFourZeroiVsFGenHi}{+58}
\newcommand{\NPFourZeroQOneFourwMFourZeroFOne}{0.735}
\newcommand{\NPFourZeroQOneFourwMFourZeroGen}{356}
\newcommand{\NPFourZeroQOneFourwMFourZeroiFOne}{0.740}
\newcommand{\NPFourZeroQOneFourwMFourZeroiGen}{361}
\newcommand{\NPFourZeroQOneFourwMFourZeroVsFFOneLo}{-0.015}
\newcommand{\NPFourZeroQOneFourwMFourZeroVsFFOneHi}{+0.005}
\newcommand{\NPFourZeroQOneFourwMFourZeroVsFGenLo}{-19}
\newcommand{\NPFourZeroQOneFourwMFourZeroVsFGenHi}{+8}
\newcommand{\NPFourZeroQOneFourwMFourZeroiVsFFOneLo}{-0.009}
\newcommand{\NPFourZeroQOneFourwMFourZeroiVsFFOneHi}{+0.009}
\newcommand{\NPFourZeroQOneFourwMFourZeroiVsFGenLo}{-15}
\newcommand{\NPFourZeroQOneFourwMFourZeroiVsFGenHi}{+19}
\newcommand{\NPFourZeroNhMFourZeroBox}{0.575}
\newcommand{\NPFourZeroNhMFourZeroGen}{907}
\newcommand{\NPFourZeroNhMFourZeroiBox}{0.582}
\newcommand{\NPFourZeroNhMFourZeroiGen}{906}
\newcommand{\NPFourZeroNhMFourZeroVsFBoxLo}{-0.030}
\newcommand{\NPFourZeroNhMFourZeroVsFBoxHi}{+0.015}
\newcommand{\NPFourZeroNhMFourZeroVsFGenLo}{+36}
\newcommand{\NPFourZeroNhMFourZeroVsFGenHi}{+168}
\newcommand{\NPFourZeroNhMFourZeroiVsFBoxLo}{-0.022}
\newcommand{\NPFourZeroNhMFourZeroiVsFBoxHi}{+0.020}
\newcommand{\NPFourZeroNhMFourZeroiVsFGenLo}{+25}
\newcommand{\NPFourZeroNhMFourZeroiVsFGenHi}{+181}
\newcommand{\NPFourZeroROnerMFourZeroCon}{0.608}
\newcommand{\NPFourZeroROnerMFourZeroGen}{594}
\newcommand{\NPFourZeroROnerMFourZeroiCon}{0.603}
\newcommand{\NPFourZeroROnerMFourZeroiGen}{573}
\newcommand{\NPFourZeroROnerMFourZeroVsFConLo}{-0.015}
\newcommand{\NPFourZeroROnerMFourZeroVsFConHi}{+0.070}
\newcommand{\NPFourZeroROnerMFourZeroVsFGenLo}{+1}
\newcommand{\NPFourZeroROnerMFourZeroVsFGenHi}{+199}
\newcommand{\NPFourZeroROnerMFourZeroiVsFConLo}{-0.020}
\newcommand{\NPFourZeroROnerMFourZeroiVsFConHi}{+0.065}
\newcommand{\NPFourZeroROnerMFourZeroiVsFGenLo}{+9}
\newcommand{\NPFourZeroROnerMFourZeroiVsFGenHi}{+147}
\newcommand{\NPFourZeroNwMFourZeroBox}{0.613}
\newcommand{\NPFourZeroNwMFourZeroGen}{831}
\newcommand{\NPFourZeroNwMFourZeroVsFBoxLo}{-0.022}
\newcommand{\NPFourZeroNwMFourZeroVsFBoxHi}{+0.024}
\newcommand{\NPFourZeroNwMFourZeroVsFGenLo}{-55}
\newcommand{\NPFourZeroNwMFourZeroVsFGenHi}{+112}
\newcommand{\NSwQEightrMFiveVsPKpFiveConPt}{+0.237}
\newcommand{\NSwNwMFiveVsPKpFiveBoxPt}{+0.110}
\newcommand{\NGrROnewMTwoZeroFOne}{0.410}
\newcommand{\NGrROnewMTwoZeroGen}{433}
\newcommand{\NGrROnewMTwoZeroiFOne}{0.418}
\newcommand{\NGrROnewMTwoZeroiGen}{419}
\newcommand{\NGrROnewFRFOne}{0.331}
\newcommand{\NGrROnewFRGen}{527}
\newcommand{\NGrROnewCBFOne}{0.358}
\newcommand{\NGrROnewCBGen}{548}
\newcommand{\NGrROnewPKpTwoZeroFOne}{0.376}
\newcommand{\NGrROnewPKpTwoZeroGen}{417}
\newcommand{\NGrROnewFFOne}{0.399}
\newcommand{\NGrROnewFGen}{368}
\newcommand{\NGrROnewMTwoZeroVsFFOneLo}{-0.011}
\newcommand{\NGrROnewMTwoZeroVsFFOneHi}{+0.033}
\newcommand{\NGrROnewMTwoZeroVsFGenLo}{+12}
\newcommand{\NGrROnewMTwoZeroVsFGenHi}{+122}
\newcommand{\NGrROnewMTwoZeroiVsFFOneLo}{-0.002}
\newcommand{\NGrROnewMTwoZeroiVsFFOneHi}{+0.040}
\newcommand{\NGrROnewMTwoZeroiVsFGenLo}{-0}
\newcommand{\NGrROnewMTwoZeroiVsFGenHi}{+111}
\newcommand{\NGrROnewMTwoZeroVsPKpTwoZeroFOneLo}{+0.013}
\newcommand{\NGrROnewMTwoZeroVsPKpTwoZeroFOneHi}{+0.055}
\newcommand{\NGrROnewMTwoZeroVsPKpTwoZeroGenLo}{-46}
\newcommand{\NGrROnewMTwoZeroVsPKpTwoZeroGenHi}{+75}
\newcommand{\NGrROnewMTwoZeroVsCBFOneLo}{+0.029}
\newcommand{\NGrROnewMTwoZeroVsCBFOneHi}{+0.075}
\newcommand{\NGrROnewMTwoZeroVsCBGenLo}{-187}
\newcommand{\NGrROnewMTwoZeroVsCBGenHi}{-44}
\newcommand{\NGrROnewMTwoZeroVsFRFOneLo}{+0.055}
\newcommand{\NGrROnewMTwoZeroVsFRFOneHi}{+0.104}
\newcommand{\NGrROnewMTwoZeroVsFRGenLo}{-153}
\newcommand{\NGrROnewMTwoZeroVsFRGenHi}{-33}
\newcommand{\NGrROnehMTwoZeroFOne}{0.354}
\newcommand{\NGrROnehMTwoZeroGen}{482}
\newcommand{\NGrROnehMTwoZeroiFOne}{0.349}
\newcommand{\NGrROnehMTwoZeroiGen}{514}
\newcommand{\NGrROnehFRFOne}{0.317}
\newcommand{\NGrROnehFRGen}{560}
\newcommand{\NGrROnehCBFOne}{0.298}
\newcommand{\NGrROnehCBGen}{650}
\newcommand{\NGrROnehPKpTwoZeroFOne}{0.320}
\newcommand{\NGrROnehPKpTwoZeroGen}{561}
\newcommand{\NGrROnehFFOne}{0.346}
\newcommand{\NGrROnehFGen}{468}
\newcommand{\NGrROnehMTwoZeroVsFFOneLo}{-0.011}
\newcommand{\NGrROnehMTwoZeroVsFFOneHi}{+0.026}
\newcommand{\NGrROnehMTwoZeroVsFGenLo}{-22}
\newcommand{\NGrROnehMTwoZeroVsFGenHi}{+53}
\newcommand{\NGrROnehMTwoZeroiVsFFOneLo}{-0.016}
\newcommand{\NGrROnehMTwoZeroiVsFFOneHi}{+0.022}
\newcommand{\NGrROnehMTwoZeroiVsFGenLo}{-5}
\newcommand{\NGrROnehMTwoZeroiVsFGenHi}{+101}
\newcommand{\NGrROnehMTwoZeroVsPKpTwoZeroFOneLo}{+0.015}
\newcommand{\NGrROnehMTwoZeroVsPKpTwoZeroFOneHi}{+0.055}
\newcommand{\NGrROnehMTwoZeroVsPKpTwoZeroGenLo}{-158}
\newcommand{\NGrROnehMTwoZeroVsPKpTwoZeroGenHi}{-6}
\newcommand{\NGrROnehMTwoZeroVsCBFOneLo}{+0.035}
\newcommand{\NGrROnehMTwoZeroVsCBFOneHi}{+0.079}
\newcommand{\NGrROnehMTwoZeroVsCBGenLo}{-257}
\newcommand{\NGrROnehMTwoZeroVsCBGenHi}{-94}
\newcommand{\NGrROnehMTwoZeroVsFRFOneLo}{+0.016}
\newcommand{\NGrROnehMTwoZeroVsFRFOneHi}{+0.060}
\newcommand{\NGrROnehMTwoZeroVsFRGenLo}{-145}
\newcommand{\NGrROnehMTwoZeroVsFRGenHi}{-17}
\newcommand{\NGrNrMTwoZeroCon}{0.280}
\newcommand{\NGrNrMTwoZeroGen}{1488}
\newcommand{\NGrNrMTwoZeroiCon}{0.280}
\newcommand{\NGrNrMTwoZeroiGen}{1568}
\newcommand{\NGrNrFRCon}{0.297}
\newcommand{\NGrNrFRGen}{1421}
\newcommand{\NGrNrCBCon}{0.180}
\newcommand{\NGrNrCBGen}{1376}
\newcommand{\NGrNrPKpTwoZeroCon}{0.297}
\newcommand{\NGrNrPKpTwoZeroGen}{1620}
\newcommand{\NGrNrFCon}{0.282}
\newcommand{\NGrNrFGen}{1492}
\newcommand{\NGrNrMTwoZeroVsFConLo}{-0.045}
\newcommand{\NGrNrMTwoZeroVsFConHi}{+0.043}
\newcommand{\NGrNrMTwoZeroVsFGenLo}{-214}
\newcommand{\NGrNrMTwoZeroVsFGenHi}{+211}
\newcommand{\NGrNrMTwoZeroiVsFConLo}{-0.045}
\newcommand{\NGrNrMTwoZeroiVsFConHi}{+0.040}
\newcommand{\NGrNrMTwoZeroiVsFGenLo}{-134}
\newcommand{\NGrNrMTwoZeroiVsFGenHi}{+278}
\newcommand{\NGrNrMTwoZeroVsPKpTwoZeroConLo}{-0.065}
\newcommand{\NGrNrMTwoZeroVsPKpTwoZeroConHi}{+0.025}
\newcommand{\NGrNrMTwoZeroVsPKpTwoZeroGenLo}{-371}
\newcommand{\NGrNrMTwoZeroVsPKpTwoZeroGenHi}{+95}
\newcommand{\NGrNrMTwoZeroVsCBConLo}{+0.037}
\newcommand{\NGrNrMTwoZeroVsCBConHi}{+0.163}
\newcommand{\NGrNrMTwoZeroVsCBGenLo}{-145}
\newcommand{\NGrNrMTwoZeroVsCBGenHi}{+367}
\newcommand{\NGrNrMTwoZeroVsFRConLo}{-0.060}
\newcommand{\NGrNrMTwoZeroVsFRConHi}{+0.028}
\newcommand{\NGrNrMTwoZeroVsFRGenLo}{-138}
\newcommand{\NGrNrMTwoZeroVsFRGenHi}{+275}
\newcommand{\NFTwoNwMXOneVsMTwoZeroBoxPt}{+0.005}
\newcommand{\NFTwoNwMXOneVsMTwoZeroGenPt}{-34}
\newcommand{\NFOneTwoQOneFourrAKCon}{0.340}
\newcommand{\NFOneTwoQOneFourrAKGen}{1728}
\newcommand{\NFOneTwoQOneFourrLTwoCon}{0.025}
\newcommand{\NFOneTwoQOneFourrLTwoGen}{5366}
\newcommand{\NFOneTwoQOneFourrLThreeTwoCon}{0.255}
\newcommand{\NFOneTwoQOneFourrLThreeTwoGen}{1344}
\newcommand{\NFOneTwoQOneFourrPKCon}{0.398}
\newcommand{\NFOneTwoQOneFourrPKGen}{1324}
\newcommand{\NFOneTwoQOneFourrQZTwoZeroCon}{0.360}
\newcommand{\NFOneTwoQOneFourrQZTwoZeroGen}{1575}
\newcommand{\NFOneTwoQOneFourrQZFiveZeroCon}{0.623}
\newcommand{\NFOneTwoQOneFourrQZFiveZeroGen}{669}
\newcommand{\NFOneTwoQOneFourrMTwoZeroVsPKConLo}{+0.128}
\newcommand{\NFOneTwoQOneFourrMTwoZeroVsPKConHi}{+0.318}
\newcommand{\NFOneTwoQOneFourrMTwoZeroVsPKGenLo}{-1121}
\newcommand{\NFOneTwoQOneFourrMTwoZeroVsPKGenHi}{-639}
\newcommand{\NFOneTwoQOneFourrMTwoZeroVsAKConLo}{+0.198}
\newcommand{\NFOneTwoQOneFourrMTwoZeroVsAKConHi}{+0.367}
\newcommand{\NFOneTwoQOneFourrMTwoZeroVsAKGenLo}{-1537}
\newcommand{\NFOneTwoQOneFourrMTwoZeroVsAKGenHi}{-1018}
\newcommand{\NFOneTwoQOneFourrMTwoZeroVsQZTwoZeroConLo}{+0.168}
\newcommand{\NFOneTwoQOneFourrMTwoZeroVsQZTwoZeroConHi}{+0.350}
\newcommand{\NFOneTwoQOneFourrMTwoZeroVsQZTwoZeroGenLo}{-1407}
\newcommand{\NFOneTwoQOneFourrMTwoZeroVsQZTwoZeroGenHi}{-859}
\newcommand{\NFOneTwoQOneFourrMTwoZeroVsQZFiveZeroConLo}{-0.052}
\newcommand{\NFOneTwoQOneFourrMTwoZeroVsQZFiveZeroConHi}{+0.048}
\newcommand{\NFOneTwoQOneFourrMTwoZeroVsQZFiveZeroGenLo}{-345}
\newcommand{\NFOneTwoQOneFourrMTwoZeroVsQZFiveZeroGenHi}{-89}
\newcommand{\NFOneTwoQOneFourrMTwoZeroVsLTwoConLo}{+0.507}
\newcommand{\NFOneTwoQOneFourrMTwoZeroVsLTwoConHi}{+0.682}
\newcommand{\NFOneTwoQOneFourrMTwoZeroVsLTwoGenLo}{-5119}
\newcommand{\NFOneTwoQOneFourrMTwoZeroVsLTwoGenHi}{-4685}
\newcommand{\NFOneTwoQOneFourrMTwoZeroVsLThreeTwoConLo}{+0.273}
\newcommand{\NFOneTwoQOneFourrMTwoZeroVsLThreeTwoConHi}{+0.460}
\newcommand{\NFOneTwoQOneFourrMTwoZeroVsLThreeTwoGenLo}{-1148}
\newcommand{\NFOneTwoQOneFourrMTwoZeroVsLThreeTwoGenHi}{-651}
\newcommand{\NFOneTwoROnerAKCon}{0.388}
\newcommand{\NFOneTwoROnerAKGen}{1390}
\newcommand{\NFOneTwoROnerLTwoCon}{0.198}
\newcommand{\NFOneTwoROnerLTwoGen}{5908}
\newcommand{\NFOneTwoROnerLThreeTwoCon}{0.260}
\newcommand{\NFOneTwoROnerLThreeTwoGen}{1930}
\newcommand{\NFOneTwoROnerPKCon}{0.427}
\newcommand{\NFOneTwoROnerPKGen}{1556}
\newcommand{\NFOneTwoROnerQZTwoZeroCon}{0.398}
\newcommand{\NFOneTwoROnerQZTwoZeroGen}{1811}
\newcommand{\NFOneTwoROnerQZFiveZeroCon}{0.530}
\newcommand{\NFOneTwoROnerQZFiveZeroGen}{860}
\newcommand{\NFOneTwoROnerMTwoZeroVsPKConLo}{+0.035}
\newcommand{\NFOneTwoROnerMTwoZeroVsPKConHi}{+0.185}
\newcommand{\NFOneTwoROnerMTwoZeroVsPKGenLo}{-1101}
\newcommand{\NFOneTwoROnerMTwoZeroVsPKGenHi}{-401}
\newcommand{\NFOneTwoROnerMTwoZeroVsAKConLo}{+0.070}
\newcommand{\NFOneTwoROnerMTwoZeroVsAKConHi}{+0.223}
\newcommand{\NFOneTwoROnerMTwoZeroVsAKGenLo}{-849}
\newcommand{\NFOneTwoROnerMTwoZeroVsAKGenHi}{-339}
\newcommand{\NFOneTwoROnerMTwoZeroVsQZTwoZeroConLo}{+0.070}
\newcommand{\NFOneTwoROnerMTwoZeroVsQZTwoZeroConHi}{+0.205}
\newcommand{\NFOneTwoROnerMTwoZeroVsQZTwoZeroGenLo}{-1288}
\newcommand{\NFOneTwoROnerMTwoZeroVsQZTwoZeroGenHi}{-748}
\newcommand{\NFOneTwoROnerMTwoZeroVsQZFiveZeroConLo}{-0.048}
\newcommand{\NFOneTwoROnerMTwoZeroVsQZFiveZeroConHi}{+0.058}
\newcommand{\NFOneTwoROnerMTwoZeroVsQZFiveZeroGenLo}{-253}
\newcommand{\NFOneTwoROnerMTwoZeroVsQZFiveZeroGenHi}{+135}
\newcommand{\NFOneTwoROnerMTwoZeroVsLTwoConLo}{+0.258}
\newcommand{\NFOneTwoROnerMTwoZeroVsLTwoConHi}{+0.420}
\newcommand{\NFOneTwoROnerMTwoZeroVsLTwoGenLo}{-5325}
\newcommand{\NFOneTwoROnerMTwoZeroVsLTwoGenHi}{-4881}
\newcommand{\NFOneTwoROnerMTwoZeroVsLThreeTwoConLo}{+0.198}
\newcommand{\NFOneTwoROnerMTwoZeroVsLThreeTwoConHi}{+0.350}
\newcommand{\NFOneTwoROnerMTwoZeroVsLThreeTwoGenLo}{-1434}
\newcommand{\NFOneTwoROnerMTwoZeroVsLThreeTwoGenHi}{-837}
\newcommand{\NFOneTwoQEightwAKFOne}{0.655}
\newcommand{\NFOneTwoQEightwAKGen}{1021}
\newcommand{\NFOneTwoQEightwLTwoFOne}{0.334}
\newcommand{\NFOneTwoQEightwLTwoGen}{2013}
\newcommand{\NFOneTwoQEightwLThreeTwoFOne}{0.461}
\newcommand{\NFOneTwoQEightwLThreeTwoGen}{1777}
\newcommand{\NFOneTwoQEightwPKFOne}{0.665}
\newcommand{\NFOneTwoQEightwPKGen}{917}
\newcommand{\NFOneTwoQEightwQZTwoZeroFOne}{0.710}
\newcommand{\NFOneTwoQEightwQZTwoZeroGen}{623}
\newcommand{\NFOneTwoQEightwQZFiveZeroFOne}{0.757}
\newcommand{\NFOneTwoQEightwQZFiveZeroGen}{468}
\newcommand{\NFOneTwoQEightwLTwoVsFGenPt}{+1563}
\newcommand{\NFOneTwoQEightwMTwoZeroVsPKFOneLo}{+0.044}
\newcommand{\NFOneTwoQEightwMTwoZeroVsPKFOneHi}{+0.098}
\newcommand{\NFOneTwoQEightwMTwoZeroVsPKGenLo}{-498}
\newcommand{\NFOneTwoQEightwMTwoZeroVsPKGenHi}{-270}
\newcommand{\NFOneTwoQEightwMTwoZeroVsAKFOneLo}{+0.047}
\newcommand{\NFOneTwoQEightwMTwoZeroVsAKFOneHi}{+0.113}
\newcommand{\NFOneTwoQEightwMTwoZeroVsAKGenLo}{-592}
\newcommand{\NFOneTwoQEightwMTwoZeroVsAKGenHi}{-382}
\newcommand{\NFOneTwoQEightwMTwoZeroVsQZTwoZeroFOneLo}{+0.008}
\newcommand{\NFOneTwoQEightwMTwoZeroVsQZTwoZeroFOneHi}{+0.042}
\newcommand{\NFOneTwoQEightwMTwoZeroVsQZTwoZeroGenLo}{-152}
\newcommand{\NFOneTwoQEightwMTwoZeroVsQZTwoZeroGenHi}{-25}
\newcommand{\NFOneTwoQEightwMTwoZeroVsQZFiveZeroFOneLo}{-0.039}
\newcommand{\NFOneTwoQEightwMTwoZeroVsQZFiveZeroFOneHi}{-0.007}
\newcommand{\NFOneTwoQEightwMTwoZeroVsQZFiveZeroGenLo}{+33}
\newcommand{\NFOneTwoQEightwMTwoZeroVsQZFiveZeroGenHi}{+112}
\newcommand{\NFOneTwoQEightwMTwoZeroVsLTwoFOneLo}{+0.355}
\newcommand{\NFOneTwoQEightwMTwoZeroVsLTwoFOneHi}{+0.449}
\newcommand{\NFOneTwoQEightwMTwoZeroVsLTwoGenLo}{-1615}
\newcommand{\NFOneTwoQEightwMTwoZeroVsLTwoGenHi}{-1337}
\newcommand{\NFOneTwoQEightwMTwoZeroVsLThreeTwoFOneLo}{+0.230}
\newcommand{\NFOneTwoQEightwMTwoZeroVsLThreeTwoFOneHi}{+0.320}
\newcommand{\NFOneTwoQEightwMTwoZeroVsLThreeTwoGenLo}{-1383}
\newcommand{\NFOneTwoQEightwMTwoZeroVsLThreeTwoGenHi}{-1100}
\newcommand{\NFOneTwoQOneFourwAKFOne}{0.666}
\newcommand{\NFOneTwoQOneFourwAKGen}{721}
\newcommand{\NFOneTwoQOneFourwLTwoFOne}{0.319}
\newcommand{\NFOneTwoQOneFourwLTwoGen}{1379}
\newcommand{\NFOneTwoQOneFourwLThreeTwoFOne}{0.416}
\newcommand{\NFOneTwoQOneFourwLThreeTwoGen}{1416}
\newcommand{\NFOneTwoQOneFourwPKFOne}{0.689}
\newcommand{\NFOneTwoQOneFourwPKGen}{599}
\newcommand{\NFOneTwoQOneFourwQZTwoZeroFOne}{0.709}
\newcommand{\NFOneTwoQOneFourwQZTwoZeroGen}{420}
\newcommand{\NFOneTwoQOneFourwQZFiveZeroFOne}{0.725}
\newcommand{\NFOneTwoQOneFourwQZFiveZeroGen}{355}
\newcommand{\NFOneTwoQOneFourwMTwoZeroVsPKFOneLo}{+0.017}
\newcommand{\NFOneTwoQOneFourwMTwoZeroVsPKFOneHi}{+0.061}
\newcommand{\NFOneTwoQOneFourwMTwoZeroVsPKGenLo}{-279}
\newcommand{\NFOneTwoQOneFourwMTwoZeroVsPKGenHi}{-138}
\newcommand{\NFOneTwoQOneFourwMTwoZeroVsAKFOneLo}{+0.033}
\newcommand{\NFOneTwoQOneFourwMTwoZeroVsAKFOneHi}{+0.090}
\newcommand{\NFOneTwoQOneFourwMTwoZeroVsAKGenLo}{-403}
\newcommand{\NFOneTwoQOneFourwMTwoZeroVsAKGenHi}{-253}
\newcommand{\NFOneTwoQOneFourwMTwoZeroVsQZTwoZeroFOneLo}{+0.001}
\newcommand{\NFOneTwoQOneFourwMTwoZeroVsQZTwoZeroFOneHi}{+0.039}
\newcommand{\NFOneTwoQOneFourwMTwoZeroVsQZTwoZeroGenLo}{-66}
\newcommand{\NFOneTwoQOneFourwMTwoZeroVsQZTwoZeroGenHi}{+15}
\newcommand{\NFOneTwoQOneFourwMTwoZeroVsQZFiveZeroFOneLo}{-0.010}
\newcommand{\NFOneTwoQOneFourwMTwoZeroVsQZFiveZeroFOneHi}{+0.015}
\newcommand{\NFOneTwoQOneFourwMTwoZeroVsQZFiveZeroGenLo}{+18}
\newcommand{\NFOneTwoQOneFourwMTwoZeroVsQZFiveZeroGenHi}{+67}
\newcommand{\NFOneTwoQOneFourwMTwoZeroVsLTwoFOneLo}{+0.359}
\newcommand{\NFOneTwoQOneFourwMTwoZeroVsLTwoFOneHi}{+0.460}
\newcommand{\NFOneTwoQOneFourwMTwoZeroVsLTwoGenLo}{-1096}
\newcommand{\NFOneTwoQOneFourwMTwoZeroVsLTwoGenHi}{-880}
\newcommand{\NFOneTwoQOneFourwMTwoZeroVsLThreeTwoFOneLo}{+0.264}
\newcommand{\NFOneTwoQOneFourwMTwoZeroVsLThreeTwoFOneHi}{+0.360}
\newcommand{\NFOneTwoQOneFourwMTwoZeroVsLThreeTwoGenLo}{-1153}
\newcommand{\NFOneTwoQOneFourwMTwoZeroVsLThreeTwoGenHi}{-894}
\newcommand{\NFOneTwoQEighthAKFOne}{0.660}
\newcommand{\NFOneTwoQEighthAKGen}{957}
\newcommand{\NFOneTwoQEighthLTwoFOne}{0.469}
\newcommand{\NFOneTwoQEighthLTwoGen}{1632}
\newcommand{\NFOneTwoQEighthLThreeTwoFOne}{0.520}
\newcommand{\NFOneTwoQEighthLThreeTwoGen}{1531}
\newcommand{\NFOneTwoQEighthPKFOne}{0.688}
\newcommand{\NFOneTwoQEighthPKGen}{818}
\newcommand{\NFOneTwoQEighthQZTwoZeroFOne}{0.697}
\newcommand{\NFOneTwoQEighthQZTwoZeroGen}{721}
\newcommand{\NFOneTwoQEighthQZFiveZeroFOne}{0.719}
\newcommand{\NFOneTwoQEighthQZFiveZeroGen}{609}
\newcommand{\NFOneTwoQEighthMTwoZeroVsPKFOneLo}{+0.018}
\newcommand{\NFOneTwoQEighthMTwoZeroVsPKFOneHi}{+0.059}
\newcommand{\NFOneTwoQEighthMTwoZeroVsPKGenLo}{-283}
\newcommand{\NFOneTwoQEighthMTwoZeroVsPKGenHi}{-137}
\newcommand{\NFOneTwoQEighthMTwoZeroVsAKFOneLo}{+0.036}
\newcommand{\NFOneTwoQEighthMTwoZeroVsAKFOneHi}{+0.096}
\newcommand{\NFOneTwoQEighthMTwoZeroVsAKGenLo}{-437}
\newcommand{\NFOneTwoQEighthMTwoZeroVsAKGenHi}{-260}
\newcommand{\NFOneTwoQEighthMTwoZeroVsQZTwoZeroFOneLo}{+0.009}
\newcommand{\NFOneTwoQEighthMTwoZeroVsQZTwoZeroFOneHi}{+0.050}
\newcommand{\NFOneTwoQEighthMTwoZeroVsQZTwoZeroGenLo}{-191}
\newcommand{\NFOneTwoQEighthMTwoZeroVsQZTwoZeroGenHi}{-39}
\newcommand{\NFOneTwoQEighthMTwoZeroVsQZFiveZeroFOneLo}{-0.008}
\newcommand{\NFOneTwoQEighthMTwoZeroVsQZFiveZeroFOneHi}{+0.023}
\newcommand{\NFOneTwoQEighthMTwoZeroVsQZFiveZeroGenLo}{-41}
\newcommand{\NFOneTwoQEighthMTwoZeroVsQZFiveZeroGenHi}{+44}
\newcommand{\NFOneTwoQEighthMTwoZeroVsLTwoFOneLo}{+0.215}
\newcommand{\NFOneTwoQEighthMTwoZeroVsLTwoFOneHi}{+0.303}
\newcommand{\NFOneTwoQEighthMTwoZeroVsLTwoGenLo}{-1150}
\newcommand{\NFOneTwoQEighthMTwoZeroVsLTwoGenHi}{-898}
\newcommand{\NFOneTwoQEighthMTwoZeroVsLThreeTwoFOneLo}{+0.164}
\newcommand{\NFOneTwoQEighthMTwoZeroVsLThreeTwoFOneHi}{+0.248}
\newcommand{\NFOneTwoQEighthMTwoZeroVsLThreeTwoGenLo}{-1062}
\newcommand{\NFOneTwoQEighthMTwoZeroVsLThreeTwoGenHi}{-788}
\newcommand{\NFOneTwoQOneFourhAKFOne}{0.687}
\newcommand{\NFOneTwoQOneFourhAKGen}{734}
\newcommand{\NFOneTwoQOneFourhLTwoFOne}{0.461}
\newcommand{\NFOneTwoQOneFourhLTwoGen}{1130}
\newcommand{\NFOneTwoQOneFourhLThreeTwoFOne}{0.527}
\newcommand{\NFOneTwoQOneFourhLThreeTwoGen}{945}
\newcommand{\NFOneTwoQOneFourhPKFOne}{0.710}
\newcommand{\NFOneTwoQOneFourhPKGen}{605}
\newcommand{\NFOneTwoQOneFourhQZTwoZeroFOne}{0.712}
\newcommand{\NFOneTwoQOneFourhQZTwoZeroGen}{532}
\newcommand{\NFOneTwoQOneFourhQZFiveZeroFOne}{0.733}
\newcommand{\NFOneTwoQOneFourhQZFiveZeroGen}{483}
\newcommand{\NFOneTwoQOneFourhMTwoZeroVsPKFOneLo}{+0.007}
\newcommand{\NFOneTwoQOneFourhMTwoZeroVsPKFOneHi}{+0.041}
\newcommand{\NFOneTwoQOneFourhMTwoZeroVsPKGenLo}{-176}
\newcommand{\NFOneTwoQOneFourhMTwoZeroVsPKGenHi}{-72}
\newcommand{\NFOneTwoQOneFourhMTwoZeroVsAKFOneLo}{+0.021}
\newcommand{\NFOneTwoQOneFourhMTwoZeroVsAKFOneHi}{+0.073}
\newcommand{\NFOneTwoQOneFourhMTwoZeroVsAKGenLo}{-318}
\newcommand{\NFOneTwoQOneFourhMTwoZeroVsAKGenHi}{-185}
\newcommand{\NFOneTwoQOneFourhMTwoZeroVsQZTwoZeroFOneLo}{+0.008}
\newcommand{\NFOneTwoQOneFourhMTwoZeroVsQZTwoZeroFOneHi}{+0.037}
\newcommand{\NFOneTwoQOneFourhMTwoZeroVsQZTwoZeroGenLo}{-84}
\newcommand{\NFOneTwoQOneFourhMTwoZeroVsQZTwoZeroGenHi}{-8}
\newcommand{\NFOneTwoQOneFourhMTwoZeroVsQZFiveZeroFOneLo}{-0.013}
\newcommand{\NFOneTwoQOneFourhMTwoZeroVsQZFiveZeroFOneHi}{+0.014}
\newcommand{\NFOneTwoQOneFourhMTwoZeroVsQZFiveZeroGenLo}{-26}
\newcommand{\NFOneTwoQOneFourhMTwoZeroVsQZFiveZeroGenHi}{+28}
\newcommand{\NFOneTwoQOneFourhMTwoZeroVsLTwoFOneLo}{+0.228}
\newcommand{\NFOneTwoQOneFourhMTwoZeroVsLTwoFOneHi}{+0.319}
\newcommand{\NFOneTwoQOneFourhMTwoZeroVsLTwoGenLo}{-745}
\newcommand{\NFOneTwoQOneFourhMTwoZeroVsLTwoGenHi}{-545}
\newcommand{\NFOneTwoQOneFourhMTwoZeroVsLThreeTwoFOneLo}{+0.167}
\newcommand{\NFOneTwoQOneFourhMTwoZeroVsLThreeTwoFOneHi}{+0.247}
\newcommand{\NFOneTwoQOneFourhMTwoZeroVsLThreeTwoGenLo}{-545}
\newcommand{\NFOneTwoQOneFourhMTwoZeroVsLThreeTwoGenHi}{-379}
\newcommand{\NFOneTwoNhAKBox}{0.466}
\newcommand{\NFOneTwoNhAKGen}{1442}
\newcommand{\NFOneTwoNhLTwoBox}{0.278}
\newcommand{\NFOneTwoNhLTwoGen}{1800}
\newcommand{\NFOneTwoNhLThreeTwoBox}{0.316}
\newcommand{\NFOneTwoNhLThreeTwoGen}{1850}
\newcommand{\NFOneTwoNhPKBox}{0.444}
\newcommand{\NFOneTwoNhPKGen}{1178}
\newcommand{\NFOneTwoNhQZTwoZeroBox}{0.453}
\newcommand{\NFOneTwoNhQZTwoZeroGen}{1256}
\newcommand{\NFOneTwoNhQZFiveZeroBox}{0.529}
\newcommand{\NFOneTwoNhQZFiveZeroGen}{1036}
\newcommand{\NFOneTwoNhMTwoZeroVsPKBoxLo}{+0.091}
\newcommand{\NFOneTwoNhMTwoZeroVsPKBoxHi}{+0.149}
\newcommand{\NFOneTwoNhMTwoZeroVsPKGenLo}{-309}
\newcommand{\NFOneTwoNhMTwoZeroVsPKGenHi}{-97}
\newcommand{\NFOneTwoNhMTwoZeroVsAKBoxLo}{+0.065}
\newcommand{\NFOneTwoNhMTwoZeroVsAKBoxHi}{+0.129}
\newcommand{\NFOneTwoNhMTwoZeroVsAKGenLo}{-604}
\newcommand{\NFOneTwoNhMTwoZeroVsAKGenHi}{-335}
\newcommand{\NFOneTwoNhMTwoZeroVsQZTwoZeroBoxLo}{+0.082}
\newcommand{\NFOneTwoNhMTwoZeroVsQZTwoZeroBoxHi}{+0.141}
\newcommand{\NFOneTwoNhMTwoZeroVsQZTwoZeroGenLo}{-406}
\newcommand{\NFOneTwoNhMTwoZeroVsQZTwoZeroGenHi}{-154}
\newcommand{\NFOneTwoNhMTwoZeroVsQZFiveZeroBoxLo}{+0.011}
\newcommand{\NFOneTwoNhMTwoZeroVsQZFiveZeroBoxHi}{+0.059}
\newcommand{\NFOneTwoNhMTwoZeroVsQZFiveZeroGenLo}{-162}
\newcommand{\NFOneTwoNhMTwoZeroVsQZFiveZeroGenHi}{+36}
\newcommand{\NFOneTwoNhMTwoZeroVsLTwoBoxLo}{+0.246}
\newcommand{\NFOneTwoNhMTwoZeroVsLTwoBoxHi}{+0.325}
\newcommand{\NFOneTwoNhMTwoZeroVsLTwoGenLo}{-1013}
\newcommand{\NFOneTwoNhMTwoZeroVsLTwoGenHi}{-636}
\newcommand{\NFOneTwoNhMTwoZeroVsLThreeTwoBoxLo}{+0.210}
\newcommand{\NFOneTwoNhMTwoZeroVsLThreeTwoBoxHi}{+0.288}
\newcommand{\NFOneTwoNhMTwoZeroVsLThreeTwoGenLo}{-1062}
\newcommand{\NFOneTwoNhMTwoZeroVsLThreeTwoGenHi}{-687}
\newcommand{\NFOneTwoROnewAKFOne}{0.378}
\newcommand{\NFOneTwoROnewAKGen}{601}
\newcommand{\NFOneTwoROnewLTwoFOne}{0.175}
\newcommand{\NFOneTwoROnewLTwoGen}{1237}
\newcommand{\NFOneTwoROnewLThreeTwoFOne}{0.235}
\newcommand{\NFOneTwoROnewLThreeTwoGen}{1040}
\newcommand{\NFOneTwoROnewPKFOne}{0.390}
\newcommand{\NFOneTwoROnewPKGen}{434}
\newcommand{\NFOneTwoROnewQZTwoZeroFOne}{0.398}
\newcommand{\NFOneTwoROnewQZTwoZeroGen}{363}
\newcommand{\NFOneTwoROnewQZFiveZeroFOne}{0.400}
\newcommand{\NFOneTwoROnewQZFiveZeroGen}{361}
\newcommand{\NFOneTwoROnewMTwoZeroVsPKFOneLo}{+0.001}
\newcommand{\NFOneTwoROnewMTwoZeroVsPKFOneHi}{+0.040}
\newcommand{\NFOneTwoROnewMTwoZeroVsPKGenLo}{-54}
\newcommand{\NFOneTwoROnewMTwoZeroVsPKGenHi}{+54}
\newcommand{\NFOneTwoROnewMTwoZeroVsAKFOneLo}{+0.011}
\newcommand{\NFOneTwoROnewMTwoZeroVsAKFOneHi}{+0.055}
\newcommand{\NFOneTwoROnewMTwoZeroVsAKGenLo}{-238}
\newcommand{\NFOneTwoROnewMTwoZeroVsAKGenHi}{-107}
\newcommand{\NFOneTwoROnewMTwoZeroVsQZTwoZeroFOneLo}{-0.010}
\newcommand{\NFOneTwoROnewMTwoZeroVsQZTwoZeroFOneHi}{+0.034}
\newcommand{\NFOneTwoROnewMTwoZeroVsQZTwoZeroGenLo}{+26}
\newcommand{\NFOneTwoROnewMTwoZeroVsQZTwoZeroGenHi}{+116}
\newcommand{\NFOneTwoROnewMTwoZeroVsQZFiveZeroFOneLo}{-0.009}
\newcommand{\NFOneTwoROnewMTwoZeroVsQZFiveZeroFOneHi}{+0.031}
\newcommand{\NFOneTwoROnewMTwoZeroVsQZFiveZeroGenLo}{+29}
\newcommand{\NFOneTwoROnewMTwoZeroVsQZFiveZeroGenHi}{+120}
\newcommand{\NFOneTwoROnewMTwoZeroVsLTwoFOneLo}{+0.201}
\newcommand{\NFOneTwoROnewMTwoZeroVsLTwoFOneHi}{+0.270}
\newcommand{\NFOneTwoROnewMTwoZeroVsLTwoGenLo}{-924}
\newcommand{\NFOneTwoROnewMTwoZeroVsLTwoGenHi}{-685}
\newcommand{\NFOneTwoROnewMTwoZeroVsLThreeTwoFOneLo}{+0.145}
\newcommand{\NFOneTwoROnewMTwoZeroVsLThreeTwoFOneHi}{+0.206}
\newcommand{\NFOneTwoROnewMTwoZeroVsLThreeTwoGenLo}{-713}
\newcommand{\NFOneTwoROnewMTwoZeroVsLThreeTwoGenHi}{-499}
\newcommand{\NFOneTwoROnehAKFOne}{0.325}
\newcommand{\NFOneTwoROnehAKGen}{650}
\newcommand{\NFOneTwoROnehLTwoFOne}{0.225}
\newcommand{\NFOneTwoROnehLTwoGen}{962}
\newcommand{\NFOneTwoROnehLThreeTwoFOne}{0.236}
\newcommand{\NFOneTwoROnehLThreeTwoGen}{918}
\newcommand{\NFOneTwoROnehPKFOne}{0.331}
\newcommand{\NFOneTwoROnehPKGen}{549}
\newcommand{\NFOneTwoROnehQZTwoZeroFOne}{0.333}
\newcommand{\NFOneTwoROnehQZTwoZeroGen}{529}
\newcommand{\NFOneTwoROnehQZFiveZeroFOne}{0.345}
\newcommand{\NFOneTwoROnehQZFiveZeroGen}{463}
\newcommand{\NFOneTwoROnehMTwoZeroVsPKFOneLo}{+0.004}
\newcommand{\NFOneTwoROnehMTwoZeroVsPKFOneHi}{+0.044}
\newcommand{\NFOneTwoROnehMTwoZeroVsPKGenLo}{-124}
\newcommand{\NFOneTwoROnehMTwoZeroVsPKGenHi}{-15}
\newcommand{\NFOneTwoROnehMTwoZeroVsAKFOneLo}{+0.008}
\newcommand{\NFOneTwoROnehMTwoZeroVsAKFOneHi}{+0.050}
\newcommand{\NFOneTwoROnehMTwoZeroVsAKGenLo}{-235}
\newcommand{\NFOneTwoROnehMTwoZeroVsAKGenHi}{-106}
\newcommand{\NFOneTwoROnehMTwoZeroVsQZTwoZeroFOneLo}{+0.003}
\newcommand{\NFOneTwoROnehMTwoZeroVsQZTwoZeroFOneHi}{+0.041}
\newcommand{\NFOneTwoROnehMTwoZeroVsQZTwoZeroGenLo}{-95}
\newcommand{\NFOneTwoROnehMTwoZeroVsQZTwoZeroGenHi}{-1}
\newcommand{\NFOneTwoROnehMTwoZeroVsQZFiveZeroFOneLo}{-0.008}
\newcommand{\NFOneTwoROnehMTwoZeroVsQZFiveZeroFOneHi}{+0.026}
\newcommand{\NFOneTwoROnehMTwoZeroVsQZFiveZeroGenLo}{-34}
\newcommand{\NFOneTwoROnehMTwoZeroVsQZFiveZeroGenHi}{+70}
\newcommand{\NFOneTwoROnehMTwoZeroVsLTwoFOneLo}{+0.100}
\newcommand{\NFOneTwoROnehMTwoZeroVsLTwoFOneHi}{+0.164}
\newcommand{\NFOneTwoROnehMTwoZeroVsLTwoGenLo}{-586}
\newcommand{\NFOneTwoROnehMTwoZeroVsLTwoGenHi}{-383}
\newcommand{\NFOneTwoROnehMTwoZeroVsLThreeTwoFOneLo}{+0.090}
\newcommand{\NFOneTwoROnehMTwoZeroVsLThreeTwoFOneHi}{+0.147}
\newcommand{\NFOneTwoROnehMTwoZeroVsLThreeTwoGenLo}{-556}
\newcommand{\NFOneTwoROnehMTwoZeroVsLThreeTwoGenHi}{-328}
\newcommand{\NFOneTwoNrAKCon}{0.203}
\newcommand{\NFOneTwoNrAKGen}{2362}
\newcommand{\NFOneTwoNrLTwoCon}{0.117}
\newcommand{\NFOneTwoNrLTwoGen}{458}
\newcommand{\NFOneTwoNrLThreeTwoCon}{0.172}
\newcommand{\NFOneTwoNrLThreeTwoGen}{1273}
\newcommand{\NFOneTwoNrPKCon}{0.297}
\newcommand{\NFOneTwoNrPKGen}{1372}
\newcommand{\NFOneTwoNrQZTwoZeroCon}{0.300}
\newcommand{\NFOneTwoNrQZTwoZeroGen}{1087}
\newcommand{\NFOneTwoNrQZFiveZeroCon}{0.318}
\newcommand{\NFOneTwoNrQZFiveZeroGen}{1627}
\newcommand{\NFOneTwoNrMTwoZeroVsPKConLo}{-0.068}
\newcommand{\NFOneTwoNrMTwoZeroVsPKConHi}{+0.030}
\newcommand{\NFOneTwoNrMTwoZeroVsPKGenLo}{-91}
\newcommand{\NFOneTwoNrMTwoZeroVsPKGenHi}{+323}
\newcommand{\NFOneTwoNrMTwoZeroVsAKConLo}{+0.030}
\newcommand{\NFOneTwoNrMTwoZeroVsAKConHi}{+0.130}
\newcommand{\NFOneTwoNrMTwoZeroVsAKGenLo}{-1117}
\newcommand{\NFOneTwoNrMTwoZeroVsAKGenHi}{-641}
\newcommand{\NFOneTwoNrMTwoZeroVsQZTwoZeroConLo}{-0.077}
\newcommand{\NFOneTwoNrMTwoZeroVsQZTwoZeroConHi}{+0.035}
\newcommand{\NFOneTwoNrMTwoZeroVsQZTwoZeroGenLo}{+179}
\newcommand{\NFOneTwoNrMTwoZeroVsQZTwoZeroGenHi}{+635}
\newcommand{\NFOneTwoNrMTwoZeroVsQZFiveZeroConLo}{-0.080}
\newcommand{\NFOneTwoNrMTwoZeroVsQZFiveZeroConHi}{+0.005}
\newcommand{\NFOneTwoNrMTwoZeroVsQZFiveZeroGenLo}{-381}
\newcommand{\NFOneTwoNrMTwoZeroVsQZFiveZeroGenHi}{+83}
\newcommand{\NFOneTwoNrMTwoZeroVsLTwoConLo}{+0.102}
\newcommand{\NFOneTwoNrMTwoZeroVsLTwoConHi}{+0.225}
\newcommand{\NFOneTwoNrMTwoZeroVsLTwoGenLo}{+780}
\newcommand{\NFOneTwoNrMTwoZeroVsLTwoGenHi}{+1261}
\newcommand{\NFOneTwoNrMTwoZeroVsLThreeTwoConLo}{+0.048}
\newcommand{\NFOneTwoNrMTwoZeroVsLThreeTwoConHi}{+0.165}
\newcommand{\NFOneTwoNrMTwoZeroVsLThreeTwoGenLo}{-73}
\newcommand{\NFOneTwoNrMTwoZeroVsLThreeTwoGenHi}{+500}
\newcommand{\NFOneTwoQOneFourrMFourZeroCon}{0.657}
\newcommand{\NFOneTwoQOneFourrMFourZeroGen}{437}
\newcommand{\NFOneTwoQOneFourrMFourZeroiCon}{0.652}
\newcommand{\NFOneTwoQOneFourrMFourZeroiGen}{442}
\newcommand{\NFOneTwoQOneFourrMFourZeroVsFConLo}{-0.020}
\newcommand{\NFOneTwoQOneFourrMFourZeroVsFConHi}{+0.043}
\newcommand{\NFOneTwoQOneFourrMFourZeroVsFGenLo}{-41}
\newcommand{\NFOneTwoQOneFourrMFourZeroVsFGenHi}{+46}
\newcommand{\NFOneTwoQOneFourrMFourZeroiVsFConLo}{-0.022}
\newcommand{\NFOneTwoQOneFourrMFourZeroiVsFConHi}{+0.040}
\newcommand{\NFOneTwoQOneFourrMFourZeroiVsFGenLo}{-45}
\newcommand{\NFOneTwoQOneFourrMFourZeroiVsFGenHi}{+61}
\newcommand{\NFOneTwoQEighthMFourZeroFOne}{0.732}
\newcommand{\NFOneTwoQEighthMFourZeroGen}{567}
\newcommand{\NFOneTwoQEighthMFourZeroiFOne}{0.732}
\newcommand{\NFOneTwoQEighthMFourZeroiGen}{580}
\newcommand{\NFOneTwoQEighthMFourZeroVsFFOneLo}{-0.002}
\newcommand{\NFOneTwoQEighthMFourZeroVsFFOneHi}{+0.021}
\newcommand{\NFOneTwoQEighthMFourZeroVsFGenLo}{-58}
\newcommand{\NFOneTwoQEighthMFourZeroVsFGenHi}{+12}
\newcommand{\NFOneTwoQEighthMFourZeroiVsFFOneLo}{-0.003}
\newcommand{\NFOneTwoQEighthMFourZeroiVsFFOneHi}{+0.022}
\newcommand{\NFOneTwoQEighthMFourZeroiVsFGenLo}{-44}
\newcommand{\NFOneTwoQEighthMFourZeroiVsFGenHi}{+23}
\newcommand{\NFOneTwoQOneFourhMFourZeroFOne}{0.733}
\newcommand{\NFOneTwoQOneFourhMFourZeroGen}{482}
\newcommand{\NFOneTwoQOneFourhMFourZeroiFOne}{0.737}
\newcommand{\NFOneTwoQOneFourhMFourZeroiGen}{473}
\newcommand{\NFOneTwoQOneFourhMFourZeroVsFFOneLo}{-0.018}
\newcommand{\NFOneTwoQOneFourhMFourZeroVsFFOneHi}{+0.006}
\newcommand{\NFOneTwoQOneFourhMFourZeroVsFGenLo}{-23}
\newcommand{\NFOneTwoQOneFourhMFourZeroVsFGenHi}{+24}
\newcommand{\NFOneTwoQOneFourhMFourZeroiVsFFOneLo}{-0.013}
\newcommand{\NFOneTwoQOneFourhMFourZeroiVsFFOneHi}{+0.010}
\newcommand{\NFOneTwoQOneFourhMFourZeroiVsFGenLo}{-28}
\newcommand{\NFOneTwoQOneFourhMFourZeroiVsFGenHi}{+12}
\newcommand{\NFOneTwoROnewMFourZeroFOne}{0.409}
\newcommand{\NFOneTwoROnewMFourZeroGen}{390}
\newcommand{\NFOneTwoROnewMFourZeroiFOne}{0.413}
\newcommand{\NFOneTwoROnewMFourZeroiGen}{383}
\newcommand{\NFOneTwoROnewMFourZeroVsFFOneLo}{-0.009}
\newcommand{\NFOneTwoROnewMFourZeroVsFFOneHi}{+0.030}
\newcommand{\NFOneTwoROnewMFourZeroVsFGenLo}{-14}
\newcommand{\NFOneTwoROnewMFourZeroVsFGenHi}{+60}
\newcommand{\NFOneTwoROnewMFourZeroiVsFFOneLo}{-0.005}
\newcommand{\NFOneTwoROnewMFourZeroiVsFFOneHi}{+0.032}
\newcommand{\NFOneTwoROnewMFourZeroiVsFGenLo}{-24}
\newcommand{\NFOneTwoROnewMFourZeroiVsFGenHi}{+57}
\newcommand{\NFOneTwoROnehMFourZeroFOne}{0.360}
\newcommand{\NFOneTwoROnehMFourZeroGen}{470}
\newcommand{\NFOneTwoROnehMFourZeroiFOne}{0.347}
\newcommand{\NFOneTwoROnehMFourZeroiGen}{495}
\newcommand{\NFOneTwoROnehMFourZeroVsFFOneLo}{-0.002}
\newcommand{\NFOneTwoROnehMFourZeroVsFFOneHi}{+0.030}
\newcommand{\NFOneTwoROnehMFourZeroVsFGenLo}{-36}
\newcommand{\NFOneTwoROnehMFourZeroVsFGenHi}{+41}
\newcommand{\NFOneTwoROnehMFourZeroiVsFFOneLo}{-0.016}
\newcommand{\NFOneTwoROnehMFourZeroiVsFFOneHi}{+0.017}
\newcommand{\NFOneTwoROnehMFourZeroiVsFGenLo}{-9}
\newcommand{\NFOneTwoROnehMFourZeroiVsFGenHi}{+66}
\newcommand{\NFOneTwoNrMFourZeroCon}{0.280}
\newcommand{\NFOneTwoNrMFourZeroGen}{1457}
\newcommand{\NFOneTwoNrMFourZeroiCon}{0.292}
\newcommand{\NFOneTwoNrMFourZeroiGen}{1525}
\newcommand{\NFOneTwoNrMFourZeroVsFConLo}{-0.040}
\newcommand{\NFOneTwoNrMFourZeroVsFConHi}{+0.043}
\newcommand{\NFOneTwoNrMFourZeroVsFGenLo}{-222}
\newcommand{\NFOneTwoNrMFourZeroVsFGenHi}{+153}
\newcommand{\NFOneTwoNrMFourZeroiVsFConLo}{-0.033}
\newcommand{\NFOneTwoNrMFourZeroiVsFConHi}{+0.055}
\newcommand{\NFOneTwoNrMFourZeroiVsFGenLo}{-172}
\newcommand{\NFOneTwoNrMFourZeroiVsFGenHi}{+238}
\newcommand{\NFOneTwoNwMFourZeroiBox}{0.600}
\newcommand{\NFOneTwoNwMFourZeroiGen}{833}
\newcommand{\NFOneTwoNwMFourZeroiVsFBoxLo}{-0.035}
\newcommand{\NFOneTwoNwMFourZeroiVsFBoxHi}{+0.013}
\newcommand{\NFOneTwoNwMFourZeroiVsFGenLo}{-52}
\newcommand{\NFOneTwoNwMFourZeroiVsFGenHi}{+122}
\newcommand{\NSkvQEightrSKOneZeroCon}{0.468}
\newcommand{\NSkvQEightrSKOneZeroGen}{930}
\newcommand{\NSkvQEightrSKTwoZeroCon}{0.520}
\newcommand{\NSkvQEightrSKTwoZeroGen}{802}
\newcommand{\NSkvQEightrSKOneZeroVsMTwoZeroConLo}{-0.163}
\newcommand{\NSkvQEightrSKOneZeroVsMTwoZeroConHi}{-0.013}
\newcommand{\NSkvQEightrSKOneZeroVsMTwoZeroGenLo}{+92}
\newcommand{\NSkvQEightrSKOneZeroVsMTwoZeroGenHi}{+394}
\newcommand{\NSkvQEightrSKTwoZeroVsMTwoZeroConLo}{-0.083}
\newcommand{\NSkvQEightrSKTwoZeroVsMTwoZeroConHi}{+0.013}
\newcommand{\NSkvQEightrSKTwoZeroVsMTwoZeroGenLo}{-16}
\newcommand{\NSkvQEightrSKTwoZeroVsMTwoZeroGenHi}{+243}
\newcommand{\NSkvQOneFourrSKOneZeroCon}{0.585}
\newcommand{\NSkvQOneFourrSKOneZeroGen}{730}
\newcommand{\NSkvQOneFourrSKTwoZeroCon}{0.618}
\newcommand{\NSkvQOneFourrSKTwoZeroGen}{596}
\newcommand{\NSkvQOneFourrSKOneZeroVsMTwoZeroConLo}{-0.090}
\newcommand{\NSkvQOneFourrSKOneZeroVsMTwoZeroConHi}{+0.020}
\newcommand{\NSkvQOneFourrSKOneZeroVsMTwoZeroGenLo}{+139}
\newcommand{\NSkvQOneFourrSKOneZeroVsMTwoZeroGenHi}{+430}
\newcommand{\NSkvQOneFourrSKTwoZeroVsMTwoZeroConLo}{-0.043}
\newcommand{\NSkvQOneFourrSKTwoZeroVsMTwoZeroConHi}{+0.040}
\newcommand{\NSkvQOneFourrSKTwoZeroVsMTwoZeroGenLo}{+51}
\newcommand{\NSkvQOneFourrSKTwoZeroVsMTwoZeroGenHi}{+239}
\newcommand{\NSkvROnerSKOneZeroCon}{0.482}
\newcommand{\NSkvROnerSKOneZeroGen}{764}
\newcommand{\NSkvROnerSKTwoZeroCon}{0.542}
\newcommand{\NSkvROnerSKTwoZeroGen}{641}
\newcommand{\NSkvROnerSKOneZeroVsMTwoZeroConLo}{-0.102}
\newcommand{\NSkvROnerSKOneZeroVsMTwoZeroConHi}{-0.005}
\newcommand{\NSkvROnerSKOneZeroVsMTwoZeroGenLo}{-241}
\newcommand{\NSkvROnerSKOneZeroVsMTwoZeroGenHi}{+171}
\newcommand{\NSkvROnerSKTwoZeroVsMTwoZeroConLo}{-0.037}
\newcommand{\NSkvROnerSKTwoZeroVsMTwoZeroConHi}{+0.052}
\newcommand{\NSkvROnerSKTwoZeroVsMTwoZeroGenLo}{-311}
\newcommand{\NSkvROnerSKTwoZeroVsMTwoZeroGenHi}{-6}
\newcommand{\NSkvQEightwSKOneZeroFOne}{0.692}
\newcommand{\NSkvQEightwSKOneZeroGen}{717}
\newcommand{\NSkvQEightwSKTwoZeroFOne}{0.724}
\newcommand{\NSkvQEightwSKTwoZeroGen}{584}
\newcommand{\NSkvQEightwSKOneZeroVsMTwoZeroFOneLo}{-0.066}
\newcommand{\NSkvQEightwSKOneZeroVsMTwoZeroFOneHi}{-0.021}
\newcommand{\NSkvQEightwSKOneZeroVsMTwoZeroGenLo}{+117}
\newcommand{\NSkvQEightwSKOneZeroVsMTwoZeroGenHi}{+250}
\newcommand{\NSkvQEightwSKTwoZeroVsMTwoZeroFOneLo}{-0.028}
\newcommand{\NSkvQEightwSKTwoZeroVsMTwoZeroFOneHi}{+0.007}
\newcommand{\NSkvQEightwSKTwoZeroVsMTwoZeroGenLo}{+5}
\newcommand{\NSkvQEightwSKTwoZeroVsMTwoZeroGenHi}{+90}
\newcommand{\NSkvQOneFourwSKOneZeroFOne}{0.709}
\newcommand{\NSkvQOneFourwSKOneZeroGen}{451}
\newcommand{\NSkvQOneFourwSKTwoZeroFOne}{0.731}
\newcommand{\NSkvQOneFourwSKTwoZeroGen}{395}
\newcommand{\NSkvQOneFourwSKOneZeroVsMTwoZeroFOneLo}{-0.038}
\newcommand{\NSkvQOneFourwSKOneZeroVsMTwoZeroFOneHi}{-0.001}
\newcommand{\NSkvQOneFourwSKOneZeroVsMTwoZeroGenLo}{+25}
\newcommand{\NSkvQOneFourwSKOneZeroVsMTwoZeroGenHi}{+87}
\newcommand{\NSkvQOneFourwSKTwoZeroVsMTwoZeroFOneLo}{-0.006}
\newcommand{\NSkvQOneFourwSKTwoZeroVsMTwoZeroFOneHi}{+0.013}
\newcommand{\NSkvQOneFourwSKTwoZeroVsMTwoZeroGenLo}{-19}
\newcommand{\NSkvQOneFourwSKTwoZeroVsMTwoZeroGenHi}{+19}
\newcommand{\NSkvNwSKOneZeroBox}{0.493}
\newcommand{\NSkvNwSKOneZeroGen}{1304}
\newcommand{\NSkvNwSKTwoZeroBox}{0.543}
\newcommand{\NSkvNwSKTwoZeroGen}{1156}
\newcommand{\NSkvNwSKOneZeroVsMTwoZeroBoxLo}{-0.142}
\newcommand{\NSkvNwSKOneZeroVsMTwoZeroBoxHi}{-0.072}
\newcommand{\NSkvNwSKOneZeroVsMTwoZeroGenLo}{+190}
\newcommand{\NSkvNwSKOneZeroVsMTwoZeroGenHi}{+464}
\newcommand{\NSkvNwSKTwoZeroVsMTwoZeroBoxLo}{-0.082}
\newcommand{\NSkvNwSKTwoZeroVsMTwoZeroBoxHi}{-0.031}
\newcommand{\NSkvNwSKTwoZeroVsMTwoZeroGenLo}{+68}
\newcommand{\NSkvNwSKTwoZeroVsMTwoZeroGenHi}{+296}
\newcommand{\NSkvQEighthSKOneZeroFOne}{0.708}
\newcommand{\NSkvQEighthSKOneZeroGen}{700}
\newcommand{\NSkvQEighthSKTwoZeroFOne}{0.712}
\newcommand{\NSkvQEighthSKTwoZeroGen}{643}
\newcommand{\NSkvQEighthSKOneZeroVsFGenPt}{+110}
\newcommand{\NSkvQEighthSKOneZeroVsMTwoZeroFOneLo}{-0.035}
\newcommand{\NSkvQEighthSKOneZeroVsMTwoZeroFOneHi}{-0.000}
\newcommand{\NSkvQEighthSKOneZeroVsMTwoZeroGenLo}{+40}
\newcommand{\NSkvQEighthSKOneZeroVsMTwoZeroGenHi}{+137}
\newcommand{\NSkvQEighthSKTwoZeroVsMTwoZeroFOneLo}{-0.031}
\newcommand{\NSkvQEighthSKTwoZeroVsMTwoZeroFOneHi}{+0.002}
\newcommand{\NSkvQEighthSKTwoZeroVsMTwoZeroGenLo}{-22}
\newcommand{\NSkvQEighthSKTwoZeroVsMTwoZeroGenHi}{+81}
\newcommand{\NSkvQOneFourhSKOneZeroFOne}{0.720}
\newcommand{\NSkvQOneFourhSKOneZeroGen}{534}
\newcommand{\NSkvQOneFourhSKTwoZeroFOne}{0.728}
\newcommand{\NSkvQOneFourhSKTwoZeroGen}{496}
\newcommand{\NSkvQOneFourhSKOneZeroVsMTwoZeroFOneLo}{-0.028}
\newcommand{\NSkvQOneFourhSKOneZeroVsMTwoZeroFOneHi}{-0.000}
\newcommand{\NSkvQOneFourhSKOneZeroVsMTwoZeroGenLo}{+20}
\newcommand{\NSkvQOneFourhSKOneZeroVsMTwoZeroGenHi}{+74}
\newcommand{\NSkvQOneFourhSKTwoZeroVsMTwoZeroFOneLo}{-0.017}
\newcommand{\NSkvQOneFourhSKTwoZeroVsMTwoZeroFOneHi}{+0.005}
\newcommand{\NSkvQOneFourhSKTwoZeroVsMTwoZeroGenLo}{-16}
\newcommand{\NSkvQOneFourhSKTwoZeroVsMTwoZeroGenHi}{+42}
\newcommand{\NSkvNhSKOneZeroBox}{0.456}
\newcommand{\NSkvNhSKOneZeroGen}{1227}
\newcommand{\NSkvNhSKTwoZeroBox}{0.500}
\newcommand{\NSkvNhSKTwoZeroGen}{1106}
\newcommand{\NSkvNhSKOneZeroVsMTwoZeroBoxLo}{-0.136}
\newcommand{\NSkvNhSKOneZeroVsMTwoZeroBoxHi}{-0.078}
\newcommand{\NSkvNhSKOneZeroVsMTwoZeroGenLo}{+142}
\newcommand{\NSkvNhSKOneZeroVsMTwoZeroGenHi}{+368}
\newcommand{\NSkvNhSKTwoZeroVsMTwoZeroBoxLo}{-0.089}
\newcommand{\NSkvNhSKTwoZeroVsMTwoZeroBoxHi}{-0.039}
\newcommand{\NSkvNhSKTwoZeroVsMTwoZeroGenLo}{+25}
\newcommand{\NSkvNhSKTwoZeroVsMTwoZeroGenHi}{+239}
\newcommand{\NSkvROnewSKOneZeroFOne}{0.373}
\newcommand{\NSkvROnewSKOneZeroGen}{535}
\newcommand{\NSkvROnewSKTwoZeroFOne}{0.400}
\newcommand{\NSkvROnewSKTwoZeroGen}{435}
\newcommand{\NSkvROnewSKOneZeroVsMTwoZeroFOneLo}{-0.060}
\newcommand{\NSkvROnewSKOneZeroVsMTwoZeroFOneHi}{-0.016}
\newcommand{\NSkvROnewSKOneZeroVsMTwoZeroGenLo}{+41}
\newcommand{\NSkvROnewSKOneZeroVsMTwoZeroGenHi}{+167}
\newcommand{\NSkvROnewSKTwoZeroVsMTwoZeroFOneLo}{-0.029}
\newcommand{\NSkvROnewSKTwoZeroVsMTwoZeroFOneHi}{+0.009}
\newcommand{\NSkvROnewSKTwoZeroVsMTwoZeroGenLo}{-54}
\newcommand{\NSkvROnewSKTwoZeroVsMTwoZeroGenHi}{+58}
\newcommand{\NSkvROnehSKOneZeroFOne}{0.329}
\newcommand{\NSkvROnehSKOneZeroGen}{543}
\newcommand{\NSkvROnehSKTwoZeroFOne}{0.339}
\newcommand{\NSkvROnehSKTwoZeroGen}{542}
\newcommand{\NSkvROnehSKOneZeroVsMTwoZeroFOneLo}{-0.046}
\newcommand{\NSkvROnehSKOneZeroVsMTwoZeroFOneHi}{-0.006}
\newcommand{\NSkvROnehSKOneZeroVsMTwoZeroGenLo}{-0}
\newcommand{\NSkvROnehSKOneZeroVsMTwoZeroGenHi}{+128}
\newcommand{\NSkvROnehSKTwoZeroVsMTwoZeroFOneLo}{-0.034}
\newcommand{\NSkvROnehSKTwoZeroVsMTwoZeroFOneHi}{+0.003}
\newcommand{\NSkvROnehSKTwoZeroVsMTwoZeroGenLo}{+4}
\newcommand{\NSkvROnehSKTwoZeroVsMTwoZeroGenHi}{+118}
\newcommand{\NSkvNrSKOneZeroCon}{0.323}
\newcommand{\NSkvNrSKOneZeroGen}{1728}
\newcommand{\NSkvNrSKTwoZeroCon}{0.292}
\newcommand{\NSkvNrSKTwoZeroGen}{1646}
\newcommand{\NSkvNrSKOneZeroVsMTwoZeroConLo}{-0.005}
\newcommand{\NSkvNrSKOneZeroVsMTwoZeroConHi}{+0.090}
\newcommand{\NSkvNrSKOneZeroVsMTwoZeroGenLo}{+18}
\newcommand{\NSkvNrSKOneZeroVsMTwoZeroGenHi}{+495}
\newcommand{\NSkvNrSKTwoZeroVsMTwoZeroConLo}{-0.025}
\newcommand{\NSkvNrSKTwoZeroVsMTwoZeroConHi}{+0.050}
\newcommand{\NSkvNrSKTwoZeroVsMTwoZeroGenLo}{-69}
\newcommand{\NSkvNrSKTwoZeroVsMTwoZeroGenHi}{+393}
\newcommand{\NDarQEightrDAThreeTwoCon}{0.275}
\newcommand{\NDarQEightrDAThreeTwoGen}{1499}
\newcommand{\NDarQEightrDAPTwoZeroCon}{0.265}
\newcommand{\NDarQEightrDAPTwoZeroGen}{1567}
\newcommand{\NDarQEightrDAThreeTwoVsMTwoZeroConLo}{-0.372}
\newcommand{\NDarQEightrDAThreeTwoVsMTwoZeroConHi}{-0.185}
\newcommand{\NDarQEightrDAThreeTwoVsMTwoZeroGenLo}{+604}
\newcommand{\NDarQEightrDAThreeTwoVsMTwoZeroGenHi}{+1032}
\newcommand{\NDarQEightrDAPTwoZeroVsMTwoZeroConLo}{-0.388}
\newcommand{\NDarQEightrDAPTwoZeroVsMTwoZeroConHi}{-0.198}
\newcommand{\NDarQEightrDAPTwoZeroVsMTwoZeroGenLo}{+656}
\newcommand{\NDarQEightrDAPTwoZeroVsMTwoZeroGenHi}{+1109}
\newcommand{\NDarQOneFourrDAThreeTwoCon}{0.362}
\newcommand{\NDarQOneFourrDAThreeTwoGen}{1103}
\newcommand{\NDarQOneFourrDAPTwoZeroCon}{0.312}
\newcommand{\NDarQOneFourrDAPTwoZeroGen}{1142}
\newcommand{\NDarQOneFourrDAThreeTwoVsMTwoZeroConLo}{-0.347}
\newcommand{\NDarQOneFourrDAThreeTwoVsMTwoZeroConHi}{-0.163}
\newcommand{\NDarQOneFourrDAThreeTwoVsMTwoZeroGenLo}{+479}
\newcommand{\NDarQOneFourrDAThreeTwoVsMTwoZeroGenHi}{+819}
\newcommand{\NDarQOneFourrDAPTwoZeroVsMTwoZeroConLo}{-0.405}
\newcommand{\NDarQOneFourrDAPTwoZeroVsMTwoZeroConHi}{-0.212}
\newcommand{\NDarQOneFourrDAPTwoZeroVsMTwoZeroGenLo}{+499}
\newcommand{\NDarQOneFourrDAPTwoZeroVsMTwoZeroGenHi}{+886}
\newcommand{\NDarROnerDAThreeTwoCon}{0.287}
\newcommand{\NDarROnerDAThreeTwoGen}{1559}
\newcommand{\NDarROnerDAPTwoZeroCon}{0.287}
\newcommand{\NDarROnerDAPTwoZeroGen}{1620}
\newcommand{\NDarROnerDAThreeTwoVsMTwoZeroConLo}{-0.330}
\newcommand{\NDarROnerDAThreeTwoVsMTwoZeroConHi}{-0.168}
\newcommand{\NDarROnerDAThreeTwoVsMTwoZeroGenLo}{+466}
\newcommand{\NDarROnerDAThreeTwoVsMTwoZeroGenHi}{+1042}
\newcommand{\NDarROnerDAPTwoZeroVsMTwoZeroConLo}{-0.325}
\newcommand{\NDarROnerDAPTwoZeroVsMTwoZeroConHi}{-0.170}
\newcommand{\NDarROnerDAPTwoZeroVsMTwoZeroGenLo}{+491}
\newcommand{\NDarROnerDAPTwoZeroVsMTwoZeroGenHi}{+1133}
\newcommand{\NDarQEightwDAThreeTwoFOne}{0.431}
\newcommand{\NDarQEightwDAThreeTwoGen}{1867}
\newcommand{\NDarQEightwDAPTwoZeroFOne}{0.644}
\newcommand{\NDarQEightwDAPTwoZeroGen}{853}
\newcommand{\NDarQEightwDAThreeTwoVsMTwoZeroFOneLo}{-0.350}
\newcommand{\NDarQEightwDAThreeTwoVsMTwoZeroFOneHi}{-0.261}
\newcommand{\NDarQEightwDAThreeTwoVsMTwoZeroGenLo}{+1187}
\newcommand{\NDarQEightwDAThreeTwoVsMTwoZeroGenHi}{+1496}
\newcommand{\NDarQEightwDAPTwoZeroVsMTwoZeroFOneLo}{-0.121}
\newcommand{\NDarQEightwDAPTwoZeroVsMTwoZeroFOneHi}{-0.061}
\newcommand{\NDarQEightwDAPTwoZeroVsMTwoZeroGenLo}{+221}
\newcommand{\NDarQEightwDAPTwoZeroVsMTwoZeroGenHi}{+416}
\newcommand{\NDarQOneFourwDAThreeTwoFOne}{0.438}
\newcommand{\NDarQOneFourwDAThreeTwoGen}{1335}
\newcommand{\NDarQOneFourwDAPTwoZeroFOne}{0.636}
\newcommand{\NDarQOneFourwDAPTwoZeroGen}{637}
\newcommand{\NDarQOneFourwDAThreeTwoVsMTwoZeroFOneLo}{-0.335}
\newcommand{\NDarQOneFourwDAThreeTwoVsMTwoZeroFOneHi}{-0.245}
\newcommand{\NDarQOneFourwDAThreeTwoVsMTwoZeroGenLo}{+826}
\newcommand{\NDarQOneFourwDAThreeTwoVsMTwoZeroGenHi}{+1052}
\newcommand{\NDarQOneFourwDAPTwoZeroVsMTwoZeroFOneLo}{-0.127}
\newcommand{\NDarQOneFourwDAPTwoZeroVsMTwoZeroFOneHi}{-0.059}
\newcommand{\NDarQOneFourwDAPTwoZeroVsMTwoZeroGenLo}{+171}
\newcommand{\NDarQOneFourwDAPTwoZeroVsMTwoZeroGenHi}{+317}
\newcommand{\NDarNwDAThreeTwoBox}{0.297}
\newcommand{\NDarNwDAThreeTwoGen}{2180}
\newcommand{\NDarNwDAPTwoZeroBox}{0.437}
\newcommand{\NDarNwDAPTwoZeroGen}{1513}
\newcommand{\NDarNwDAThreeTwoVsMTwoZeroBoxLo}{-0.343}
\newcommand{\NDarNwDAThreeTwoVsMTwoZeroBoxHi}{-0.262}
\newcommand{\NDarNwDAThreeTwoVsMTwoZeroGenLo}{+1007}
\newcommand{\NDarNwDAThreeTwoVsMTwoZeroGenHi}{+1399}
\newcommand{\NDarNwDAPTwoZeroVsMTwoZeroBoxLo}{-0.198}
\newcommand{\NDarNwDAPTwoZeroVsMTwoZeroBoxHi}{-0.127}
\newcommand{\NDarNwDAPTwoZeroVsMTwoZeroGenLo}{+392}
\newcommand{\NDarNwDAPTwoZeroVsMTwoZeroGenHi}{+682}
\newcommand{\NDarQEighthDAThreeTwoFOne}{0.510}
\newcommand{\NDarQEighthDAThreeTwoGen}{1435}
\newcommand{\NDarQEighthDAPTwoZeroFOne}{0.676}
\newcommand{\NDarQEighthDAPTwoZeroGen}{852}
\newcommand{\NDarQEighthDAThreeTwoVsMTwoZeroFOneLo}{-0.256}
\newcommand{\NDarQEighthDAThreeTwoVsMTwoZeroFOneHi}{-0.174}
\newcommand{\NDarQEighthDAThreeTwoVsMTwoZeroGenLo}{+700}
\newcommand{\NDarQEighthDAThreeTwoVsMTwoZeroGenHi}{+943}
\newcommand{\NDarQEighthDAPTwoZeroVsMTwoZeroFOneLo}{-0.077}
\newcommand{\NDarQEighthDAPTwoZeroVsMTwoZeroFOneHi}{-0.026}
\newcommand{\NDarQEighthDAPTwoZeroVsMTwoZeroGenLo}{+161}
\newcommand{\NDarQEighthDAPTwoZeroVsMTwoZeroGenHi}{+320}
\newcommand{\NDarQOneFourhDAThreeTwoFOne}{0.524}
\newcommand{\NDarQOneFourhDAThreeTwoGen}{987}
\newcommand{\NDarQOneFourhDAPTwoZeroFOne}{0.681}
\newcommand{\NDarQOneFourhDAPTwoZeroGen}{627}
\newcommand{\NDarQOneFourhDAThreeTwoVsMTwoZeroFOneLo}{-0.253}
\newcommand{\NDarQOneFourhDAThreeTwoVsMTwoZeroFOneHi}{-0.170}
\newcommand{\NDarQOneFourhDAThreeTwoVsMTwoZeroGenLo}{+413}
\newcommand{\NDarQOneFourhDAThreeTwoVsMTwoZeroGenHi}{+595}
\newcommand{\NDarQOneFourhDAPTwoZeroVsMTwoZeroFOneLo}{-0.076}
\newcommand{\NDarQOneFourhDAPTwoZeroVsMTwoZeroFOneHi}{-0.032}
\newcommand{\NDarQOneFourhDAPTwoZeroVsMTwoZeroGenLo}{+87}
\newcommand{\NDarQOneFourhDAPTwoZeroVsMTwoZeroGenHi}{+200}
\newcommand{\NDarNhDAThreeTwoBox}{0.312}
\newcommand{\NDarNhDAThreeTwoGen}{1730}
\newcommand{\NDarNhDAPTwoZeroBox}{0.397}
\newcommand{\NDarNhDAPTwoZeroGen}{1495}
\newcommand{\NDarNhDAThreeTwoVsMTwoZeroBoxLo}{-0.290}
\newcommand{\NDarNhDAThreeTwoVsMTwoZeroBoxHi}{-0.215}
\newcommand{\NDarNhDAThreeTwoVsMTwoZeroGenLo}{+599}
\newcommand{\NDarNhDAThreeTwoVsMTwoZeroGenHi}{+916}
\newcommand{\NDarNhDAPTwoZeroVsMTwoZeroBoxLo}{-0.200}
\newcommand{\NDarNhDAPTwoZeroVsMTwoZeroBoxHi}{-0.133}
\newcommand{\NDarNhDAPTwoZeroVsMTwoZeroGenLo}{+383}
\newcommand{\NDarNhDAPTwoZeroVsMTwoZeroGenHi}{+665}
\newcommand{\NDarROnewDAThreeTwoFOne}{0.225}
\newcommand{\NDarROnewDAThreeTwoGen}{1055}
\newcommand{\NDarROnewDAPTwoZeroFOne}{0.350}
\newcommand{\NDarROnewDAPTwoZeroGen}{582}
\newcommand{\NDarROnewDAThreeTwoVsMTwoZeroFOneLo}{-0.216}
\newcommand{\NDarROnewDAThreeTwoVsMTwoZeroFOneHi}{-0.156}
\newcommand{\NDarROnewDAThreeTwoVsMTwoZeroGenLo}{+515}
\newcommand{\NDarROnewDAThreeTwoVsMTwoZeroGenHi}{+738}
\newcommand{\NDarROnewDAPTwoZeroVsMTwoZeroFOneLo}{-0.086}
\newcommand{\NDarROnewDAPTwoZeroVsMTwoZeroFOneHi}{-0.035}
\newcommand{\NDarROnewDAPTwoZeroVsMTwoZeroGenLo}{+80}
\newcommand{\NDarROnewDAPTwoZeroVsMTwoZeroGenHi}{+225}
\newcommand{\NDarROnehDAThreeTwoFOne}{0.240}
\newcommand{\NDarROnehDAThreeTwoGen}{862}
\newcommand{\NDarROnehDAPTwoZeroFOne}{0.315}
\newcommand{\NDarROnehDAPTwoZeroGen}{587}
\newcommand{\NDarROnehDAThreeTwoVsMTwoZeroFOneLo}{-0.144}
\newcommand{\NDarROnehDAThreeTwoVsMTwoZeroFOneHi}{-0.088}
\newcommand{\NDarROnehDAThreeTwoVsMTwoZeroGenLo}{+277}
\newcommand{\NDarROnehDAThreeTwoVsMTwoZeroGenHi}{+486}
\newcommand{\NDarROnehDAPTwoZeroVsMTwoZeroFOneLo}{-0.062}
\newcommand{\NDarROnehDAPTwoZeroVsMTwoZeroFOneHi}{-0.018}
\newcommand{\NDarROnehDAPTwoZeroVsMTwoZeroGenLo}{+44}
\newcommand{\NDarROnehDAPTwoZeroVsMTwoZeroGenHi}{+168}
\newcommand{\NDarNrDAThreeTwoCon}{0.250}
\newcommand{\NDarNrDAThreeTwoGen}{1172}
\newcommand{\NDarNrDAPTwoZeroCon}{0.217}
\newcommand{\NDarNrDAPTwoZeroGen}{1113}
\newcommand{\NDarNrDAThreeTwoVsMTwoZeroConLo}{-0.087}
\newcommand{\NDarNrDAThreeTwoVsMTwoZeroConHi}{+0.030}
\newcommand{\NDarNrDAThreeTwoVsMTwoZeroGenLo}{-572}
\newcommand{\NDarNrDAThreeTwoVsMTwoZeroGenHi}{-61}
\newcommand{\NDarNrDAPTwoZeroVsMTwoZeroConLo}{-0.122}
\newcommand{\NDarNrDAPTwoZeroVsMTwoZeroConHi}{+0.003}
\newcommand{\NDarNrDAPTwoZeroVsMTwoZeroGenLo}{-660}
\newcommand{\NDarNrDAPTwoZeroVsMTwoZeroGenHi}{-103}
\newcommand{\NETwoeQEightrQZFiveZeroRatio}{1.32}
\newcommand{\NETwoeQEightrMTwoZeroRatio}{0.76}
\newcommand{\NETwoeQEightrMFourZeroRatio}{0.78}
\newcommand{\NETwoeNwQZTwoZeroRatio}{1.37}
\newcommand{\NETwoepQEightrLTwoRatio}{5.19}
\newcommand{\NETwoepNwLTwoRatio}{2.56}
\newcommand{\NAggMTwoZeroanyVsFAccLossMin}{0.011}
\newcommand{\NAggMTwoZeroanyVsFAccLossMax}{0.075}
\newcommand{\NAggGOneTwoCBVsFAccLossCells}{12}
\newcommand{\NAggGOneTwoCBVsFGenLongerCells}{11}
\newcommand{\NAggGOneTwoFRVsFAccLossCells}{10}
\newcommand{\NAggGOneTwoFRVsFGenLongerCells}{11}
\newcommand{\NAggGOneTwoPKpTwoZeroVsFAccLossCells}{10}
\newcommand{\NAggGOneTwoPKpTwoZeroVsFGenLongerCells}{11}
\newcommand{\NAggGOneTwoMTwoZeroVsFAccLossCells}{3}
\newcommand{\NAggGOneTwoMTwoZeroVsPKpTwoZeroAccGainCells}{6}
\newcommand{\NAggGOneTwoMTwoZeroVsPKpTwoZeroGenShorterCells}{9}
\newcommand{\NAggGOneTwoMTwoZeroVsCBAccGainCells}{12}
\newcommand{\NAggGOneTwoMTwoZeroVsPKpTwoZeroGenShorterMin}{54}
\newcommand{\NAggGOneTwoMTwoZeroVsFGenMedDiffMin}{-36}
\newcommand{\NAggGOneTwoMTwoZeroVsPKpTwoZeroGenShorterMax}{537}
\newcommand{\NAggGOneTwoMTwoZeroVsFGenMedDiffMax}{+41}
\newcommand{\NAggGOneTwoeBOneMTwoZeroFasterCells}{7}
\newcommand{\NAggGOneTwoeBFourMTwoZeroFasterCells}{10}
\newcommand{\NAggFOneTwoPKVsFGenLongerCells}{11}
\newcommand{\NAggFOneTwoPKVsFAccLossCells}{9}
\newcommand{\NAggFOneTwoAKVsFGenLongerCells}{12}
\newcommand{\NAggFOneTwoAKVsFAccLossCells}{12}
\newcommand{\NAggFOneTwoMTwoZeroVsQZTwoZeroAccGainCells}{10}
\newcommand{\NAggFOneTwoMTwoZeroVsQZTwoZeroGenShorterCells}{9}
\newcommand{\NAggFOneTwoQZTwoZeroVsFGenLongerCells}{10}
\newcommand{\NAggFOneTwoQZTwoZeroVsFAccLossCells}{9}
\newcommand{\NAggFOneTwoMTwoZeroVsQZFiveZeroAccGainCells}{2}
\newcommand{\NAggFOneTwoMTwoZeroVsQZFiveZeroAccLossCells}{1}
\newcommand{\NAggFOneTwoQZFiveZeroVsFGenLongerCells}{4}
\newcommand{\NAggFOneTwoQZFiveZeroVsFAccLossCells}{3}
\newcommand{\NAggFOneTwoLTwoVsFGenLongerCells}{11}
\newcommand{\NAggFOneTwoLTwoVsFAccLossCells}{12}
\newcommand{\NAggFOneTwoLThreeTwoVsFGenLongerCells}{11}
\newcommand{\NAggFOneTwoLThreeTwoVsFAccLossCells}{12}
\newcommand{\NAggFOneTwosAKShareMin}{23.2}
\newcommand{\NAggFOneTwosAKShareMax}{29.9}
\newcommand{\NAggFOneTwosLTwoShareMin}{0.3}
\newcommand{\NAggFOneTwosLTwoShareMax}{1.5}
\newcommand{\NAggFOneTwosLThreeTwoShareMin}{4.2}
\newcommand{\NAggFOneTwosLThreeTwoShareMax}{24.3}
\newcommand{\NAggCcMRatioMin}{1.78}
\newcommand{\NAggCcMRatioMax}{2.85}
\newcommand{\NAggCcFRRatioMin}{14.41}
\newcommand{\NAggCcFRRatioMax}{28.98}
\newcommand{\NAggMechAttnFBeforeMin}{0.10}
\newcommand{\NAggMechAttnFBeforeMax}{0.47}
\newcommand{\NAggMechTransferEightKShareMin}{0.67}
\newcommand{\NAggMechTransferEightKIdealMin}{0.04}
\newcommand{\NAggMechTransferEightKShareMax}{0.84}
\newcommand{\NAggMechTransferEightKIdealMax}{0.22}
\newcommand{\NAggMechTransferEightVIdealMin}{0.08}
\newcommand{\NAggMechTransferEightVShareMax}{0.78}
\newcommand{\NAggMechTransferEightVIdealMax}{0.37}
\newcommand{\NAggMechTransferEightVqThreeShareMin}{0.28}
\newcommand{\NAggCapPTwoZeroRatioMin}{1.6}
\newcommand{\NAggCapPFourZeroRatioMin}{1.4}
\newcommand{\NAggCapOneFourPTwoZeroRatioMax}{2.5}
\newcommand{\NAggCapOneFourPFourZeroRatioMax}{1.8}
\newcommand{\NPOneamortAggRulerUsesFMin}{2.6}
\newcommand{\NPOneamortAggRulerUsesFMax}{4.6}
\newcommand{\NPOneamortAggShortUsesFMin}{4.3}
\newcommand{\NPOneamortAggShortUsesFMax}{15.6}
\newcommand{\NPOnethreeExtensionsCells}{24}
\newcommand{\NPOnethreeExtensionsCost}{19}
\newcommand{\NPOnethreeExtensionsGainWhereCost}{12}
\newcommand{\NPOnelifeMFourRatio}{0.762}
\newcommand{\NPOnelifeMFourpFourZeroRatio}{0.839}
\newcommand{\NPOnelifeNFourRatio}{0.953}
\newcommand{\NAggSkvSKOneZeroVsMTwoZeroAccNegCells}{10}
\newcommand{\NAggSkvSKOneZeroVsMTwoZeroGenPosCells}{10}
\newcommand{\NAggSkvSKOneZeroVsFAccNegCells}{9}
\newcommand{\NAggSkvSKOneZeroVsFGenPosCells}{11}
\newcommand{\NAggSkvSKTwoZeroVsMTwoZeroAccLevelCells}{10}
\newcommand{\NAggSkvSKTwoZeroVsMTwoZeroGenNegCells}{1}
\newcommand{\NAggSkvSKTwoZeroVsMTwoZeroGenPosCells}{5}
\newcommand{\NAggDarDAThreeTwoVsFAccNegCells}{11}
\newcommand{\NAggDarDAThreeTwoVsFGenPosCells}{11}
\newcommand{\NAggDarDAPTwoZeroVsFAccNegCells}{11}
\newcommand{\NAggDarDAPTwoZeroVsFGenPosCells}{11}
\newcommand{\NAggDarDAPTwoZeroVsMTwoZeroGenPosCells}{11}
\newcommand{\NAggRadixFrxOverFMin}{0.96}
\newcommand{\NAggRadixFrxOverFMax}{1.03}
\newcommand{\NSfQEightrPKFourLatLowTTwoZero}{2.09}
\newcommand{\NSfQEightrCBFourLatLowTTwoZero}{1.88}
\newcommand{\NAggSfQEightrPKFourRatioMin}{0.79}
\newcommand{\NAggSfQEightrPKFourRatioMax}{0.93}
\newcommand{\NAggSfQEightrCBFourRatioMin}{0.85}
\newcommand{\NAggSfQEightrCBFourRatioMax}{0.91}
\newcommand{\NAggSfQEightrMFourOverPubMax}{1.97}
\newcommand{\NSfQEightwPKFourLatLowTTwoZero}{0.66}
\newcommand{\NAggSfQEightwMFourOverPubMin}{1.08}
\newcommand{\NSfQOneFourrCBFourLatLowTTwoZero}{0.97}
\newcommand{\NAggSfTokensPubOverFMin}{1.37}
\newcommand{\NAggSfTokensPubOverFMax}{3.59}
\newcommand{\NAggSfLatLowMFourMin}{0.31}
\newcommand{\NAggSfLatLowMFourMax}{0.49}
\newcommand{\NSelQEightrCtlPKPt}{0.131}
\newcommand{\NSelQEightrCtlORPt}{0.239}
\newcommand{\NSelQEightrCtlORChunkHeadsPt}{0.093}
\newcommand{\NSelNwFillNopropOROverPK}{0.31}
\newcommand{\NSelLEightwFillNopropOROverPK}{0.09}
\newcommand{\NAggSelFillRhoRkMin}{-0.84}
\newcommand{\NAggSelFillRhoRkMax}{-0.70}
\newcommand{\NAggPropRedMVsPKMin}{28}
\newcommand{\NAggPropRedMVsPKMax}{42}
\newcommand{\NAggPropFillerDnllCutMin}{42}
\newcommand{\NAggPropFillerDnllCutMax}{58}
\newcommand{\NAggPropRedOKVVsPKMin}{74}
\newcommand{\NAggPropRedOKVVsPKMax}{82}
\newcommand{\NAggPropRedTUVsPKMin}{30}
\newcommand{\NAggPropRedTUVsPKMax}{58}
\newcommand{\NAggLinkRhoGenMin}{+0.68}
\newcommand{\NAggLinkRhoGenMax}{+0.87}
\newcommand{\NAggLinkRhoAccMin}{-0.77}
\newcommand{\NAggLinkRhoAccMax}{-0.67}
\newcommand{\NAggRowsRkFillRulerMin}{0.485}
\newcommand{\NAggRowsRkFillRulerMax}{0.526}
\newcommand{\NAggRowsRkFillWikiMin}{0.111}
\newcommand{\NAggRowsRkFillWikiMax}{0.197}
\newcommand{\NAggTbMTwoZeroVsFTaskNolossCells}{15}
\newcommand{\NAggTbMFourZeroVsFTaskNolossCells}{16}
\newcommand{\NNtQEightwFFOne}{0.454}
\newcommand{\NNtQEightwNRVsFFOnePt}{-0.232}
\newcommand{\NNtQEighthFFOne}{0.623}
\newcommand{\NNtQEighthNRVsFFOnePt}{-0.323}
\newcommand{\NAggNtStrongRecoversMin}{81}
\newcommand{\NAggNtStrongRecoversMax}{101}
\newcommand{\NAggNtRepVsFGenMax}{+2.1}
\newcommand{\NLenQEightrFCheckSharePct}{25}
\newcommand{\NLenLEightrFCheckSharePct}{30}
\newcommand{\NAggLenMShareChkRulerMin}{53}
\newcommand{\NAggLenMShareChkRulerMax}{60}
\newcommand{\NAggLenRepCpreWaitRatioMin}{1.7}
\newcommand{\NAggLenRepCpreWaitRatioMax}{6.4}
\newcommand{\NAggLenRepCpreSoRatioMin}{0.68}
\newcommand{\NAggLenRepCpreSoRatioMax}{1.17}
\newcommand{\NAggLenRepMWaitRatioMin}{1.06}
\newcommand{\NAggLenRepMWaitRatioMax}{1.15}
\newcommand{\NAggLenRhoWaitRatioExtraWaitMin}{0.67}
\newcommand{\NAggLenRhoWaitRatioExtraWaitMax}{0.88}

%% file: sections/01-introduction.tex
\section{Introduction}
\label{sec:intro}

\begin{figure}[t]
\centering
\includegraphics[width=\columnwidth]{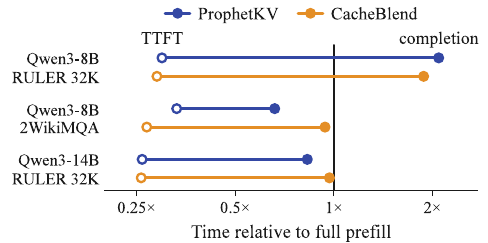}
\caption{Time to first token (TTFT, hollow) and mean completion time (filled) of requests served at low load in SGLang on an H20 GPU, as multiples of full prefill's; both repairs at 20\% recompute.}
\label{fig:served-ttft-e2e}
\end{figure}

\begin{figure*}[t]
\centering
\includegraphics[width=\textwidth]{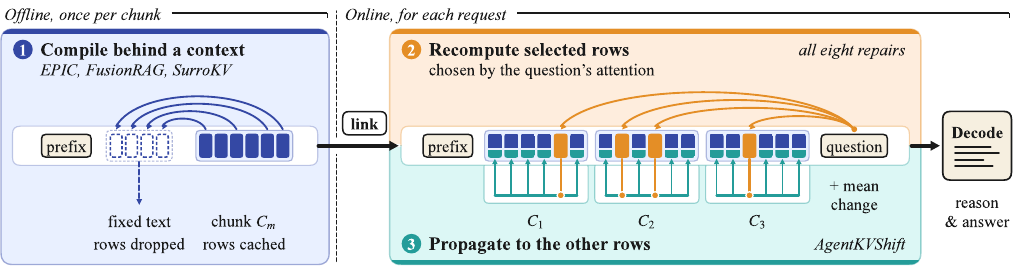}
\caption{Where a PIC repair acts: offline, (1) compiling each chunk behind a context whose rows are then dropped; online, (2) recomputing selected chunk rows and (3) propagating a shift to the other rows. In italics, the revisited repairs that act at each step; the drawing shows the recipe's choice. A bar is one token's KV row.}
\label{fig:framework}
\end{figure*}

Prefix caching \citep{zheng-sglang-2023} reuses a key--value (KV) cache only for an identical leading prefix, since each token's keys and values depend on all tokens before it.
Long prompts are often assembled from chunks that recur across requests, such as retrieved passages, code files, or worked examples.
Because recurring chunks arrive in changing combinations and orders, prefix caching rarely hits.

Building on modular attention reuse \citep{gim-prompt-2023}, position-independent caching (PIC) compiles each chunk's KV cache once offline, splices cached chunks into later prompts at any position, and recomputes a small share of tokens online to repair the cross-chunk attention lost by compiling chunks alone \citep{yao-cacheblend-2024,hu-epic-2024}.
Published repairs have been evaluated almost entirely on short answers with reasoning disabled, where they report a much earlier time to first token (TTFT) at close to the accuracy of full prefill.

On reasoning models, which decode hundreds to thousands of chain-of-thought tokens before answering, an early first token no longer tells how much time PIC saves.
We therefore revisit eight PIC repairs \citep{yao-cacheblend-2024,hu-epic-2024,wang-prophetkv-2026,pandey-agentkvshift-2026,wang-prefix-2026,yan-qcfuse-2026,anon-surrokv-2026,wang-daric-2026} in a unified reimplementation on four reasoning models and three long-context retrieval-QA benchmarks, reporting generated tokens alongside accuracy.
In a short-answer control with reasoning disabled, the stronger repairs recover \NAggNtStrongRecoversMin{}--\NAggNtStrongRecoversMax{}\% of the accuracy that compiling chunks alone loses.
With reasoning enabled, however, the published repairs at 20\% recompute or less lose accuracy against full prefill in \NAggFOneTwoPKVsFAccLossCells{}--\NAggGOneTwoCBVsFAccLossCells{} of the 12 model--benchmark cells and generate more tokens in \NAggFOneTwoQZTwoZeroVsFGenLongerCells{} or \NAggFOneTwoPKVsFGenLongerCells{}.
Those extra tokens can outweigh the prefill saving end to end.
In serving with SGLang on an H20 GPU at low load, two of these repairs still reach the first token several times sooner than full prefill, yet on Qwen3-8B RULER 32K their requests even finish later than full prefill's (Figure~\ref{fig:served-ttft-e2e}).

Raising the recompute share reduces both the accuracy loss and the extra tokens, as QCFuse does at its default 50\% \citep{yan-qcfuse-2026}, yet recomputing half of the context surrenders much of the prefill saving.
To curb both costs at a small recompute share, we trace where a repaired cache's damage to the reasoning comes from, measuring that damage by replaying full prefill's output over the cache.
The damage tilts the model toward checking what it has read, and these checks are where much of the extra reasoning goes.
The damage starts with compiling each chunk alone, which displaces the chunk's keys and values.
We find that this displacement is largely the same across the real prompts containing the chunk, so compiling each chunk once behind one fixed text can compensate much of it offline.
Once the prompt is known, the rest is left to online recomputation.
Comparing selection rules, we find that the online step should favor the rows holding most of the remaining key error that the reasoning reads.
The recomputed rows also reveal how their chunk shifted, so propagating their mean change to the rest of the chunk further reduces the damage that recomputation leaves.
A repair thus takes three steps, each acting on what the one before leaves.

We assemble these three steps into a training-free recipe (Figure~\ref{fig:framework}).
Offline, it compiles each chunk once behind a fixed, generic Wikipedia text whose rows are then discarded \citep{hu-epic-2024,anon-surrokv-2026}.
Online, it recomputes the 20\% of chunk rows that the question attends to most \citep{wang-prophetkv-2026} and propagates their per-chunk mean change to the remaining rows \citep{pandey-agentkvshift-2026}.
Across the 12 cells, it scores higher than each published repair at the same 20\% recompute in \NAggFOneTwoMTwoZeroVsQZTwoZeroAccGainCells{}--\NAggGOneTwoMTwoZeroVsCBAccGainCells{} cells and generates fewer tokens in \NAggFOneTwoMTwoZeroVsQZTwoZeroGenShorterCells{}--\NAggDarDAPTwoZeroVsMTwoZeroGenPosCells{}.
On larger dense and mixture-of-experts (MoE) models up to 32B, it also scores above CacheBlend \citep{yao-cacheblend-2024}, the one published repair we ran there at the same share, in all six cells.
On summarization, code, and math, it has no resolved task-score loss against full prefill in \NAggTbMTwoZeroVsFTaskNolossCells{} of the 16 cells.
In serving with SGLang on an H20 GPU, its peak completion rate over the loads tested is \NAggCapPTwoZeroRatioMin{}--\NAggCapOneFourPTwoZeroRatioMax{}$\times$ full prefill's.

This paper makes three contributions:
\begin{enumerate}
  \item \textbf{A unified re-evaluation that reveals the hidden cost of PIC on reasoning models.}
  We reimplement eight PIC repairs in one engine and find that the published ones generate more tokens and lose accuracy.
  Short-answer evaluations cannot show the longer reasoning.
  \item \textbf{A diagnosis of where the damage to the reasoning comes from.}
  We trace the damage to compiling each chunk alone, show how it lengthens the reasoning, and conclude that a repair needs three steps, each acting on what the one before leaves.
  \item \textbf{An assembled, training-free recipe.}
  We assemble the three steps into a recipe from existing mechanisms.
  It scores higher than each published repair at the same 20\% recompute in \NAggFOneTwoMTwoZeroVsQZTwoZeroAccGainCells{}--\NAggGOneTwoMTwoZeroVsCBAccGainCells{} of the 12 cells and generates fewer tokens in \NAggFOneTwoMTwoZeroVsQZTwoZeroGenShorterCells{}--\NAggDarDAPTwoZeroVsMTwoZeroGenPosCells{}.
  In serving on an H20 GPU, its peak completion rate over the loads tested is \NAggCapPTwoZeroRatioMin{}--\NAggCapOneFourPTwoZeroRatioMax{}$\times$ full prefill's.
\end{enumerate}

%% file: sections/02-background.tex
\section{Background and Related Work}
\label{sec:background}

\subsection{Position-Independent Caching}

Consider a prompt $X = [P, C_1, \dots, C_M, Q]$ composed of an optional shared prefix $P$, $M$ reusable chunks $C_1, \dots, C_M$, and a request-specific suffix $Q$ containing the user query.
In this notation, prefix caching reuses the cache of a chunk $C_m$ only when $P, C_1, \dots, C_{m-1}$ repeat exactly.
PIC removes this constraint through two stages formalized by EPIC \citep{hu-epic-2024}.
In the offline compile stage, each chunk $C_m$ is prefilled once, either alone or behind a compile context, and its cache $(\tilde{K}_m, \tilde{V}_m)$ is stored.
In the online link stage, an arriving request loads these caches, moves each $\tilde{K}_m$ to the chunk's position in the prompt by a rotary position embedding (RoPE) rotation \citep{su-roformer-2021}, and places the chunks between $P$ and $Q$.

Even after RoPE restores contiguous global positions, a chunk that follows other text in the prompt but was compiled alone has its keys and values displaced relative to full prefill, because it never attended to that text.
Its opening tokens also become spurious attention sinks \citep{xiao-efficient-2023}, as EPIC observes.
Online repair mitigates this mismatch by recomputing a selected subset $S$ of chunk tokens over the assembled prompt (recompute share $\rho = |S| / \sum_{m=1}^{M} |C_m|$).

\subsection{Published PIC Repairs}

Training-free PIC repairs differ in how they select $S$ and whether they also act outside $S$.
Query-agnostic selectors choose $S$ without inspecting $Q$.
EPIC targets boundary sinks by recomputing the first $k$ tokens of each non-initial chunk or by prepending four dummy tokens offline and dropping their KV states.
DARIC \citep{wang-daric-2026} generalizes this static rule by assigning progressively more initial tokens to later chunks and recomputing a trailing prompt window.
CacheBlend \citep{yao-cacheblend-2024} instead selects the tokens whose KV deviates most from their cached states, narrowing the candidates layer by layer from the first, and RedKnot \citep{liu-redknot-2026} recomputes whole KV heads, those that offline profiling marks as global.
Because query-agnostic rules can spend budget on chunks unrelated to the request, query-aware selectors use $Q$ to target relevant evidence.
ProphetKV \citep{wang-prophetkv-2026} runs the user query over the spliced cache and recomputes the chunk tokens receiving the highest layer-averaged query attention.
To avoid loading every layer before selection, QCFuse \citep{yan-qcfuse-2026} probes the query against preselected chunk anchors and scores context keys at calibrated critical layers.
RelaxKV \citep{qi-relaxkv-2026} recomputes query-chosen tokens against only a query-relevant part of the context.

Some repairs also act outside $S$, either at compile time or on unselected rows online (Figure~\ref{fig:framework}).
Offline, FusionRAG \citep{wang-prefix-2026} prefills each chunk behind retrieved corpus neighbors whose KV states are discarded, then applies last-layer query-guided selection online.
Concurrent SurroKV \citep{anon-surrokv-2026} replaces corpus neighbors with one fixed Wikipedia passage whose rows are dropped and pairs it with ProphetKV's selection.
Transition Overlap \citep{xu-transition-2026} and QV-PIC \citep{liu-qvpic-2026} drop a short compile context the same way, and without any recomputation even random chunks as a compile context beat none \citep{cestola-evaluating-2026}.
Online, AgentKVShift \citep{pandey-agentkvshift-2026} selects per-chunk key and value probes by first-layer divergence, recomputes their union, and propagates the probes' mean key and value residuals to the chunk's unselected rows.
None of these repairs acts at all three steps of Figure~\ref{fig:framework}, and none of their papers follows the damage to the reasoning through the steps; Section~\ref{sec:diagnosis} does, and the recipe of Section~\ref{sec:recipe} acts at all three.

Outside training-free repair, trained PIC methods adapt the model to standalone caches by fine-tuning its backbone or added link tokens \citep{lu-turborag-2024,ma-blockattention-2024,yang-kvlink-2025}, or train adapter tokens, a cache writer or a recompute selector for it \citep{chen-kvpacket-2026,xie-sempic-2026,pan-cacheweave-2026}; APE \citep{yang-ape-2025} instead adjusts attention over contexts encoded in parallel.
Orthogonal KV quantization and compression reduce cache storage and transfer volume \citep{liu-kivi-2024,liu-cachegen-2023}, and AgentKVShift evaluates PIC repairs over low-bit KV stores.

\subsection{What Short-Answer Evaluations Missed}

Prior PIC evaluations, from those of CacheBlend, EPIC, FusionRAG and SurroKV to two studies that re-implement several repairs in one framework \citep{cestola-experimental-2026,cestola-evaluating-2026}, have scored short answers in QA and brief summarization, almost always with non-reasoning models or with thinking disabled.
When answers run to a few tokens, prefill can take nearly all of a request's time, as FusionRAG observes, and a faithful repair can take longer than full prefill to process the prompt \citep{zhang-safekv-2026}.

Only scattered reports touch on output length or reasoning, without measuring their end-to-end cost.
EPIC notes occasional overly long outputs under its own repair, QCFuse's repository reports slightly more decoded tokens with thinking enabled at 20\% recompute, and ProphetKV evaluates a thinking model on task scores and TTFT without comparing generated token counts across methods.
Yet when a reasoning model decodes hundreds to thousands of chain-of-thought tokens before answering, decoding time can rival or exceed prefill time.
A cache error that lengthens the reasoning then adds decoding steps that offset the prefill time PIC saves, so TTFT and short-answer accuracy cannot tell whether a published repair still pays off once reasoning runs to the end.

%% file: sections/03-revisiting.tex
\section{Revisiting Published PIC Repairs under Reasoning}
\label{sec:revisiting}

To test whether existing PIC repairs still pay off once reasoning runs to the end, we reimplement eight training-free methods from Section~\ref{sec:background} in one engine.
Sharing one engine isolates their compile and repair rules from framework and kernel differences.

\subsection{Unified Reimplementation and Evaluation Setup}
\label{sec:setup}

Appendix~\ref{app:ports} lists the eight BF16 repairs with their mechanisms and recompute shares (Table~\ref{tab:repairs}) and each port's implementation details.

We evaluate 12 retrieval-QA cells, crossing four reasoning models with three multi-hop QA tasks.
The models are Qwen3-8B and Qwen3-14B \citep{yang-qwen3-2025}, DeepSeek-R1-Distill-Llama-8B \citep{deepseekai-deepseekr1-2025} and Nemotron-Nano-8B \citep{bercovich-llamanemotron-2025}; the tasks are RULER's multi-hop QA at 32K tokens \citep{hsieh-ruler-2024} and LongBench-E's 2WikiMQA and HotpotQA \citep{bai-longbench-2023,ho-constructing-2020,yang-hotpotqa-2018}, with \NFcQEightrFNq{}, \NFcQEightwFNq{} and \NFcQEighthFNq{} questions.
Each dataset's passages form the chunks; without an offline compile context, chunks compile alone on Qwen3 and after the shared prefix $P$ on the Llama models.
Reasoning runs enable thinking ($T{=}0.6$, top-$p{=}0.95$, four seeds), with generation capped at 6,144 tokens on RULER 32K and 8,192 on LongBench-E. %

We report mean generated tokens (reasoning plus answer) and accuracy: normalized containment (\emph{contains}) on RULER 32K and post-thinking token F1 on LongBench-E (boxed F1 for Nemotron-Nano-8B).
Repairs are paired against full prefill with 95\% bootstrap intervals over questions; a difference is \emph{resolved} when its interval excludes zero and \emph{level} otherwise.

\subsection{Short-Answer Control with Thinking Disabled}
\label{sec:nothink}

Before turning to reasoning, we check our ports in the short-answer regime on Qwen3-8B over the same 2WikiMQA and HotpotQA prompts, disabling thinking, decoding greedily up to 32 tokens, and scoring the full output by F1.
Full prefill scores \NNtQEightwFFOne{} and \NNtQEighthFFOne{} F1, while compiling alone shifts F1 by \NNtQEightwNRVsFFOnePt{} and \NNtQEighthNRVsFFOnePt{}.

\begin{figure*}[t]
\centering
\includegraphics[width=\textwidth]{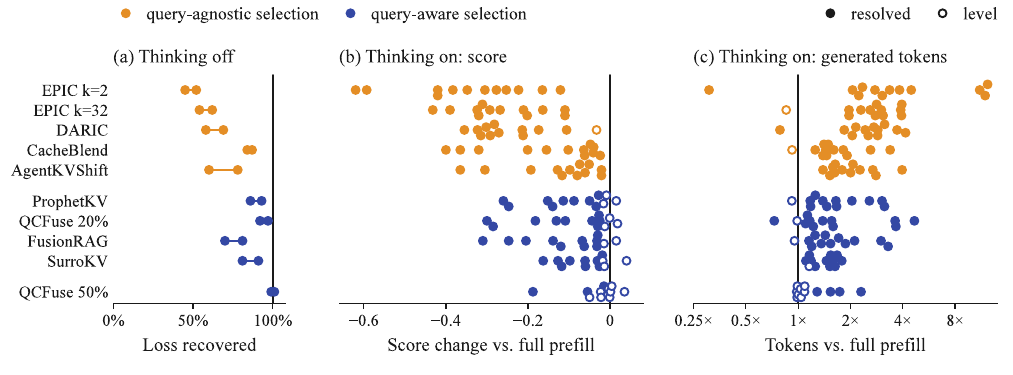}
\caption{The eight repairs against full prefill. (a)~Thinking off (Qwen3-8B, 2WikiMQA-E and HotpotQA-E): the share of the loss from compiling chunks alone that each repair recovers. (b,~c)~Thinking on, 12 cells: score change and generated tokens, one mark per cell; filled: resolved, hollow: level.}
\label{fig:revisiting-grid}
\end{figure*}

Here our ports repair most of that loss (Figure~\ref{fig:revisiting-grid}a).
The stronger selective repairs (CacheBlend, ProphetKV, QCFuse at 20\%/50\%, SurroKV at 10\%) recover \NAggNtStrongRecoversMin{}--\NAggNtStrongRecoversMax{}\% of the gap, and across all repairs, outputs differ by at most \NAggNtRepVsFGenMax{} tokens.

\subsection{Accuracy Loss and Longer Reasoning with Thinking Enabled}
\label{sec:grid-results}

Enabling thinking on the prompts of Section~\ref{sec:nothink} shifts length sharply.
Every repair except QCFuse at 50\% now generates resolvably more tokens than full prefill on both benchmarks, adding \NSkvQEighthSKOneZeroVsFGenPt{} to \NFOneTwoQEightwLTwoVsFGenPt{} tokens per sample.
Across all 12 cells with reasoning run to the end (Figure~\ref{fig:revisiting-grid}b--c), the published repairs at 20\% recompute or less generate more tokens than full prefill in \NAggFOneTwoQZTwoZeroVsFGenLongerCells{} or \NAggFOneTwoPKVsFGenLongerCells{} cells and lose accuracy against it in \NAggFOneTwoPKVsFAccLossCells{}--\NAggGOneTwoCBVsFAccLossCells{}.
Reasoning thus exposes extra decoding that short-answer benchmarks miss.

Raising the recompute share curbs both costs, with QCFuse at its default 50\% cutting accuracy losses to \NAggFOneTwoQZFiveZeroVsFAccLossCells{} cells and extra tokens to \NAggFOneTwoQZFiveZeroVsFGenLongerCells{}.
Recomputing half the context, however, surrenders much of the prefill saving that motivates PIC.

\subsection{Fast to Start, Slow to Finish in Serving}
\label{sec:served-cost}

We next measure in a serving engine whether the extra decoding outweighs the prefill saving.
We serve the 20\% plans of ProphetKV and CacheBlend beside full prefill in SGLang, one H20 GPU per system, on Qwen3-8B RULER 32K, Qwen3-8B 2WikiMQA-E, and Qwen3-14B RULER 32K (setup in Section~\ref{sec:serving-h20}).
At each workload's lowest rate, both plans reach the first token several times sooner than full prefill (Figure~\ref{fig:served-ttft-e2e}), yet generate \NAggSfTokensPubOverFMin{}--\NAggSfTokensPubOverFMax{}$\times$ its tokens per request.
On Qwen3-8B RULER 32K, their mean completion time is \NSfQEightrCBFourLatLowTTwoZero{}--\NSfQEightrPKFourLatLowTTwoZero{}$\times$ full prefill's (Figure~\ref{fig:served-ttft-e2e}), so their requests finish later despite starting sooner, and over the arrival rates tested they reach only \NAggSfQEightrPKFourRatioMin{}--\NAggSfQEightrPKFourRatioMax{}$\times$ and \NAggSfQEightrCBFourRatioMin{}--\NAggSfQEightrCBFourRatioMax{}$\times$ full prefill's peak completion rate, respectively.
On the other two workloads, both plans still finish sooner than full prefill, at \NSfQEightwPKFourLatLowTTwoZero{}--\NSfQOneFourrCBFourLatLowTTwoZero{}$\times$ its mean completion time.

%% file: sections/04-diagnosis.tex
\section{Diagnosing Where the Damage Comes From}
\label{sec:diagnosis}

To curb the accuracy loss and the extra decoding without recomputing half the context, we trace where a repaired cache's damage to the reasoning comes from.
We measure this damage by \emph{replay}, teacher-forcing the first 512 tokens of full prefill's own output over a repaired cache: the extra negative log-likelihood (NLL) per token over full prefill's is the cache's \emph{replay damage}. %
Our three primary cells, each replayed on 32 questions, are Qwen3-8B and R1-Distill-Llama-8B on RULER 32K, whose passages are short, and Nemotron-Nano-8B on 2WikiMQA-E, whose passages are long.
Replay follows the cost of generation: across nine repair variants in each cell, replay damage correlates with the extra tokens a variant generates over full prefill (Spearman $\NAggLinkRhoGenMin{}$ to $\NAggLinkRhoGenMax{}$) and with its accuracy change ($\NAggLinkRhoAccMin{}$ to $\NAggLinkRhoAccMax{}$).

\begin{figure}[t]
\centering
\includegraphics[width=\columnwidth]{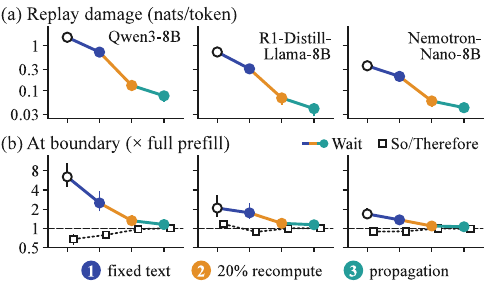}
\caption{(a) Replay damage and (b) the likelihood of ``Wait'' (squares: ``So'' or ``Therefore'') where full prefill's thinking starts a sentence or a line, relative to full prefill's cache, after each step of Figure~\ref{fig:framework}. Hollow circles: chunks compiled after the shared prefix alone.}
\label{fig:diagnosis-steps}
\end{figure}

\begin{figure*}[t]
\centering
\includegraphics[width=\textwidth]{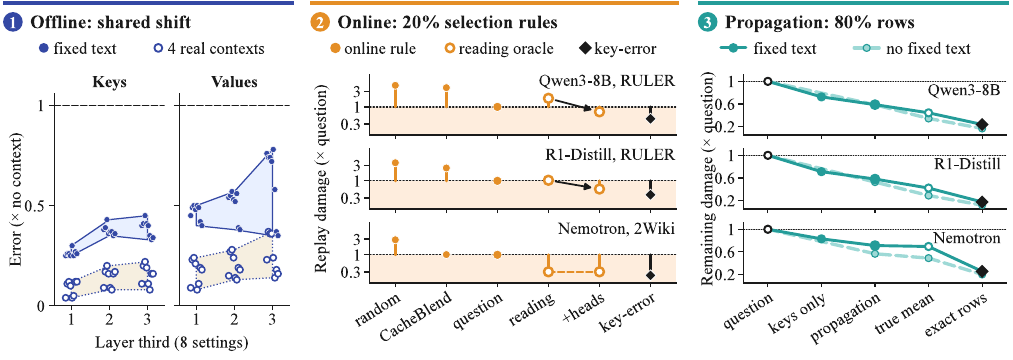}
\caption{What each step acts on. (a) Error a chunk keeps behind the fixed text and as the noise-corrected mean of its caches behind four real contexts (no context${}=1$). (b) Replay damage of six 20\% selection rules (question's attention${}=1$; +heads: the reading oracle with every chunk's head forced in). (c) Damage left after the 20\% recompute, by correction. (b, c) Qwen3-8B and R1-Distill-Llama-8B on RULER 32K, Nemotron-Nano-8B on 2WikiMQA-E.}
\label{fig:mechanism}
\end{figure*}

\subsection{How the Damage Lengthens the Reasoning}
\label{sec:diag-length}

A damaged cache tilts the model toward checking what it has read.
We replay full prefill's thinking to its end, up to 2,048 tokens, and compare the probability each cache gives ``Wait'' wherever that thinking starts a sentence or a line. %
Compiled after the shared prefix alone, the cache makes ``Wait'' \NAggLenRepCpreWaitRatioMin{}--\NAggLenRepCpreWaitRatioMax{} times as likely there as full prefill's cache does, while ``So'' or ``Therefore'', the words that carry the reasoning on, change far less (\NAggLenRepCpreSoRatioMin{}--\NAggLenRepCpreSoRatioMax{} times; Figure~\ref{fig:diagnosis-steps}b).

The tilt carries over to generation as extra checks.
Across the nine variants of each cell, the more a variant raises ``Wait'' in replay, the more extra times it writes ``Wait'' in generation (Spearman \NAggLenRhoWaitRatioExtraWaitMin{} to \NAggLenRhoWaitRatioExtraWaitMax{}).
On RULER 32K, \NAggLenMShareChkRulerMin{}--\NAggLenMShareChkRulerMax{}\% of the extra thinking the recipe still adds lies in paragraphs that open with a check (``Wait'', ``Hmm'', ``let me check''), against \NLenQEightrFCheckSharePct{}--\NLenLEightrFCheckSharePct{}\% of full prefill's own thinking there.

\subsection{Compiling Chunks Behind a Fixed Text}
\label{sec:diag-offline}

The damage begins offline, when a chunk is compiled without the passages that precede it in a prompt.
Under full prefill, in eight model--dataset settings, a chunk's tokens send \NAggMechAttnFBeforeMin{}--\NAggMechAttnFBeforeMax{} of their attention in the middle and upper thirds of the layers to those passages; compiled without them, the chunk directs that attention elsewhere, which displaces its keys and values.
Because the passages before a chunk vary across requests, this displacement might be expected to be specific to each prompt.
It is mostly shared (Figure~\ref{fig:mechanism}a): against a chunk's cache behind an unseen real context, the mean of its caches over all real contexts from other prompts leaves only \NAggMechTransferEightKIdealMin{}--\NAggMechTransferEightKIdealMax{} of the key error and \NAggMechTransferEightVIdealMin{}--\NAggMechTransferEightVIdealMax{} of the value error that compiling with no context leaves, across eight model--dataset settings.

Much of the displacement can therefore be compensated before any request arrives, by compiling each chunk once behind a fixed text whose KV rows are then dropped, as EPIC and SurroKV do; CacheClip \citep{yang-cacheclip-2025} does the same with the system prompt but keeps one copy of it.
By SurroKV's analysis, such a text helps as far as the shift it causes comes close to the mean shift of real contexts.
One Wikipedia text of about 2.2K tokens achieves \NAggMechTransferEightKShareMin{}--\NAggMechTransferEightKShareMax{} of the mean's error reduction in the keys and \NAggMechTransferEightVqThreeShareMin{}--\NAggMechTransferEightVShareMax{} in the values, and before any online recomputation it cuts the replay damage in the three primary cells by \NAggPropFillerDnllCutMin{}--\NAggPropFillerDnllCutMax{}\% against compiling after the shared prefix $P$ alone (Figure~\ref{fig:diagnosis-steps}a). %

\subsection{Choosing Which Rows to Recompute}
\label{sec:diag-selection}

With every chunk behind the fixed text, we compare rules for choosing the chunk rows to recompute online, at the 20\% share used to compare repairs throughout the paper; besides the three primary cells, we run Qwen3-8B and R1-Distill-Llama-8B on 2WikiMQA-E.
ProphetKV motivates selecting rows by the question's attention with the overlap between the rows the question attends to and those that decoding later attends to.
That motivation suggests an obvious upper bound, a reading oracle that recomputes the rows full prefill's own reasoning attends to most.
Yet on RULER 32K it does no better than the question's attention, and on Qwen3-8B it leaves more damage (\NSelQEightrCtlORPt{} against \NSelQEightrCtlPKPt{}); on 2WikiMQA-E it leaves only \NSelLEightwFillNopropOROverPK{}--\NSelNwFillNopropOROverPK{} of the question's damage with each of the three models.

The split follows where the remaining error sits.
We weight each row's key error, layer by layer, by the attention that the queries of full prefill's reasoning give the row over the damaged keys, and call the sum the row's \emph{reading-weighted key error}.
On RULER 32K, about half of this error, \NAggRowsRkFillRulerMin{}--\NAggRowsRkFillRulerMax{}, sits in the first 8 rows of each chunk, its head; on 2WikiMQA-E, only \NAggRowsRkFillWikiMin{}--\NAggRowsRkFillWikiMax{} does.
The question's attention selects nearly all of these heads in every cell, but the reading oracle selects only about half of them on RULER 32K; with every head forced in, it beats the question's choice there (on Qwen3-8B, \NSelQEightrCtlORChunkHeadsPt{} against \NSelQEightrCtlPKPt{}; Figure~\ref{fig:mechanism}b).
Across the selection rules we compare, the more of this error a rule's rows hold, the less damage the rule leaves (mean Spearman's $\rho$ within questions, $\NAggSelFillRhoRkMin{}$ to $\NAggSelFillRhoRkMax{}$ across the five cells), and an oracle that recomputes the rows holding the most of it leaves the least damage of all the rules in every cell, or is level with the least.
Online selection should therefore favor the rows that hold most of the remaining key error that the reasoning reads.

\subsection{Propagating the Shift to Rows Not Recomputed}
\label{sec:diag-propagation}

After the 20\% recompute, most of the remaining damage lies in the other 80\% of the chunk rows, which keep their offline states: giving them full prefill's keys and values would remove \NAggPropRedOKVVsPKMin{}--\NAggPropRedOKVVsPKMax{}\% of it (Figure~\ref{fig:mechanism}c).
Part of their displacement is a shift that all rows of a chunk share: giving each of them the true mean change of its chunk's rows not recomputed, layer by layer, removes \NAggPropRedTUVsPKMin{}--\NAggPropRedTUVsPKMax{}\% of the damage.

Online, the recomputed rows show how their chunk shifted, because their states now reflect the prompt's context.
Adding their mean change from the cached states, layer by layer, to the other rows of their chunk, keys at half strength and values at full, removes \NAggPropRedMVsPKMin{}--\NAggPropRedMVsPKMax{}\% of the damage, in each of the three cells most of what the true mean change removes.

Taken together, the three steps lower the replay damage one after another (Figure~\ref{fig:diagnosis-steps}a), and each also lowers the tilt toward ``Wait'' of Section~\ref{sec:diag-length}, to \NAggLenRepMWaitRatioMin{}--\NAggLenRepMWaitRatioMax{} times with all three (Figure~\ref{fig:diagnosis-steps}b).
A repair at a small recompute share thus takes three steps, each acting on what the one before leaves.

%% file: sections/05-recipe.tex
\section{Assembled Recipe and Offline Evaluation}
\label{sec:recipe}

We assemble the three steps into a training-free recipe and evaluate it offline.

\begin{figure*}[t]
\centering
\includegraphics[width=\textwidth]{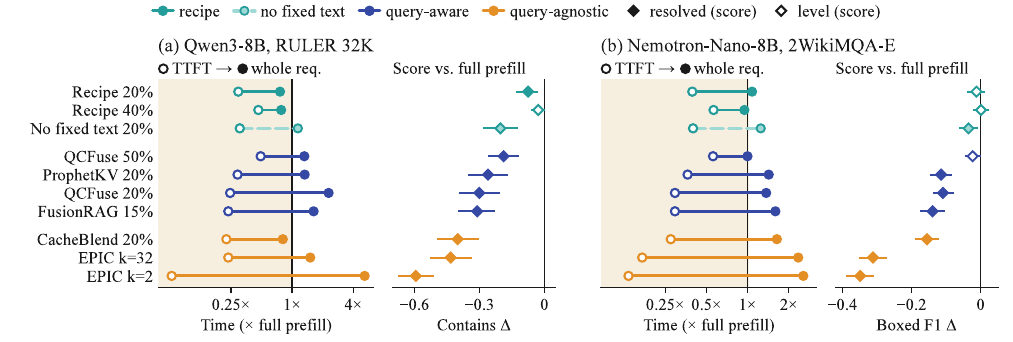}
\caption{Estimated request time on one H20 at batch 1, from the first token (hollow) to completion (filled), beside each setting's score change against full prefill (diamonds, filled where resolved), on the two cells where the published repairs were timed.}
\label{fig:time}
\end{figure*}

\begin{figure}[t]
\centering
\includegraphics[width=\columnwidth]{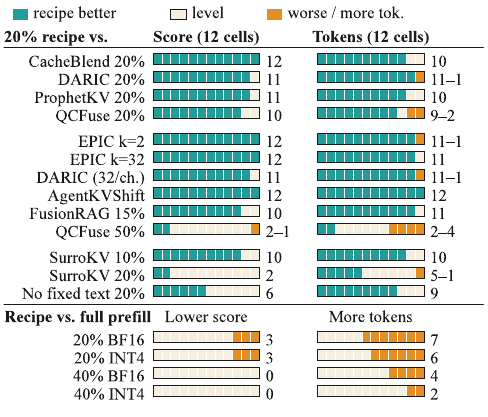}
\caption{Of the 12 cells, those where the 20\% recipe (BF16) scores higher or lower than each repair and generates fewer or more tokens, each resolved by its 95\% interval; bottom: those where the recipe, at each share and store, scores lower or generates more tokens than full prefill (per-cell values in Appendix~\ref{app:core-cells}).}
\label{fig:core-grid}
\end{figure}

\subsection{Assembled Recipe}
\label{sec:recipe-spec}

The recipe implements the three steps of Figure~\ref{fig:framework} with the mechanisms Section~\ref{sec:diagnosis} credits (Algorithm~\ref{alg:recipe} in Appendix~\ref{app:recipe}).
Offline, each chunk $C_m$ is compiled once after the shared prefix $P$ and the 2.2K-token Wikipedia text of Section~\ref{sec:diag-offline}, whose rows are then dropped, and its cache is stored in BF16 or in a KIVI-style INT4 store \citep{liu-kivi-2024}. %
Online, every token of the suffix $Q$ scores the chunk tokens by its attention, averaged over heads and layers; the 20\% of the chunk rows scored highest are recomputed through all layers, and for each chunk their mean change is added, layer by layer, to its other rows, keys at half strength and values at full.
We report a 40\% setting beside the 20\% one.

\subsection{Results on the 12 Cells}
\label{sec:core-eval}

We evaluate the recipe on the 12 cells of Section~\ref{sec:setup}, with the same questions and seeds (Figure~\ref{fig:core-grid}).
Against the published repairs run at the same 20\% share (CacheBlend, DARIC, ProphetKV and QCFuse), the recipe scores higher in \NAggFOneTwoMTwoZeroVsQZTwoZeroAccGainCells{}--\NAggGOneTwoMTwoZeroVsCBAccGainCells{} of the 12 cells, scores lower in none, and generates fewer tokens in \NAggFOneTwoMTwoZeroVsQZTwoZeroGenShorterCells{}--\NAggDarDAPTwoZeroVsMTwoZeroGenPosCells{}.
Figure~\ref{fig:core-grid} also reports the published repairs run at their own shares.
Against QCFuse at its default 50\%, the 20\% recipe is level in most cells, below in \NAggFOneTwoMTwoZeroVsQZFiveZeroAccLossCells{} and above in \NAggFOneTwoMTwoZeroVsQZFiveZeroAccGainCells{}, and QCFuse at 50\% generates fewer tokens on the four 2WikiMQA-E cells; at 40\%, the recipe is below QCFuse at 50\% in no cell and generates more tokens in none.
Concurrent SurroKV pairs a fixed 1,024-token Wikipedia text with ProphetKV's selection; against its default 10\%, the 20\% recipe scores higher in \NAggSkvSKOneZeroVsMTwoZeroAccNegCells{} cells, and SurroKV generates more tokens in \NAggSkvSKOneZeroVsMTwoZeroGenPosCells{}; at 20\%, SurroKV is level in \NAggSkvSKTwoZeroVsMTwoZeroAccLevelCells{} cells, scores lower on the two Nemotron LongBench-E cells, and generates more tokens in \NAggSkvSKTwoZeroVsMTwoZeroGenPosCells{} and fewer in \NAggSkvSKTwoZeroVsMTwoZeroGenNegCells{}.

Against full prefill, the 20\% recipe loses accuracy resolvably in \NAggGOneTwoMTwoZeroVsFAccLossCells{} of the 12 cells with each store, by \NAggMTwoZeroanyVsFAccLossMin{} to \NAggMTwoZeroanyVsFAccLossMax{}, and its extra tokens sit in the tail, the BF16 median within \NAggGOneTwoMTwoZeroVsFGenMedDiffMin{} to \NAggGOneTwoMTwoZeroVsFGenMedDiffMax{} tokens of full prefill's.
At 40\%, no resolved accuracy loss remains with either store, and the INT4 store's accuracy differs resolvably from BF16's in no cell at either share.

\subsection{Time to Finish a Request}
\label{sec:time}

Since an early first token no longer tells how much time PIC saves, we compare the repairs by the time a whole request takes: its first token plus decoding its own tokens, estimated on one H20 at batch 1 from measured prefill and decoding steps (Appendix~\ref{app:estimate}), on the two cells where the published repairs were timed (Figure~\ref{fig:time}).
In Figure~\ref{fig:time}, every published repair reaches the first token sooner than full prefill, yet on each cell six of the seven published settings finish later than full prefill, taking \NETwoeQEightrQZFiveZeroRatio{}--\NETwoepQEightrLTwoRatio{} times its request time on Qwen3-8B RULER 32K and \NETwoeNwQZTwoZeroRatio{}--\NETwoepNwLTwoRatio{} times on Nemotron 2WikiMQA-E.
The recipe's online steps reach the first token later than CacheBlend's and QCFuse's at the same share and FusionRAG's at its 15\%, and about as soon as ProphetKV's; still, on Qwen3-8B RULER 32K the recipe with the BF16 store finishes in \NETwoeQEightrMTwoZeroRatio{} of full prefill's request time at 20\% and \NETwoeQEightrMFourZeroRatio{} at 40\%.
QCFuse at 50\% takes \NETwoeQEightrQZFiveZeroRatio{} there.

\subsection{Ablations and Design Choices}
\label{sec:ablations}

The recipe's selection is ProphetKV's, compared with other rules in Section~\ref{sec:diag-selection}; the ablations remove its other two steps.
Without the fixed text, with chunks compiled as in Section~\ref{sec:setup}, the same online steps lose accuracy against full prefill in \NAggGOneTwoPKpTwoZeroVsFAccLossCells{} cells and generate more tokens in \NAggGOneTwoPKpTwoZeroVsFGenLongerCells{}, while the recipe scores higher than them in \NAggGOneTwoMTwoZeroVsPKpTwoZeroAccGainCells{} cells, lower in none, and generates \NAggGOneTwoMTwoZeroVsPKpTwoZeroGenShorterMin{}--\NAggGOneTwoMTwoZeroVsPKpTwoZeroGenShorterMax{} fewer tokens in \NAggGOneTwoMTwoZeroVsPKpTwoZeroGenShorterCells{} (Figure~\ref{fig:core-grid}).
On Qwen3-8B RULER 32K and Nemotron 2WikiMQA-E, we also remove the propagation.
The propagation changes boxed F1 by \NPubNwMTwoZeroVsMnpBoxPt{} on Nemotron 2WikiMQA-E and the generated tokens by \NPubQEightrMTwoZeroVsMnpGenPt{} on Qwen3-8B RULER 32K, where contains is level.

\subsection{Tasks, Corpora, Chunk Sizes, and Larger Models}
\label{sec:breadth}

Figure~\ref{fig:breadth} in Appendix~\ref{app:breadth} carries the comparison beyond Wikipedia retrieval QA, short passages and 8B--14B models.
On summarization, code and math (LongBench-E's MultiNews, SAMSum and RepoBench-P \citep{fabbri-multinews-2019,gliwa-samsum-2019,liu-repobench-2023}, and MATH-500 \citep{hendrycks-measuring-2021,lightman-lets-2023}), CacheBlend at 20\% loses score on RepoBench-P for all four models and on MultiNews for three, and generates more tokens on MATH-500 for all four, while the recipe has no resolved score loss in \NAggTbMTwoZeroVsFTaskNolossCells{} of the 16 cells at 20\% and in all \NAggTbMFourZeroVsFTaskNolossCells{} at 40\%. %
On dense Qwen3-32B and MoE Qwen3-30B-A3B, CacheBlend loses accuracy and adds tokens in all six cells, and the 20\% recipe scores above it in all six.
Section~\ref{sec:serving} measures the recipe when served.

%% file: sections/06-serving.tex
\section{End-to-End Serving and Amortization}
\label{sec:serving}

We now serve the recipe under load and measure how many uses repay its offline compile.

\begin{figure}[t]
\centering
\includegraphics[width=\columnwidth]{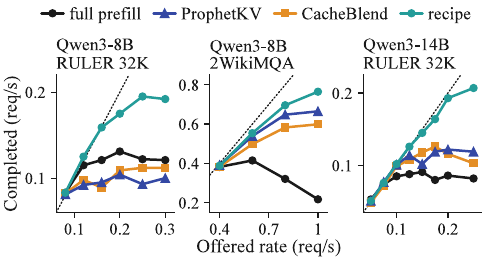}
\caption{Requests completed per second against the offered rate, 20 requests left out at each end of a trace, on one H20 GPU per system: full prefill, the recipe and the two published plans of Section~\ref{sec:served-cost}, the last three at 20\% recompute with INT4 stores. Dotted line: the offered rate.}
\label{fig:served-rate}
\end{figure}

\subsection{Serving on H20}
\label{sec:serving-h20}

We serve each system on its own H20 GPU in SGLang 0.5.11 with our plugin, which stores compiled chunks in INT4 in host memory, dequantizes them on load, and runs the repair synchronously.
The time PIC saves depends on how fast the GPU runs prefill relative to decoding.
A cell's systems run at the same time on one node over the same 100-request Poisson traces, two seeds per rate, and the two published plans of Section~\ref{sec:served-cost} compile each chunk after the shared prefix $P$ and run without propagation. %
A system's completion rate counts the requests it finishes per second while the middle of a trace arrives, leaving out 10, 20 or 30 requests at each end. %
SGLang's radix cache is off in every system.
Turned on for full prefill, as SGLang's default has it, the cache changes its peak completion rate by \NAggRadixFrxOverFMin{}--\NAggRadixFrxOverFMax{}$\times$ at the rates where both settings were run.

Served beside the two published plans in the runs of Section~\ref{sec:served-cost}, the recipe at 20\% completes requests at low load in \NAggSfLatLowMFourMin{}--\NAggSfLatLowMFourMax{}$\times$ full prefill's mean time and finishes them sooner than both published plans in every cell.
Across the offered rates, its peak completion rate is \NAggSfQEightwMFourOverPubMin{}--\NAggSfQEightrMFourOverPubMax{}$\times$ theirs over the three windows, and Figure~\ref{fig:served-rate} draws the middle one.
In separate runs on the same cells, which serve the recipe at 20\% and 40\% beside full prefill, its peak completion rate over the rates tested is \NAggCapPTwoZeroRatioMin{}--\NAggCapOneFourPTwoZeroRatioMax{}$\times$ full prefill's at 20\% and \NAggCapPFourZeroRatioMin{}--\NAggCapOneFourPFourZeroRatioMax{}$\times$ at 40\%.
In every cell and window, its peak is higher at 20\% than at 40\%.

\subsection{Offline Compile Cost and Amortization}
\label{sec:serving-amort}

The runs above compile every chunk during warm-up, outside the requests.
Over five cells in our engine, compiling a prompt's chunks behind the shared prefix and the fixed text takes \NAggCcMRatioMin{}--\NAggCcMRatioMax{}$\times$ as long as a full prefill of the prompt, against \NAggCcFRRatioMin{}--\NAggCcFRRatioMax{}$\times$ for FusionRAG's compile behind retrieved neighbors on the three cells where it was timed.
To measure how many uses repay the compile, we serve eight Qwen3-8B RULER 32K questions one request at a time, each used 16 times with its passages reordered after the first use.
The store starts empty, so a question's first use compiles its chunks inside the request.
Over all the requests, the recipe takes \NPOnelifeMFourRatio{}$\times$ full prefill's time at 20\% and \NPOnelifeMFourpFourZeroRatio{}$\times$ at 40\%, and the same online steps at 20\% without the fixed text take \NPOnelifeNFourRatio{}$\times$.
At 20\%, the recipe's mean time per request, counted over all uses so far, falls below full prefill's at the fifth use and stays below.

%% file: sections/07-conclusion.tex
\section{Conclusion}
\label{sec:conclusion}

We revisited eight PIC repairs on reasoning models and found that at small recompute shares they lose accuracy and lengthen the reasoning.
Tracing the damage behind these costs by replay showed that it starts with compiling each chunk alone, and that a repair at a small recompute share takes three steps, each acting on what the one before leaves.
The training-free recipe assembled from these steps scores higher than each published repair at the same 20\% share in \NAggFOneTwoMTwoZeroVsQZTwoZeroAccGainCells{}--\NAggGOneTwoMTwoZeroVsCBAccGainCells{} of the 12 cells and, served on H20, reaches \NAggCapPTwoZeroRatioMin{}--\NAggCapOneFourPTwoZeroRatioMax{}$\times$ full prefill's peak completion rate over the rates tested.
At 40\%, it has no resolved accuracy loss against full prefill in any of the 12 cells, and its peak completion rate is still \NAggCapPFourZeroRatioMin{}--\NAggCapOneFourPFourZeroRatioMax{}$\times$ full prefill's.
On reasoning models, an early first token no longer tells how much time PIC saves, because the longer reasoning that cache damage causes can use up part or all of the prefill saving.

%% file: sections/A-ports.tex
\newpage
\appendix
\onecolumn

\section{The Reimplemented Repairs}
\label{app:ports}

All eight repairs run in the engine of Section~\ref{sec:setup} with BF16 caches, on the same prompts and settings as full prefill.
Each takes the datasets' own passages as chunks.
A repair without an offline compile context compiles chunks alone on Qwen3 and after the shared prefix $P$ on the Llama-based models.
Table~\ref{tab:repairs} lists each repair's compile and online mechanisms and the recompute shares it is evaluated at.

\begin{table}[!ht]
\centering
\small
\caption{The eight PIC repairs, their compile and online mechanisms, and the evaluated recompute share $\rho$ over chunk tokens.}
\label{tab:repairs}
\setlength{\tabcolsep}{4pt}
\begin{tabular}{@{}>{\raggedright\arraybackslash}p{2.35cm}>{\raggedright\arraybackslash}p{11.2cm}>{\raggedright\arraybackslash}p{3.0cm}@{}}
\toprule
\textbf{Repair} & \textbf{Compile \& Online Mechanism} & \textbf{Recompute Share $\rho$} \\
\midrule
EPIC & Standalone compile; first $k$ tokens per chunk ($k{\in}\{2,32\}$). & \NAggFOneTwosLTwoShareMin{}--\NAggFOneTwosLTwoShareMax{}\%; \NAggFOneTwosLThreeTwoShareMin{}--\NAggFOneTwosLThreeTwoShareMax{}\% \\
DARIC & Standalone compile; monotonic chunk heads + trailing window. & avg.\ 32~tok/chunk; 20\% \\
CacheBlend & Standalone compile; layer-1 value-deviation top-$\rho$, one fixed set. & 20\% \\
AgentKVShift & Standalone compile; per-chunk K/V probe union + weighted shift. & \NAggFOneTwosAKShareMin{}--\NAggFOneTwosAKShareMax{}\% \\
ProphetKV & Standalone compile; all-layer query-attention global top-$\rho$. & 20\% \\
QCFuse & 15\% per-layer anchors; the question scores context keys at 3 critical layers. & 20\%; 50\% (default) \\
FusionRAG & Compile behind BGE-M3 \citep{chen-m3embedding-2024} top-10 neighbors; last-layer query top-$\rho$. & 15\% \\
SurroKV & Compile behind fixed 1,024-token Wikipedia text; ProphetKV selection. & 10\% (20\% in \S\ref{sec:recipe}) \\
\bottomrule
\end{tabular}
\end{table}

Against full prefill in the 12 cells of Figure~\ref{fig:revisiting-grid}, the repairs' counts are as follows.

Query-agnostic selectors degrade most severely.
EPIC loses accuracy in all \NAggFOneTwoLTwoVsFAccLossCells{} cells at $k=2$ and all \NAggFOneTwoLThreeTwoVsFAccLossCells{} at $k=32$, and lengthens outputs in \NAggFOneTwoLTwoVsFGenLongerCells{} and \NAggFOneTwoLThreeTwoVsFGenLongerCells{}, respectively.
At 32 tokens per chunk on average, DARIC loses accuracy in \NAggDarDAThreeTwoVsFAccNegCells{} cells and adds tokens in \NAggDarDAThreeTwoVsFGenPosCells{}, and scaled to 20\% it loses accuracy in \NAggDarDAPTwoZeroVsFAccNegCells{} and adds tokens in \NAggDarDAPTwoZeroVsFGenPosCells{}.
CacheBlend (20\%) loses accuracy in \NAggGOneTwoCBVsFAccLossCells{} cells and adds tokens in \NAggGOneTwoCBVsFGenLongerCells{}.
AgentKVShift loses accuracy in all \NAggFOneTwoAKVsFAccLossCells{} cells and adds tokens in all \NAggFOneTwoAKVsFGenLongerCells{}, despite recomputing \NAggFOneTwosAKShareMin{}--\NAggFOneTwosAKShareMax{}\% via its probe union.

The query-aware selectors and the repairs with an offline compile context lose accuracy in fewer cells, but still in most of them.
ProphetKV and QCFuse at 20\% lose accuracy in \NAggFOneTwoPKVsFAccLossCells{} and \NAggFOneTwoQZTwoZeroVsFAccLossCells{} cells and lengthen outputs in \NAggFOneTwoPKVsFGenLongerCells{} and \NAggFOneTwoQZTwoZeroVsFGenLongerCells{}.
FusionRAG (15\%) loses accuracy in \NAggGOneTwoFRVsFAccLossCells{} cells and adds tokens in \NAggGOneTwoFRVsFGenLongerCells{}.
Concurrent SurroKV at its default 10\% loses accuracy in \NAggSkvSKOneZeroVsFAccNegCells{} and adds tokens in \NAggSkvSKOneZeroVsFGenPosCells{} (20\% is evaluated in Section~\ref{sec:recipe}).

Table~\ref{tab:ports} lists what each port follows, the released code where one exists and the published algorithm otherwise, and its adaptations to our prompts and engine.

\begin{table}[!ht]
\centering
\small
\caption{What each reimplemented repair follows and its adaptations in our engine.}
\label{tab:ports}
\setlength{\tabcolsep}{4pt}
\setlength{\extrarowheight}{1.5pt}
\begin{tabular}{@{}>{\raggedright\arraybackslash}p{2.0cm}>{\raggedright\arraybackslash}p{7.25cm}>{\raggedright\arraybackslash}p{7.3cm}@{}}
\toprule
\textbf{Repair} & \textbf{What the port follows} & \textbf{Adaptations} \\
\midrule
EPIC \citep{hu-epic-2024} & Released code's LegoLink-$k$ (EPIC's name for the rule): the first $k$ tokens of each chunk after the first are recomputed; $k{=}2$, its headline setting, and $k{=}32$, its largest. & The first chunk's first $k$ tokens are recomputed too, since in our prompts it follows the shared prefix $P$. \\
DARIC \citep{wang-daric-2026} & Released code's selection rule: the first tokens of each document chunk, more for later chunks, and a window before the prompt's last token; the other rows keep their compiled keys and values, the keys re-rotated. Run at the paper's defaults (a linear allocation that gives the first document chunk $r{=}0.1$ of the average, and 32 tokens a chunk on average) and with the same allocation scaled to 20\% of the chunk rows. & Its instruction chunk is our prefix $P$, computed exactly, and its question chunk our suffix $Q$, which every method computes whole; window rows that fall before $Q$ are recomputed. \\ %
CacheBlend \citep{yao-cacheblend-2024} & Released code: the rows with the largest layer-1 value deviation from a fresh layer-1 pass over the prompt are recomputed, one set for every layer. & Run at 20\% of the chunk rows, where the paper's default is 15\%, with the released code's single selection in place of the paper's narrowing layer by layer. The fresh layer-1 values are read from full prefill, and the request-time estimate of Figure~\ref{fig:time} charges one layer of full prefill for them. \\
AgentKVShift \citep{pandey-agentkvshift-2026} & The paper's algorithm (no released code): per chunk of $n$ rows, the $\lceil 0.2n \rceil$ rows with the largest layer-1 key deviation and, separately, those with the largest value deviation are recomputed as key and value probes; each layer's mean change of the key probes is added to the keys of the rows outside them, and that of the value probes to the values of the rows outside those, each row weighted by $\min(\text{its deviation}, 1)$. & Keys are averaged before RoPE, and the change is applied before each layer's probe attention. The union recomputes \NAggFOneTwosAKShareMin{}--\NAggFOneTwosAKShareMax{}\% of the chunk rows, the range of the 12 cells' means. \\ %
ProphetKV \citep{wang-prophetkv-2026} & The paper's Algorithm~1 (no released code): one pass of the query over the reused cache, its attention summed over layers, the top 20\% of chunk rows recomputed through every layer, the other rows reused as compiled. & The query is every token after the chunks. \\
QCFuse \citep{yan-qcfuse-2026} & Released code: offline, each chunk keeps, per layer, 15\% of its tokens as anchors, ranked from its first 4 and then by how much an appended request to repeat the context attends to them; online, the question, read over these anchors, scores every context token at three critical layers, those its per-model calibration ranks best at locating answers, and the top $\rho$ are recomputed from layer 0. & R1-Distill-Llama-8B and Nemotron-Nano-8B, both fine-tuned from Llama-3.1-8B \citep{grattafiori-llama-2024}, use the critical layers its code gives for that base model. \\ %
FusionRAG \citep{wang-prefix-2026} & The paper (no released code): each chunk is compiled after the caches of its 10 nearest corpus passages by BGE-M3 \citep{chen-m3embedding-2024}, their rows dropped; online, the 15\% of chunk rows the question attends to most at the last layer are recomputed. & Neighbors are added in similarity order while they fit the model's context window. The question's last-layer queries are computed over the assembled cache and the prompt's instructions before it. \\
SurroKV \citep{anon-surrokv-2026} & The paper (no released code): each chunk is compiled behind the fixed Wikipedia text its Table~6 prints, whose rows are dropped; online, ProphetKV's selection at SurroKV's default of 10\% and at the recipe's 20\%, the other rows left as compiled. & The text is cut to its first 1,024 tokens by each model's own tokenizer. \\ %
\bottomrule
\end{tabular}
\end{table}

%% file: sections/D-estimate.tex
\section{Request-Time Estimate}
\label{app:estimate}

Where a setting was not served, we estimate a request's time from prefill and decode steps measured with SGLang 0.5.11 on one H20.
Its first token takes SGLang's prefill time for full prefill, scaled by the method's first-token time relative to full prefill's in our engine, with the chunk store in GPU memory.
Its decoding takes SGLang's decode step summed over its generated tokens.
With $B$ alike requests at once, the prefill and decode steps are measured at batch $B$ and divided among the requests, which gives each request's share of the GPU time.

Over all 12 cells of Section~\ref{sec:setup}, the estimate finds the recipe at 20\% with the BF16 store faster than full prefill in \NAggGOneTwoeBOneMTwoZeroFasterCells{} at batch 1 and level with it in the rest.
In GPU time per request, it is faster in \NAggGOneTwoeBFourMTwoZeroFasterCells{} of them at batch 4, level in the rest, and faster in every cell estimated at batches 8 and 16.

In the estimate, which charges the compile at its measured ratio to full prefill (Section~\ref{sec:serving-amort}), the compile is repaid in GPU time after \NPOneamortAggRulerUsesFMin{}--\NPOneamortAggRulerUsesFMax{} uses on RULER 32K and \NPOneamortAggShortUsesFMin{}--\NPOneamortAggShortUsesFMax{} on LongBench-E's shorter prompts wherever a use saves time.

%% file: sections/E-fixedtext.tex
\section{The Recipe in Detail}
\label{app:recipe}
\label{app:fixedtext}

Algorithm~\ref{alg:recipe} gives the recipe of Section~\ref{sec:recipe-spec} step by step, in the notation of Section~\ref{sec:background}.
$R_i$ rotates a key by RoPE to position $i$ of the prompt; $k_i$ and $v_i$ are a recomputed token's new key and value, and $\tilde{k}_i$ and $\tilde{v}_i$ the cached ones they replace.
The same strengths, $\alpha_K = 0.5$ for keys and $\alpha_V = 1$ for values, serve every model and task. %

\begin{algorithm}[t]
\caption{The recipe at recompute share $\rho$ (20\% or 40\%).}
\label{alg:recipe}
\begin{algorithmic}[1]
\STATE {\bfseries Input:} shared prefix $P$, chunks $C_1, \dots, C_M$, suffix $Q$, the fixed text $F$, strengths $\alpha_K$ and $\alpha_V$
\STATE \textbf{Offline}, for each chunk $C_m$: prefill $[P, F, C_m]$, keep only $C_m$'s keys and values $(\tilde{K}_m, \tilde{V}_m)$, and store them in BF16 or INT4
\STATE \textbf{Online}, for a prompt $X = [P, C_1, \dots, C_M, Q]$: load each $(\tilde{K}_m, \tilde{V}_m)$, rotate $\tilde{K}_m$ to $C_m$'s positions in $X$, and place the chunks after $P$
\STATE Score each chunk token by the attention that $Q$'s tokens pay it over this cache, summed over $Q$'s tokens and averaged over heads, then over layers
\STATE $S \gets$ the share $\rho$ of all chunk tokens with the highest scores
\STATE Recompute the tokens in $S$ layer by layer: in each layer, write their new keys and values $k_i, v_i$ into the cache in place of $\tilde{k}_i, \tilde{v}_i$, then let them attend over it
\FOR{each chunk $C_m$ with $S_m = S \cap C_m$ not empty, each layer and each KV head}
\STATE $\Delta k \gets \frac{1}{|S_m|} \sum_{i \in S_m} R_{-i}\,(k_i - \tilde{k}_i)$ \hfill (the key changes, rotated back to before RoPE)
\STATE $\Delta v \gets \frac{1}{|S_m|} \sum_{i \in S_m} (v_i - \tilde{v}_i)$
\STATE for each $j \in C_m \setminus S_m$: \ $\tilde{k}_j \gets \tilde{k}_j + \alpha_K R_j\,\Delta k$, \ $\tilde{v}_j \gets \tilde{v}_j + \alpha_V\,\Delta v$
\ENDFOR
\STATE Prefill $Q$ over the repaired cache and decode
\end{algorithmic}
\end{algorithm}

The fixed text $F$, the 2.2K-token Wikipedia text of Section~\ref{sec:diag-offline}, is the first four passages of one example of LongBench's MuSiQue \citep{trivedi-musique-2021}. %
No other length of the compile text, from 0.44K to 6.1K tokens in replay (Section~\ref{sec:diagnosis}) and 1.0K or 4.2K in generation, did resolvably better than this text, and compiling each chunk behind its retrieved neighbors, as FusionRAG does, brings no gain over the fixed text under the recipe's online steps on the two cells tested, Qwen3-8B RULER 32K and Nemotron 2WikiMQA-E. %
When $\rho$ is swept from 5\% to 50\% on these two cells, the recipe's gain over the same online steps without the fixed text, with chunks compiled as in Section~\ref{sec:setup}, is largest at the smallest share, at 5\% \NSwQEightrMFiveVsPKpFiveConPt{} contains and \NSwNwMFiveVsPKpFiveBoxPt{} boxed F1.
With the passages of another MuSiQue example in place of this text, the recipe on Nemotron 2WikiMQA-E is level with it in boxed F1 and in generated tokens (\NFTwoNwMXOneVsMTwoZeroBoxPt{} and \NFTwoNwMXOneVsMTwoZeroGenPt{} against this text).

%% file: sections/B-cells.tex
\section{Per-Cell Results Behind Figure 7}
\label{app:core-cells}

Tables~\ref{tab:cells-score} and~\ref{tab:cells-tokens} give each setting's score and mean generated tokens in the 12 cells of Section~\ref{sec:core-eval}, from which Figure~\ref{fig:core-grid} counts.

\begin{table}[!ht]
\centering
\footnotesize
\caption{Score per cell, over every question of Section~\ref{sec:setup} and four seeds: contains on RULER 32K, F1 on LongBench-E's 2WikiMQA and HotpotQA, boxed F1 for Nemotron-Nano-8B there. $\uparrow$/$\downarrow$: higher or lower, resolved by the paired 95\% interval over questions, each question averaged over its seeds; for a repair, against the 20\% recipe (BF16); for a recipe setting, against full prefill. Q: Qwen3; R1: R1-Distill-Llama; N: Nemotron-Nano.}
\label{tab:cells-score}
\setlength{\tabcolsep}{2.6pt}
\begin{tabular}{@{}l*{12}{r}@{}}
\toprule
 & \multicolumn{4}{c}{RULER 32K (contains)} & \multicolumn{4}{c}{2WikiMQA-E (F1)} & \multicolumn{4}{c}{HotpotQA-E (F1)} \\
\cmidrule(lr){2-5}\cmidrule(lr){6-9}\cmidrule(lr){10-13}
 & Q-8B & Q-14B & R1-8B & N-8B & Q-8B & Q-14B & R1-8B & N-8B & Q-8B & Q-14B & R1-8B & N-8B \\
\midrule
Full prefill & \NFcQEightrFCon{}\pcblank & \NFcQOneFourrFCon{}\pcblank & \NFcROnerFCon{}\pcblank & \NGrNrFCon{}\pcblank & \NFcQEightwFFOne{}\pcblank & \NFcQOneFourwFFOne{}\pcblank & \NGrROnewFFOne{}\pcblank & \NFcNwFBox{}\pcblank & \NFcQEighthFFOne{}\pcblank & \NFcQOneFourhFFOne{}\pcblank & \NGrROnehFFOne{}\pcblank & \NFcNhFBox{}\pcblank \\
\midrule
CacheBlend 20\% & \NFcQEightrCBCon{}\pcmarkr{\NFcQEightrMTwoZeroVsCBConLo}{\NFcQEightrMTwoZeroVsCBConHi} & \NFcQOneFourrCBCon{}\pcmarkr{\NFcQOneFourrMTwoZeroVsCBConLo}{\NFcQOneFourrMTwoZeroVsCBConHi} & \NFcROnerCBCon{}\pcmarkr{\NFcROnerMTwoZeroVsCBConLo}{\NFcROnerMTwoZeroVsCBConHi} & \NGrNrCBCon{}\pcmarkr{\NGrNrMTwoZeroVsCBConLo}{\NGrNrMTwoZeroVsCBConHi} & \NFcQEightwCBFOne{}\pcmarkr{\NFcQEightwMTwoZeroVsCBFOneLo}{\NFcQEightwMTwoZeroVsCBFOneHi} & \NFcQOneFourwCBFOne{}\pcmarkr{\NFcQOneFourwMTwoZeroVsCBFOneLo}{\NFcQOneFourwMTwoZeroVsCBFOneHi} & \NGrROnewCBFOne{}\pcmarkr{\NGrROnewMTwoZeroVsCBFOneLo}{\NGrROnewMTwoZeroVsCBFOneHi} & \NFcNwCBBox{}\pcmarkr{\NFcNwMTwoZeroVsCBBoxLo}{\NFcNwMTwoZeroVsCBBoxHi} & \NFcQEighthCBFOne{}\pcmarkr{\NFcQEighthMTwoZeroVsCBFOneLo}{\NFcQEighthMTwoZeroVsCBFOneHi} & \NFcQOneFourhCBFOne{}\pcmarkr{\NFcQOneFourhMTwoZeroVsCBFOneLo}{\NFcQOneFourhMTwoZeroVsCBFOneHi} & \NGrROnehCBFOne{}\pcmarkr{\NGrROnehMTwoZeroVsCBFOneLo}{\NGrROnehMTwoZeroVsCBFOneHi} & \NFcNhCBBox{}\pcmarkr{\NFcNhMTwoZeroVsCBBoxLo}{\NFcNhMTwoZeroVsCBBoxHi} \\
DARIC 20\% & \NDarQEightrDAPTwoZeroCon{}\pcmark{\NDarQEightrDAPTwoZeroVsMTwoZeroConLo}{\NDarQEightrDAPTwoZeroVsMTwoZeroConHi} & \NDarQOneFourrDAPTwoZeroCon{}\pcmark{\NDarQOneFourrDAPTwoZeroVsMTwoZeroConLo}{\NDarQOneFourrDAPTwoZeroVsMTwoZeroConHi} & \NDarROnerDAPTwoZeroCon{}\pcmark{\NDarROnerDAPTwoZeroVsMTwoZeroConLo}{\NDarROnerDAPTwoZeroVsMTwoZeroConHi} & \NDarNrDAPTwoZeroCon{}\pcmark{\NDarNrDAPTwoZeroVsMTwoZeroConLo}{\NDarNrDAPTwoZeroVsMTwoZeroConHi} & \NDarQEightwDAPTwoZeroFOne{}\pcmark{\NDarQEightwDAPTwoZeroVsMTwoZeroFOneLo}{\NDarQEightwDAPTwoZeroVsMTwoZeroFOneHi} & \NDarQOneFourwDAPTwoZeroFOne{}\pcmark{\NDarQOneFourwDAPTwoZeroVsMTwoZeroFOneLo}{\NDarQOneFourwDAPTwoZeroVsMTwoZeroFOneHi} & \NDarROnewDAPTwoZeroFOne{}\pcmark{\NDarROnewDAPTwoZeroVsMTwoZeroFOneLo}{\NDarROnewDAPTwoZeroVsMTwoZeroFOneHi} & \NDarNwDAPTwoZeroBox{}\pcmark{\NDarNwDAPTwoZeroVsMTwoZeroBoxLo}{\NDarNwDAPTwoZeroVsMTwoZeroBoxHi} & \NDarQEighthDAPTwoZeroFOne{}\pcmark{\NDarQEighthDAPTwoZeroVsMTwoZeroFOneLo}{\NDarQEighthDAPTwoZeroVsMTwoZeroFOneHi} & \NDarQOneFourhDAPTwoZeroFOne{}\pcmark{\NDarQOneFourhDAPTwoZeroVsMTwoZeroFOneLo}{\NDarQOneFourhDAPTwoZeroVsMTwoZeroFOneHi} & \NDarROnehDAPTwoZeroFOne{}\pcmark{\NDarROnehDAPTwoZeroVsMTwoZeroFOneLo}{\NDarROnehDAPTwoZeroVsMTwoZeroFOneHi} & \NDarNhDAPTwoZeroBox{}\pcmark{\NDarNhDAPTwoZeroVsMTwoZeroBoxLo}{\NDarNhDAPTwoZeroVsMTwoZeroBoxHi} \\
ProphetKV 20\% & \NPubQEightrPKCon{}\pcmarkr{\NPubQEightrMTwoZeroVsPKConLo}{\NPubQEightrMTwoZeroVsPKConHi} & \NFOneTwoQOneFourrPKCon{}\pcmarkr{\NFOneTwoQOneFourrMTwoZeroVsPKConLo}{\NFOneTwoQOneFourrMTwoZeroVsPKConHi} & \NFOneTwoROnerPKCon{}\pcmarkr{\NFOneTwoROnerMTwoZeroVsPKConLo}{\NFOneTwoROnerMTwoZeroVsPKConHi} & \NFOneTwoNrPKCon{}\pcmarkr{\NFOneTwoNrMTwoZeroVsPKConLo}{\NFOneTwoNrMTwoZeroVsPKConHi} & \NFOneTwoQEightwPKFOne{}\pcmarkr{\NFOneTwoQEightwMTwoZeroVsPKFOneLo}{\NFOneTwoQEightwMTwoZeroVsPKFOneHi} & \NFOneTwoQOneFourwPKFOne{}\pcmarkr{\NFOneTwoQOneFourwMTwoZeroVsPKFOneLo}{\NFOneTwoQOneFourwMTwoZeroVsPKFOneHi} & \NFOneTwoROnewPKFOne{}\pcmarkr{\NFOneTwoROnewMTwoZeroVsPKFOneLo}{\NFOneTwoROnewMTwoZeroVsPKFOneHi} & \NPubNwPKBox{}\pcmarkr{\NPubNwMTwoZeroVsPKBoxLo}{\NPubNwMTwoZeroVsPKBoxHi} & \NFOneTwoQEighthPKFOne{}\pcmarkr{\NFOneTwoQEighthMTwoZeroVsPKFOneLo}{\NFOneTwoQEighthMTwoZeroVsPKFOneHi} & \NFOneTwoQOneFourhPKFOne{}\pcmarkr{\NFOneTwoQOneFourhMTwoZeroVsPKFOneLo}{\NFOneTwoQOneFourhMTwoZeroVsPKFOneHi} & \NFOneTwoROnehPKFOne{}\pcmarkr{\NFOneTwoROnehMTwoZeroVsPKFOneLo}{\NFOneTwoROnehMTwoZeroVsPKFOneHi} & \NFOneTwoNhPKBox{}\pcmarkr{\NFOneTwoNhMTwoZeroVsPKBoxLo}{\NFOneTwoNhMTwoZeroVsPKBoxHi} \\
QCFuse 20\% & \NQzQEightrQZTwoZeroCon{}\pcmarkr{\NQzQEightrMTwoZeroVsQZTwoZeroConLo}{\NQzQEightrMTwoZeroVsQZTwoZeroConHi} & \NFOneTwoQOneFourrQZTwoZeroCon{}\pcmarkr{\NFOneTwoQOneFourrMTwoZeroVsQZTwoZeroConLo}{\NFOneTwoQOneFourrMTwoZeroVsQZTwoZeroConHi} & \NFOneTwoROnerQZTwoZeroCon{}\pcmarkr{\NFOneTwoROnerMTwoZeroVsQZTwoZeroConLo}{\NFOneTwoROnerMTwoZeroVsQZTwoZeroConHi} & \NFOneTwoNrQZTwoZeroCon{}\pcmarkr{\NFOneTwoNrMTwoZeroVsQZTwoZeroConLo}{\NFOneTwoNrMTwoZeroVsQZTwoZeroConHi} & \NFOneTwoQEightwQZTwoZeroFOne{}\pcmarkr{\NFOneTwoQEightwMTwoZeroVsQZTwoZeroFOneLo}{\NFOneTwoQEightwMTwoZeroVsQZTwoZeroFOneHi} & \NFOneTwoQOneFourwQZTwoZeroFOne{}\pcmarkr{\NFOneTwoQOneFourwMTwoZeroVsQZTwoZeroFOneLo}{\NFOneTwoQOneFourwMTwoZeroVsQZTwoZeroFOneHi} & \NFOneTwoROnewQZTwoZeroFOne{}\pcmarkr{\NFOneTwoROnewMTwoZeroVsQZTwoZeroFOneLo}{\NFOneTwoROnewMTwoZeroVsQZTwoZeroFOneHi} & \NQzNwQZTwoZeroBox{}\pcmarkr{\NQzNwMTwoZeroVsQZTwoZeroBoxLo}{\NQzNwMTwoZeroVsQZTwoZeroBoxHi} & \NFOneTwoQEighthQZTwoZeroFOne{}\pcmarkr{\NFOneTwoQEighthMTwoZeroVsQZTwoZeroFOneLo}{\NFOneTwoQEighthMTwoZeroVsQZTwoZeroFOneHi} & \NFOneTwoQOneFourhQZTwoZeroFOne{}\pcmarkr{\NFOneTwoQOneFourhMTwoZeroVsQZTwoZeroFOneLo}{\NFOneTwoQOneFourhMTwoZeroVsQZTwoZeroFOneHi} & \NFOneTwoROnehQZTwoZeroFOne{}\pcmarkr{\NFOneTwoROnehMTwoZeroVsQZTwoZeroFOneLo}{\NFOneTwoROnehMTwoZeroVsQZTwoZeroFOneHi} & \NFOneTwoNhQZTwoZeroBox{}\pcmarkr{\NFOneTwoNhMTwoZeroVsQZTwoZeroBoxLo}{\NFOneTwoNhMTwoZeroVsQZTwoZeroBoxHi} \\
\midrule
EPIC $k{=}2$ & \NEpQEightrLTwoCon{}\pcmarkr{\NEpQEightrMTwoZeroVsLTwoConLo}{\NEpQEightrMTwoZeroVsLTwoConHi} & \NFOneTwoQOneFourrLTwoCon{}\pcmarkr{\NFOneTwoQOneFourrMTwoZeroVsLTwoConLo}{\NFOneTwoQOneFourrMTwoZeroVsLTwoConHi} & \NFOneTwoROnerLTwoCon{}\pcmarkr{\NFOneTwoROnerMTwoZeroVsLTwoConLo}{\NFOneTwoROnerMTwoZeroVsLTwoConHi} & \NFOneTwoNrLTwoCon{}\pcmarkr{\NFOneTwoNrMTwoZeroVsLTwoConLo}{\NFOneTwoNrMTwoZeroVsLTwoConHi} & \NFOneTwoQEightwLTwoFOne{}\pcmarkr{\NFOneTwoQEightwMTwoZeroVsLTwoFOneLo}{\NFOneTwoQEightwMTwoZeroVsLTwoFOneHi} & \NFOneTwoQOneFourwLTwoFOne{}\pcmarkr{\NFOneTwoQOneFourwMTwoZeroVsLTwoFOneLo}{\NFOneTwoQOneFourwMTwoZeroVsLTwoFOneHi} & \NFOneTwoROnewLTwoFOne{}\pcmarkr{\NFOneTwoROnewMTwoZeroVsLTwoFOneLo}{\NFOneTwoROnewMTwoZeroVsLTwoFOneHi} & \NEpNwLTwoBox{}\pcmarkr{\NEpNwMTwoZeroVsLTwoBoxLo}{\NEpNwMTwoZeroVsLTwoBoxHi} & \NFOneTwoQEighthLTwoFOne{}\pcmarkr{\NFOneTwoQEighthMTwoZeroVsLTwoFOneLo}{\NFOneTwoQEighthMTwoZeroVsLTwoFOneHi} & \NFOneTwoQOneFourhLTwoFOne{}\pcmarkr{\NFOneTwoQOneFourhMTwoZeroVsLTwoFOneLo}{\NFOneTwoQOneFourhMTwoZeroVsLTwoFOneHi} & \NFOneTwoROnehLTwoFOne{}\pcmarkr{\NFOneTwoROnehMTwoZeroVsLTwoFOneLo}{\NFOneTwoROnehMTwoZeroVsLTwoFOneHi} & \NFOneTwoNhLTwoBox{}\pcmarkr{\NFOneTwoNhMTwoZeroVsLTwoBoxLo}{\NFOneTwoNhMTwoZeroVsLTwoBoxHi} \\
EPIC $k{=}32$ & \NEpQEightrLThreeTwoCon{}\pcmarkr{\NEpQEightrMTwoZeroVsLThreeTwoConLo}{\NEpQEightrMTwoZeroVsLThreeTwoConHi} & \NFOneTwoQOneFourrLThreeTwoCon{}\pcmarkr{\NFOneTwoQOneFourrMTwoZeroVsLThreeTwoConLo}{\NFOneTwoQOneFourrMTwoZeroVsLThreeTwoConHi} & \NFOneTwoROnerLThreeTwoCon{}\pcmarkr{\NFOneTwoROnerMTwoZeroVsLThreeTwoConLo}{\NFOneTwoROnerMTwoZeroVsLThreeTwoConHi} & \NFOneTwoNrLThreeTwoCon{}\pcmarkr{\NFOneTwoNrMTwoZeroVsLThreeTwoConLo}{\NFOneTwoNrMTwoZeroVsLThreeTwoConHi} & \NFOneTwoQEightwLThreeTwoFOne{}\pcmarkr{\NFOneTwoQEightwMTwoZeroVsLThreeTwoFOneLo}{\NFOneTwoQEightwMTwoZeroVsLThreeTwoFOneHi} & \NFOneTwoQOneFourwLThreeTwoFOne{}\pcmarkr{\NFOneTwoQOneFourwMTwoZeroVsLThreeTwoFOneLo}{\NFOneTwoQOneFourwMTwoZeroVsLThreeTwoFOneHi} & \NFOneTwoROnewLThreeTwoFOne{}\pcmarkr{\NFOneTwoROnewMTwoZeroVsLThreeTwoFOneLo}{\NFOneTwoROnewMTwoZeroVsLThreeTwoFOneHi} & \NEpNwLThreeTwoBox{}\pcmarkr{\NEpNwMTwoZeroVsLThreeTwoBoxLo}{\NEpNwMTwoZeroVsLThreeTwoBoxHi} & \NFOneTwoQEighthLThreeTwoFOne{}\pcmarkr{\NFOneTwoQEighthMTwoZeroVsLThreeTwoFOneLo}{\NFOneTwoQEighthMTwoZeroVsLThreeTwoFOneHi} & \NFOneTwoQOneFourhLThreeTwoFOne{}\pcmarkr{\NFOneTwoQOneFourhMTwoZeroVsLThreeTwoFOneLo}{\NFOneTwoQOneFourhMTwoZeroVsLThreeTwoFOneHi} & \NFOneTwoROnehLThreeTwoFOne{}\pcmarkr{\NFOneTwoROnehMTwoZeroVsLThreeTwoFOneLo}{\NFOneTwoROnehMTwoZeroVsLThreeTwoFOneHi} & \NFOneTwoNhLThreeTwoBox{}\pcmarkr{\NFOneTwoNhMTwoZeroVsLThreeTwoBoxLo}{\NFOneTwoNhMTwoZeroVsLThreeTwoBoxHi} \\
DARIC (32/chunk) & \NDarQEightrDAThreeTwoCon{}\pcmark{\NDarQEightrDAThreeTwoVsMTwoZeroConLo}{\NDarQEightrDAThreeTwoVsMTwoZeroConHi} & \NDarQOneFourrDAThreeTwoCon{}\pcmark{\NDarQOneFourrDAThreeTwoVsMTwoZeroConLo}{\NDarQOneFourrDAThreeTwoVsMTwoZeroConHi} & \NDarROnerDAThreeTwoCon{}\pcmark{\NDarROnerDAThreeTwoVsMTwoZeroConLo}{\NDarROnerDAThreeTwoVsMTwoZeroConHi} & \NDarNrDAThreeTwoCon{}\pcmark{\NDarNrDAThreeTwoVsMTwoZeroConLo}{\NDarNrDAThreeTwoVsMTwoZeroConHi} & \NDarQEightwDAThreeTwoFOne{}\pcmark{\NDarQEightwDAThreeTwoVsMTwoZeroFOneLo}{\NDarQEightwDAThreeTwoVsMTwoZeroFOneHi} & \NDarQOneFourwDAThreeTwoFOne{}\pcmark{\NDarQOneFourwDAThreeTwoVsMTwoZeroFOneLo}{\NDarQOneFourwDAThreeTwoVsMTwoZeroFOneHi} & \NDarROnewDAThreeTwoFOne{}\pcmark{\NDarROnewDAThreeTwoVsMTwoZeroFOneLo}{\NDarROnewDAThreeTwoVsMTwoZeroFOneHi} & \NDarNwDAThreeTwoBox{}\pcmark{\NDarNwDAThreeTwoVsMTwoZeroBoxLo}{\NDarNwDAThreeTwoVsMTwoZeroBoxHi} & \NDarQEighthDAThreeTwoFOne{}\pcmark{\NDarQEighthDAThreeTwoVsMTwoZeroFOneLo}{\NDarQEighthDAThreeTwoVsMTwoZeroFOneHi} & \NDarQOneFourhDAThreeTwoFOne{}\pcmark{\NDarQOneFourhDAThreeTwoVsMTwoZeroFOneLo}{\NDarQOneFourhDAThreeTwoVsMTwoZeroFOneHi} & \NDarROnehDAThreeTwoFOne{}\pcmark{\NDarROnehDAThreeTwoVsMTwoZeroFOneLo}{\NDarROnehDAThreeTwoVsMTwoZeroFOneHi} & \NDarNhDAThreeTwoBox{}\pcmark{\NDarNhDAThreeTwoVsMTwoZeroBoxLo}{\NDarNhDAThreeTwoVsMTwoZeroBoxHi} \\
AgentKVShift & \NPubQEightrAKCon{}\pcmarkr{\NPubQEightrMTwoZeroVsAKConLo}{\NPubQEightrMTwoZeroVsAKConHi} & \NFOneTwoQOneFourrAKCon{}\pcmarkr{\NFOneTwoQOneFourrMTwoZeroVsAKConLo}{\NFOneTwoQOneFourrMTwoZeroVsAKConHi} & \NFOneTwoROnerAKCon{}\pcmarkr{\NFOneTwoROnerMTwoZeroVsAKConLo}{\NFOneTwoROnerMTwoZeroVsAKConHi} & \NFOneTwoNrAKCon{}\pcmarkr{\NFOneTwoNrMTwoZeroVsAKConLo}{\NFOneTwoNrMTwoZeroVsAKConHi} & \NFOneTwoQEightwAKFOne{}\pcmarkr{\NFOneTwoQEightwMTwoZeroVsAKFOneLo}{\NFOneTwoQEightwMTwoZeroVsAKFOneHi} & \NFOneTwoQOneFourwAKFOne{}\pcmarkr{\NFOneTwoQOneFourwMTwoZeroVsAKFOneLo}{\NFOneTwoQOneFourwMTwoZeroVsAKFOneHi} & \NFOneTwoROnewAKFOne{}\pcmarkr{\NFOneTwoROnewMTwoZeroVsAKFOneLo}{\NFOneTwoROnewMTwoZeroVsAKFOneHi} & \NPubNwAKBox{}\pcmarkr{\NPubNwMTwoZeroVsAKBoxLo}{\NPubNwMTwoZeroVsAKBoxHi} & \NFOneTwoQEighthAKFOne{}\pcmarkr{\NFOneTwoQEighthMTwoZeroVsAKFOneLo}{\NFOneTwoQEighthMTwoZeroVsAKFOneHi} & \NFOneTwoQOneFourhAKFOne{}\pcmarkr{\NFOneTwoQOneFourhMTwoZeroVsAKFOneLo}{\NFOneTwoQOneFourhMTwoZeroVsAKFOneHi} & \NFOneTwoROnehAKFOne{}\pcmarkr{\NFOneTwoROnehMTwoZeroVsAKFOneLo}{\NFOneTwoROnehMTwoZeroVsAKFOneHi} & \NFOneTwoNhAKBox{}\pcmarkr{\NFOneTwoNhMTwoZeroVsAKBoxLo}{\NFOneTwoNhMTwoZeroVsAKBoxHi} \\
FusionRAG 15\% & \NFcQEightrFRCon{}\pcmarkr{\NFcQEightrMTwoZeroVsFRConLo}{\NFcQEightrMTwoZeroVsFRConHi} & \NFcQOneFourrFRCon{}\pcmarkr{\NFcQOneFourrMTwoZeroVsFRConLo}{\NFcQOneFourrMTwoZeroVsFRConHi} & \NFcROnerFRCon{}\pcmarkr{\NFcROnerMTwoZeroVsFRConLo}{\NFcROnerMTwoZeroVsFRConHi} & \NGrNrFRCon{}\pcmarkr{\NGrNrMTwoZeroVsFRConLo}{\NGrNrMTwoZeroVsFRConHi} & \NFcQEightwFRFOne{}\pcmarkr{\NFcQEightwMTwoZeroVsFRFOneLo}{\NFcQEightwMTwoZeroVsFRFOneHi} & \NFcQOneFourwFRFOne{}\pcmarkr{\NFcQOneFourwMTwoZeroVsFRFOneLo}{\NFcQOneFourwMTwoZeroVsFRFOneHi} & \NGrROnewFRFOne{}\pcmarkr{\NGrROnewMTwoZeroVsFRFOneLo}{\NGrROnewMTwoZeroVsFRFOneHi} & \NFcNwFRBox{}\pcmarkr{\NFcNwMTwoZeroVsFRBoxLo}{\NFcNwMTwoZeroVsFRBoxHi} & \NFcQEighthFRFOne{}\pcmarkr{\NFcQEighthMTwoZeroVsFRFOneLo}{\NFcQEighthMTwoZeroVsFRFOneHi} & \NFcQOneFourhFRFOne{}\pcmarkr{\NFcQOneFourhMTwoZeroVsFRFOneLo}{\NFcQOneFourhMTwoZeroVsFRFOneHi} & \NGrROnehFRFOne{}\pcmarkr{\NGrROnehMTwoZeroVsFRFOneLo}{\NGrROnehMTwoZeroVsFRFOneHi} & \NFcNhFRBox{}\pcmarkr{\NFcNhMTwoZeroVsFRBoxLo}{\NFcNhMTwoZeroVsFRBoxHi} \\
QCFuse 50\% & \NQzQEightrQZFiveZeroCon{}\pcmarkr{\NQzQEightrMTwoZeroVsQZFiveZeroConLo}{\NQzQEightrMTwoZeroVsQZFiveZeroConHi} & \NFOneTwoQOneFourrQZFiveZeroCon{}\pcmarkr{\NFOneTwoQOneFourrMTwoZeroVsQZFiveZeroConLo}{\NFOneTwoQOneFourrMTwoZeroVsQZFiveZeroConHi} & \NFOneTwoROnerQZFiveZeroCon{}\pcmarkr{\NFOneTwoROnerMTwoZeroVsQZFiveZeroConLo}{\NFOneTwoROnerMTwoZeroVsQZFiveZeroConHi} & \NFOneTwoNrQZFiveZeroCon{}\pcmarkr{\NFOneTwoNrMTwoZeroVsQZFiveZeroConLo}{\NFOneTwoNrMTwoZeroVsQZFiveZeroConHi} & \NFOneTwoQEightwQZFiveZeroFOne{}\pcmarkr{\NFOneTwoQEightwMTwoZeroVsQZFiveZeroFOneLo}{\NFOneTwoQEightwMTwoZeroVsQZFiveZeroFOneHi} & \NFOneTwoQOneFourwQZFiveZeroFOne{}\pcmarkr{\NFOneTwoQOneFourwMTwoZeroVsQZFiveZeroFOneLo}{\NFOneTwoQOneFourwMTwoZeroVsQZFiveZeroFOneHi} & \NFOneTwoROnewQZFiveZeroFOne{}\pcmarkr{\NFOneTwoROnewMTwoZeroVsQZFiveZeroFOneLo}{\NFOneTwoROnewMTwoZeroVsQZFiveZeroFOneHi} & \NQzNwQZFiveZeroBox{}\pcmarkr{\NQzNwMTwoZeroVsQZFiveZeroBoxLo}{\NQzNwMTwoZeroVsQZFiveZeroBoxHi} & \NFOneTwoQEighthQZFiveZeroFOne{}\pcmarkr{\NFOneTwoQEighthMTwoZeroVsQZFiveZeroFOneLo}{\NFOneTwoQEighthMTwoZeroVsQZFiveZeroFOneHi} & \NFOneTwoQOneFourhQZFiveZeroFOne{}\pcmarkr{\NFOneTwoQOneFourhMTwoZeroVsQZFiveZeroFOneLo}{\NFOneTwoQOneFourhMTwoZeroVsQZFiveZeroFOneHi} & \NFOneTwoROnehQZFiveZeroFOne{}\pcmarkr{\NFOneTwoROnehMTwoZeroVsQZFiveZeroFOneLo}{\NFOneTwoROnehMTwoZeroVsQZFiveZeroFOneHi} & \NFOneTwoNhQZFiveZeroBox{}\pcmarkr{\NFOneTwoNhMTwoZeroVsQZFiveZeroBoxLo}{\NFOneTwoNhMTwoZeroVsQZFiveZeroBoxHi} \\
\midrule
SurroKV 10\% & \NSkvQEightrSKOneZeroCon{}\pcmark{\NSkvQEightrSKOneZeroVsMTwoZeroConLo}{\NSkvQEightrSKOneZeroVsMTwoZeroConHi} & \NSkvQOneFourrSKOneZeroCon{}\pcmark{\NSkvQOneFourrSKOneZeroVsMTwoZeroConLo}{\NSkvQOneFourrSKOneZeroVsMTwoZeroConHi} & \NSkvROnerSKOneZeroCon{}\pcmark{\NSkvROnerSKOneZeroVsMTwoZeroConLo}{\NSkvROnerSKOneZeroVsMTwoZeroConHi} & \NSkvNrSKOneZeroCon{}\pcmark{\NSkvNrSKOneZeroVsMTwoZeroConLo}{\NSkvNrSKOneZeroVsMTwoZeroConHi} & \NSkvQEightwSKOneZeroFOne{}\pcmark{\NSkvQEightwSKOneZeroVsMTwoZeroFOneLo}{\NSkvQEightwSKOneZeroVsMTwoZeroFOneHi} & \NSkvQOneFourwSKOneZeroFOne{}\pcmark{\NSkvQOneFourwSKOneZeroVsMTwoZeroFOneLo}{\NSkvQOneFourwSKOneZeroVsMTwoZeroFOneHi} & \NSkvROnewSKOneZeroFOne{}\pcmark{\NSkvROnewSKOneZeroVsMTwoZeroFOneLo}{\NSkvROnewSKOneZeroVsMTwoZeroFOneHi} & \NSkvNwSKOneZeroBox{}\pcmark{\NSkvNwSKOneZeroVsMTwoZeroBoxLo}{\NSkvNwSKOneZeroVsMTwoZeroBoxHi} & \NSkvQEighthSKOneZeroFOne{}\pcmark{\NSkvQEighthSKOneZeroVsMTwoZeroFOneLo}{\NSkvQEighthSKOneZeroVsMTwoZeroFOneHi} & \NSkvQOneFourhSKOneZeroFOne{}\pcmark{\NSkvQOneFourhSKOneZeroVsMTwoZeroFOneLo}{\NSkvQOneFourhSKOneZeroVsMTwoZeroFOneHi} & \NSkvROnehSKOneZeroFOne{}\pcmark{\NSkvROnehSKOneZeroVsMTwoZeroFOneLo}{\NSkvROnehSKOneZeroVsMTwoZeroFOneHi} & \NSkvNhSKOneZeroBox{}\pcmark{\NSkvNhSKOneZeroVsMTwoZeroBoxLo}{\NSkvNhSKOneZeroVsMTwoZeroBoxHi} \\
SurroKV 20\% & \NSkvQEightrSKTwoZeroCon{}\pcmark{\NSkvQEightrSKTwoZeroVsMTwoZeroConLo}{\NSkvQEightrSKTwoZeroVsMTwoZeroConHi} & \NSkvQOneFourrSKTwoZeroCon{}\pcmark{\NSkvQOneFourrSKTwoZeroVsMTwoZeroConLo}{\NSkvQOneFourrSKTwoZeroVsMTwoZeroConHi} & \NSkvROnerSKTwoZeroCon{}\pcmark{\NSkvROnerSKTwoZeroVsMTwoZeroConLo}{\NSkvROnerSKTwoZeroVsMTwoZeroConHi} & \NSkvNrSKTwoZeroCon{}\pcmark{\NSkvNrSKTwoZeroVsMTwoZeroConLo}{\NSkvNrSKTwoZeroVsMTwoZeroConHi} & \NSkvQEightwSKTwoZeroFOne{}\pcmark{\NSkvQEightwSKTwoZeroVsMTwoZeroFOneLo}{\NSkvQEightwSKTwoZeroVsMTwoZeroFOneHi} & \NSkvQOneFourwSKTwoZeroFOne{}\pcmark{\NSkvQOneFourwSKTwoZeroVsMTwoZeroFOneLo}{\NSkvQOneFourwSKTwoZeroVsMTwoZeroFOneHi} & \NSkvROnewSKTwoZeroFOne{}\pcmark{\NSkvROnewSKTwoZeroVsMTwoZeroFOneLo}{\NSkvROnewSKTwoZeroVsMTwoZeroFOneHi} & \NSkvNwSKTwoZeroBox{}\pcmark{\NSkvNwSKTwoZeroVsMTwoZeroBoxLo}{\NSkvNwSKTwoZeroVsMTwoZeroBoxHi} & \NSkvQEighthSKTwoZeroFOne{}\pcmark{\NSkvQEighthSKTwoZeroVsMTwoZeroFOneLo}{\NSkvQEighthSKTwoZeroVsMTwoZeroFOneHi} & \NSkvQOneFourhSKTwoZeroFOne{}\pcmark{\NSkvQOneFourhSKTwoZeroVsMTwoZeroFOneLo}{\NSkvQOneFourhSKTwoZeroVsMTwoZeroFOneHi} & \NSkvROnehSKTwoZeroFOne{}\pcmark{\NSkvROnehSKTwoZeroVsMTwoZeroFOneLo}{\NSkvROnehSKTwoZeroVsMTwoZeroFOneHi} & \NSkvNhSKTwoZeroBox{}\pcmark{\NSkvNhSKTwoZeroVsMTwoZeroBoxLo}{\NSkvNhSKTwoZeroVsMTwoZeroBoxHi} \\
No fixed text 20\% & \NFcQEightrPKpTwoZeroCon{}\pcmarkr{\NFcQEightrMTwoZeroVsPKpTwoZeroConLo}{\NFcQEightrMTwoZeroVsPKpTwoZeroConHi} & \NFcQOneFourrPKpTwoZeroCon{}\pcmarkr{\NFcQOneFourrMTwoZeroVsPKpTwoZeroConLo}{\NFcQOneFourrMTwoZeroVsPKpTwoZeroConHi} & \NFcROnerPKpTwoZeroCon{}\pcmarkr{\NFcROnerMTwoZeroVsPKpTwoZeroConLo}{\NFcROnerMTwoZeroVsPKpTwoZeroConHi} & \NGrNrPKpTwoZeroCon{}\pcmarkr{\NGrNrMTwoZeroVsPKpTwoZeroConLo}{\NGrNrMTwoZeroVsPKpTwoZeroConHi} & \NFcQEightwPKpTwoZeroFOne{}\pcmarkr{\NFcQEightwMTwoZeroVsPKpTwoZeroFOneLo}{\NFcQEightwMTwoZeroVsPKpTwoZeroFOneHi} & \NFcQOneFourwPKpTwoZeroFOne{}\pcmarkr{\NFcQOneFourwMTwoZeroVsPKpTwoZeroFOneLo}{\NFcQOneFourwMTwoZeroVsPKpTwoZeroFOneHi} & \NGrROnewPKpTwoZeroFOne{}\pcmarkr{\NGrROnewMTwoZeroVsPKpTwoZeroFOneLo}{\NGrROnewMTwoZeroVsPKpTwoZeroFOneHi} & \NFcNwPKpTwoZeroBox{}\pcmarkr{\NFcNwMTwoZeroVsPKpTwoZeroBoxLo}{\NFcNwMTwoZeroVsPKpTwoZeroBoxHi} & \NFcQEighthPKpTwoZeroFOne{}\pcmarkr{\NFcQEighthMTwoZeroVsPKpTwoZeroFOneLo}{\NFcQEighthMTwoZeroVsPKpTwoZeroFOneHi} & \NFcQOneFourhPKpTwoZeroFOne{}\pcmarkr{\NFcQOneFourhMTwoZeroVsPKpTwoZeroFOneLo}{\NFcQOneFourhMTwoZeroVsPKpTwoZeroFOneHi} & \NGrROnehPKpTwoZeroFOne{}\pcmarkr{\NGrROnehMTwoZeroVsPKpTwoZeroFOneLo}{\NGrROnehMTwoZeroVsPKpTwoZeroFOneHi} & \NFcNhPKpTwoZeroBox{}\pcmarkr{\NFcNhMTwoZeroVsPKpTwoZeroBoxLo}{\NFcNhMTwoZeroVsPKpTwoZeroBoxHi} \\
\midrule
Recipe 20\% BF16 & \NFcQEightrMTwoZeroCon{}\pcmark{\NFcQEightrMTwoZeroVsFConLo}{\NFcQEightrMTwoZeroVsFConHi} & \NFcQOneFourrMTwoZeroCon{}\pcmark{\NFcQOneFourrMTwoZeroVsFConLo}{\NFcQOneFourrMTwoZeroVsFConHi} & \NFcROnerMTwoZeroCon{}\pcmark{\NFcROnerMTwoZeroVsFConLo}{\NFcROnerMTwoZeroVsFConHi} & \NGrNrMTwoZeroCon{}\pcmark{\NGrNrMTwoZeroVsFConLo}{\NGrNrMTwoZeroVsFConHi} & \NFcQEightwMTwoZeroFOne{}\pcmark{\NFcQEightwMTwoZeroVsFFOneLo}{\NFcQEightwMTwoZeroVsFFOneHi} & \NFcQOneFourwMTwoZeroFOne{}\pcmark{\NFcQOneFourwMTwoZeroVsFFOneLo}{\NFcQOneFourwMTwoZeroVsFFOneHi} & \NGrROnewMTwoZeroFOne{}\pcmark{\NGrROnewMTwoZeroVsFFOneLo}{\NGrROnewMTwoZeroVsFFOneHi} & \NFcNwMTwoZeroBox{}\pcmark{\NFcNwMTwoZeroVsFBoxLo}{\NFcNwMTwoZeroVsFBoxHi} & \NFcQEighthMTwoZeroFOne{}\pcmark{\NFcQEighthMTwoZeroVsFFOneLo}{\NFcQEighthMTwoZeroVsFFOneHi} & \NFcQOneFourhMTwoZeroFOne{}\pcmark{\NFcQOneFourhMTwoZeroVsFFOneLo}{\NFcQOneFourhMTwoZeroVsFFOneHi} & \NGrROnehMTwoZeroFOne{}\pcmark{\NGrROnehMTwoZeroVsFFOneLo}{\NGrROnehMTwoZeroVsFFOneHi} & \NFcNhMTwoZeroBox{}\pcmark{\NFcNhMTwoZeroVsFBoxLo}{\NFcNhMTwoZeroVsFBoxHi} \\
Recipe 20\% INT4 & \NFcQEightrMTwoZeroiCon{}\pcmark{\NFcQEightrMTwoZeroiVsFConLo}{\NFcQEightrMTwoZeroiVsFConHi} & \NFcQOneFourrMTwoZeroiCon{}\pcmark{\NFcQOneFourrMTwoZeroiVsFConLo}{\NFcQOneFourrMTwoZeroiVsFConHi} & \NFcROnerMTwoZeroiCon{}\pcmark{\NFcROnerMTwoZeroiVsFConLo}{\NFcROnerMTwoZeroiVsFConHi} & \NGrNrMTwoZeroiCon{}\pcmark{\NGrNrMTwoZeroiVsFConLo}{\NGrNrMTwoZeroiVsFConHi} & \NFcQEightwMTwoZeroiFOne{}\pcmark{\NFcQEightwMTwoZeroiVsFFOneLo}{\NFcQEightwMTwoZeroiVsFFOneHi} & \NFcQOneFourwMTwoZeroiFOne{}\pcmark{\NFcQOneFourwMTwoZeroiVsFFOneLo}{\NFcQOneFourwMTwoZeroiVsFFOneHi} & \NGrROnewMTwoZeroiFOne{}\pcmark{\NGrROnewMTwoZeroiVsFFOneLo}{\NGrROnewMTwoZeroiVsFFOneHi} & \NFcNwMTwoZeroiBox{}\pcmark{\NFcNwMTwoZeroiVsFBoxLo}{\NFcNwMTwoZeroiVsFBoxHi} & \NFcQEighthMTwoZeroiFOne{}\pcmark{\NFcQEighthMTwoZeroiVsFFOneLo}{\NFcQEighthMTwoZeroiVsFFOneHi} & \NFcQOneFourhMTwoZeroiFOne{}\pcmark{\NFcQOneFourhMTwoZeroiVsFFOneLo}{\NFcQOneFourhMTwoZeroiVsFFOneHi} & \NGrROnehMTwoZeroiFOne{}\pcmark{\NGrROnehMTwoZeroiVsFFOneLo}{\NGrROnehMTwoZeroiVsFFOneHi} & \NFcNhMTwoZeroiBox{}\pcmark{\NFcNhMTwoZeroiVsFBoxLo}{\NFcNhMTwoZeroiVsFBoxHi} \\
Recipe 40\% BF16 & \NPFourZeroQEightrMFourZeroCon{}\pcmark{\NPFourZeroQEightrMFourZeroVsFConLo}{\NPFourZeroQEightrMFourZeroVsFConHi} & \NFOneTwoQOneFourrMFourZeroCon{}\pcmark{\NFOneTwoQOneFourrMFourZeroVsFConLo}{\NFOneTwoQOneFourrMFourZeroVsFConHi} & \NPFourZeroROnerMFourZeroCon{}\pcmark{\NPFourZeroROnerMFourZeroVsFConLo}{\NPFourZeroROnerMFourZeroVsFConHi} & \NFOneTwoNrMFourZeroCon{}\pcmark{\NFOneTwoNrMFourZeroVsFConLo}{\NFOneTwoNrMFourZeroVsFConHi} & \NPFourZeroQEightwMFourZeroFOne{}\pcmark{\NPFourZeroQEightwMFourZeroVsFFOneLo}{\NPFourZeroQEightwMFourZeroVsFFOneHi} & \NPFourZeroQOneFourwMFourZeroFOne{}\pcmark{\NPFourZeroQOneFourwMFourZeroVsFFOneLo}{\NPFourZeroQOneFourwMFourZeroVsFFOneHi} & \NFOneTwoROnewMFourZeroFOne{}\pcmark{\NFOneTwoROnewMFourZeroVsFFOneLo}{\NFOneTwoROnewMFourZeroVsFFOneHi} & \NPFourZeroNwMFourZeroBox{}\pcmark{\NPFourZeroNwMFourZeroVsFBoxLo}{\NPFourZeroNwMFourZeroVsFBoxHi} & \NFOneTwoQEighthMFourZeroFOne{}\pcmark{\NFOneTwoQEighthMFourZeroVsFFOneLo}{\NFOneTwoQEighthMFourZeroVsFFOneHi} & \NFOneTwoQOneFourhMFourZeroFOne{}\pcmark{\NFOneTwoQOneFourhMFourZeroVsFFOneLo}{\NFOneTwoQOneFourhMFourZeroVsFFOneHi} & \NFOneTwoROnehMFourZeroFOne{}\pcmark{\NFOneTwoROnehMFourZeroVsFFOneLo}{\NFOneTwoROnehMFourZeroVsFFOneHi} & \NPFourZeroNhMFourZeroBox{}\pcmark{\NPFourZeroNhMFourZeroVsFBoxLo}{\NPFourZeroNhMFourZeroVsFBoxHi} \\
Recipe 40\% INT4 & \NPFourZeroQEightrMFourZeroiCon{}\pcmark{\NPFourZeroQEightrMFourZeroiVsFConLo}{\NPFourZeroQEightrMFourZeroiVsFConHi} & \NFOneTwoQOneFourrMFourZeroiCon{}\pcmark{\NFOneTwoQOneFourrMFourZeroiVsFConLo}{\NFOneTwoQOneFourrMFourZeroiVsFConHi} & \NPFourZeroROnerMFourZeroiCon{}\pcmark{\NPFourZeroROnerMFourZeroiVsFConLo}{\NPFourZeroROnerMFourZeroiVsFConHi} & \NFOneTwoNrMFourZeroiCon{}\pcmark{\NFOneTwoNrMFourZeroiVsFConLo}{\NFOneTwoNrMFourZeroiVsFConHi} & \NPFourZeroQEightwMFourZeroiFOne{}\pcmark{\NPFourZeroQEightwMFourZeroiVsFFOneLo}{\NPFourZeroQEightwMFourZeroiVsFFOneHi} & \NPFourZeroQOneFourwMFourZeroiFOne{}\pcmark{\NPFourZeroQOneFourwMFourZeroiVsFFOneLo}{\NPFourZeroQOneFourwMFourZeroiVsFFOneHi} & \NFOneTwoROnewMFourZeroiFOne{}\pcmark{\NFOneTwoROnewMFourZeroiVsFFOneLo}{\NFOneTwoROnewMFourZeroiVsFFOneHi} & \NFOneTwoNwMFourZeroiBox{}\pcmark{\NFOneTwoNwMFourZeroiVsFBoxLo}{\NFOneTwoNwMFourZeroiVsFBoxHi} & \NFOneTwoQEighthMFourZeroiFOne{}\pcmark{\NFOneTwoQEighthMFourZeroiVsFFOneLo}{\NFOneTwoQEighthMFourZeroiVsFFOneHi} & \NFOneTwoQOneFourhMFourZeroiFOne{}\pcmark{\NFOneTwoQOneFourhMFourZeroiVsFFOneLo}{\NFOneTwoQOneFourhMFourZeroiVsFFOneHi} & \NFOneTwoROnehMFourZeroiFOne{}\pcmark{\NFOneTwoROnehMFourZeroiVsFFOneLo}{\NFOneTwoROnehMFourZeroiVsFFOneHi} & \NPFourZeroNhMFourZeroiBox{}\pcmark{\NPFourZeroNhMFourZeroiVsFBoxLo}{\NPFourZeroNhMFourZeroiVsFBoxHi} \\
\bottomrule
\end{tabular}
\end{table}

\begin{table}[!ht]
\centering
\footnotesize
\caption{Mean generated tokens per sample (reasoning plus answer) per cell, over the same questions and seeds as Table~\ref{tab:cells-score}. $\uparrow$/$\downarrow$: more or fewer tokens, resolved by the paired 95\% interval; for a repair, against the 20\% recipe (BF16); for a recipe setting, against full prefill. Columns as in Table~\ref{tab:cells-score}.}
\label{tab:cells-tokens}
\setlength{\tabcolsep}{2.6pt}
\begin{tabular}{@{}l*{12}{r}@{}}
\toprule
 & \multicolumn{4}{c}{RULER 32K} & \multicolumn{4}{c}{2WikiMQA-E} & \multicolumn{4}{c}{HotpotQA-E} \\
\cmidrule(lr){2-5}\cmidrule(lr){6-9}\cmidrule(lr){10-13}
 & Q-8B & Q-14B & R1-8B & N-8B & Q-8B & Q-14B & R1-8B & N-8B & Q-8B & Q-14B & R1-8B & N-8B \\
\midrule
Full prefill & \NFcQEightrFGen{}\pcblank & \NFcQOneFourrFGen{}\pcblank & \NFcROnerFGen{}\pcblank & \NGrNrFGen{}\pcblank & \NFcQEightwFGen{}\pcblank & \NFcQOneFourwFGen{}\pcblank & \NGrROnewFGen{}\pcblank & \NFcNwFGen{}\pcblank & \NFcQEighthFGen{}\pcblank & \NFcQOneFourhFGen{}\pcblank & \NGrROnehFGen{}\pcblank & \NFcNhFGen{}\pcblank \\
\midrule
CacheBlend 20\% & \NFcQEightrCBGen{}\pcmarkr{\NFcQEightrMTwoZeroVsCBGenLo}{\NFcQEightrMTwoZeroVsCBGenHi} & \NFcQOneFourrCBGen{}\pcmarkr{\NFcQOneFourrMTwoZeroVsCBGenLo}{\NFcQOneFourrMTwoZeroVsCBGenHi} & \NFcROnerCBGen{}\pcmarkr{\NFcROnerMTwoZeroVsCBGenLo}{\NFcROnerMTwoZeroVsCBGenHi} & \NGrNrCBGen{}\pcmarkr{\NGrNrMTwoZeroVsCBGenLo}{\NGrNrMTwoZeroVsCBGenHi} & \NFcQEightwCBGen{}\pcmarkr{\NFcQEightwMTwoZeroVsCBGenLo}{\NFcQEightwMTwoZeroVsCBGenHi} & \NFcQOneFourwCBGen{}\pcmarkr{\NFcQOneFourwMTwoZeroVsCBGenLo}{\NFcQOneFourwMTwoZeroVsCBGenHi} & \NGrROnewCBGen{}\pcmarkr{\NGrROnewMTwoZeroVsCBGenLo}{\NGrROnewMTwoZeroVsCBGenHi} & \NFcNwCBGen{}\pcmarkr{\NFcNwMTwoZeroVsCBGenLo}{\NFcNwMTwoZeroVsCBGenHi} & \NFcQEighthCBGen{}\pcmarkr{\NFcQEighthMTwoZeroVsCBGenLo}{\NFcQEighthMTwoZeroVsCBGenHi} & \NFcQOneFourhCBGen{}\pcmarkr{\NFcQOneFourhMTwoZeroVsCBGenLo}{\NFcQOneFourhMTwoZeroVsCBGenHi} & \NGrROnehCBGen{}\pcmarkr{\NGrROnehMTwoZeroVsCBGenLo}{\NGrROnehMTwoZeroVsCBGenHi} & \NFcNhCBGen{}\pcmarkr{\NFcNhMTwoZeroVsCBGenLo}{\NFcNhMTwoZeroVsCBGenHi} \\
DARIC 20\% & \NDarQEightrDAPTwoZeroGen{}\pcmark{\NDarQEightrDAPTwoZeroVsMTwoZeroGenLo}{\NDarQEightrDAPTwoZeroVsMTwoZeroGenHi} & \NDarQOneFourrDAPTwoZeroGen{}\pcmark{\NDarQOneFourrDAPTwoZeroVsMTwoZeroGenLo}{\NDarQOneFourrDAPTwoZeroVsMTwoZeroGenHi} & \NDarROnerDAPTwoZeroGen{}\pcmark{\NDarROnerDAPTwoZeroVsMTwoZeroGenLo}{\NDarROnerDAPTwoZeroVsMTwoZeroGenHi} & \NDarNrDAPTwoZeroGen{}\pcmark{\NDarNrDAPTwoZeroVsMTwoZeroGenLo}{\NDarNrDAPTwoZeroVsMTwoZeroGenHi} & \NDarQEightwDAPTwoZeroGen{}\pcmark{\NDarQEightwDAPTwoZeroVsMTwoZeroGenLo}{\NDarQEightwDAPTwoZeroVsMTwoZeroGenHi} & \NDarQOneFourwDAPTwoZeroGen{}\pcmark{\NDarQOneFourwDAPTwoZeroVsMTwoZeroGenLo}{\NDarQOneFourwDAPTwoZeroVsMTwoZeroGenHi} & \NDarROnewDAPTwoZeroGen{}\pcmark{\NDarROnewDAPTwoZeroVsMTwoZeroGenLo}{\NDarROnewDAPTwoZeroVsMTwoZeroGenHi} & \NDarNwDAPTwoZeroGen{}\pcmark{\NDarNwDAPTwoZeroVsMTwoZeroGenLo}{\NDarNwDAPTwoZeroVsMTwoZeroGenHi} & \NDarQEighthDAPTwoZeroGen{}\pcmark{\NDarQEighthDAPTwoZeroVsMTwoZeroGenLo}{\NDarQEighthDAPTwoZeroVsMTwoZeroGenHi} & \NDarQOneFourhDAPTwoZeroGen{}\pcmark{\NDarQOneFourhDAPTwoZeroVsMTwoZeroGenLo}{\NDarQOneFourhDAPTwoZeroVsMTwoZeroGenHi} & \NDarROnehDAPTwoZeroGen{}\pcmark{\NDarROnehDAPTwoZeroVsMTwoZeroGenLo}{\NDarROnehDAPTwoZeroVsMTwoZeroGenHi} & \NDarNhDAPTwoZeroGen{}\pcmark{\NDarNhDAPTwoZeroVsMTwoZeroGenLo}{\NDarNhDAPTwoZeroVsMTwoZeroGenHi} \\
ProphetKV 20\% & \NPubQEightrPKGen{}\pcmarkr{\NPubQEightrMTwoZeroVsPKGenLo}{\NPubQEightrMTwoZeroVsPKGenHi} & \NFOneTwoQOneFourrPKGen{}\pcmarkr{\NFOneTwoQOneFourrMTwoZeroVsPKGenLo}{\NFOneTwoQOneFourrMTwoZeroVsPKGenHi} & \NFOneTwoROnerPKGen{}\pcmarkr{\NFOneTwoROnerMTwoZeroVsPKGenLo}{\NFOneTwoROnerMTwoZeroVsPKGenHi} & \NFOneTwoNrPKGen{}\pcmarkr{\NFOneTwoNrMTwoZeroVsPKGenLo}{\NFOneTwoNrMTwoZeroVsPKGenHi} & \NFOneTwoQEightwPKGen{}\pcmarkr{\NFOneTwoQEightwMTwoZeroVsPKGenLo}{\NFOneTwoQEightwMTwoZeroVsPKGenHi} & \NFOneTwoQOneFourwPKGen{}\pcmarkr{\NFOneTwoQOneFourwMTwoZeroVsPKGenLo}{\NFOneTwoQOneFourwMTwoZeroVsPKGenHi} & \NFOneTwoROnewPKGen{}\pcmarkr{\NFOneTwoROnewMTwoZeroVsPKGenLo}{\NFOneTwoROnewMTwoZeroVsPKGenHi} & \NPubNwPKGen{}\pcmarkr{\NPubNwMTwoZeroVsPKGenLo}{\NPubNwMTwoZeroVsPKGenHi} & \NFOneTwoQEighthPKGen{}\pcmarkr{\NFOneTwoQEighthMTwoZeroVsPKGenLo}{\NFOneTwoQEighthMTwoZeroVsPKGenHi} & \NFOneTwoQOneFourhPKGen{}\pcmarkr{\NFOneTwoQOneFourhMTwoZeroVsPKGenLo}{\NFOneTwoQOneFourhMTwoZeroVsPKGenHi} & \NFOneTwoROnehPKGen{}\pcmarkr{\NFOneTwoROnehMTwoZeroVsPKGenLo}{\NFOneTwoROnehMTwoZeroVsPKGenHi} & \NFOneTwoNhPKGen{}\pcmarkr{\NFOneTwoNhMTwoZeroVsPKGenLo}{\NFOneTwoNhMTwoZeroVsPKGenHi} \\
QCFuse 20\% & \NQzQEightrQZTwoZeroGen{}\pcmarkr{\NQzQEightrMTwoZeroVsQZTwoZeroGenLo}{\NQzQEightrMTwoZeroVsQZTwoZeroGenHi} & \NFOneTwoQOneFourrQZTwoZeroGen{}\pcmarkr{\NFOneTwoQOneFourrMTwoZeroVsQZTwoZeroGenLo}{\NFOneTwoQOneFourrMTwoZeroVsQZTwoZeroGenHi} & \NFOneTwoROnerQZTwoZeroGen{}\pcmarkr{\NFOneTwoROnerMTwoZeroVsQZTwoZeroGenLo}{\NFOneTwoROnerMTwoZeroVsQZTwoZeroGenHi} & \NFOneTwoNrQZTwoZeroGen{}\pcmarkr{\NFOneTwoNrMTwoZeroVsQZTwoZeroGenLo}{\NFOneTwoNrMTwoZeroVsQZTwoZeroGenHi} & \NFOneTwoQEightwQZTwoZeroGen{}\pcmarkr{\NFOneTwoQEightwMTwoZeroVsQZTwoZeroGenLo}{\NFOneTwoQEightwMTwoZeroVsQZTwoZeroGenHi} & \NFOneTwoQOneFourwQZTwoZeroGen{}\pcmarkr{\NFOneTwoQOneFourwMTwoZeroVsQZTwoZeroGenLo}{\NFOneTwoQOneFourwMTwoZeroVsQZTwoZeroGenHi} & \NFOneTwoROnewQZTwoZeroGen{}\pcmarkr{\NFOneTwoROnewMTwoZeroVsQZTwoZeroGenLo}{\NFOneTwoROnewMTwoZeroVsQZTwoZeroGenHi} & \NQzNwQZTwoZeroGen{}\pcmarkr{\NQzNwMTwoZeroVsQZTwoZeroGenLo}{\NQzNwMTwoZeroVsQZTwoZeroGenHi} & \NFOneTwoQEighthQZTwoZeroGen{}\pcmarkr{\NFOneTwoQEighthMTwoZeroVsQZTwoZeroGenLo}{\NFOneTwoQEighthMTwoZeroVsQZTwoZeroGenHi} & \NFOneTwoQOneFourhQZTwoZeroGen{}\pcmarkr{\NFOneTwoQOneFourhMTwoZeroVsQZTwoZeroGenLo}{\NFOneTwoQOneFourhMTwoZeroVsQZTwoZeroGenHi} & \NFOneTwoROnehQZTwoZeroGen{}\pcmarkr{\NFOneTwoROnehMTwoZeroVsQZTwoZeroGenLo}{\NFOneTwoROnehMTwoZeroVsQZTwoZeroGenHi} & \NFOneTwoNhQZTwoZeroGen{}\pcmarkr{\NFOneTwoNhMTwoZeroVsQZTwoZeroGenLo}{\NFOneTwoNhMTwoZeroVsQZTwoZeroGenHi} \\
\midrule
EPIC $k{=}2$ & \NEpQEightrLTwoGen{}\pcmarkr{\NEpQEightrMTwoZeroVsLTwoGenLo}{\NEpQEightrMTwoZeroVsLTwoGenHi} & \NFOneTwoQOneFourrLTwoGen{}\pcmarkr{\NFOneTwoQOneFourrMTwoZeroVsLTwoGenLo}{\NFOneTwoQOneFourrMTwoZeroVsLTwoGenHi} & \NFOneTwoROnerLTwoGen{}\pcmarkr{\NFOneTwoROnerMTwoZeroVsLTwoGenLo}{\NFOneTwoROnerMTwoZeroVsLTwoGenHi} & \NFOneTwoNrLTwoGen{}\pcmarkr{\NFOneTwoNrMTwoZeroVsLTwoGenLo}{\NFOneTwoNrMTwoZeroVsLTwoGenHi} & \NFOneTwoQEightwLTwoGen{}\pcmarkr{\NFOneTwoQEightwMTwoZeroVsLTwoGenLo}{\NFOneTwoQEightwMTwoZeroVsLTwoGenHi} & \NFOneTwoQOneFourwLTwoGen{}\pcmarkr{\NFOneTwoQOneFourwMTwoZeroVsLTwoGenLo}{\NFOneTwoQOneFourwMTwoZeroVsLTwoGenHi} & \NFOneTwoROnewLTwoGen{}\pcmarkr{\NFOneTwoROnewMTwoZeroVsLTwoGenLo}{\NFOneTwoROnewMTwoZeroVsLTwoGenHi} & \NEpNwLTwoGen{}\pcmarkr{\NEpNwMTwoZeroVsLTwoGenLo}{\NEpNwMTwoZeroVsLTwoGenHi} & \NFOneTwoQEighthLTwoGen{}\pcmarkr{\NFOneTwoQEighthMTwoZeroVsLTwoGenLo}{\NFOneTwoQEighthMTwoZeroVsLTwoGenHi} & \NFOneTwoQOneFourhLTwoGen{}\pcmarkr{\NFOneTwoQOneFourhMTwoZeroVsLTwoGenLo}{\NFOneTwoQOneFourhMTwoZeroVsLTwoGenHi} & \NFOneTwoROnehLTwoGen{}\pcmarkr{\NFOneTwoROnehMTwoZeroVsLTwoGenLo}{\NFOneTwoROnehMTwoZeroVsLTwoGenHi} & \NFOneTwoNhLTwoGen{}\pcmarkr{\NFOneTwoNhMTwoZeroVsLTwoGenLo}{\NFOneTwoNhMTwoZeroVsLTwoGenHi} \\
EPIC $k{=}32$ & \NEpQEightrLThreeTwoGen{}\pcmarkr{\NEpQEightrMTwoZeroVsLThreeTwoGenLo}{\NEpQEightrMTwoZeroVsLThreeTwoGenHi} & \NFOneTwoQOneFourrLThreeTwoGen{}\pcmarkr{\NFOneTwoQOneFourrMTwoZeroVsLThreeTwoGenLo}{\NFOneTwoQOneFourrMTwoZeroVsLThreeTwoGenHi} & \NFOneTwoROnerLThreeTwoGen{}\pcmarkr{\NFOneTwoROnerMTwoZeroVsLThreeTwoGenLo}{\NFOneTwoROnerMTwoZeroVsLThreeTwoGenHi} & \NFOneTwoNrLThreeTwoGen{}\pcmarkr{\NFOneTwoNrMTwoZeroVsLThreeTwoGenLo}{\NFOneTwoNrMTwoZeroVsLThreeTwoGenHi} & \NFOneTwoQEightwLThreeTwoGen{}\pcmarkr{\NFOneTwoQEightwMTwoZeroVsLThreeTwoGenLo}{\NFOneTwoQEightwMTwoZeroVsLThreeTwoGenHi} & \NFOneTwoQOneFourwLThreeTwoGen{}\pcmarkr{\NFOneTwoQOneFourwMTwoZeroVsLThreeTwoGenLo}{\NFOneTwoQOneFourwMTwoZeroVsLThreeTwoGenHi} & \NFOneTwoROnewLThreeTwoGen{}\pcmarkr{\NFOneTwoROnewMTwoZeroVsLThreeTwoGenLo}{\NFOneTwoROnewMTwoZeroVsLThreeTwoGenHi} & \NEpNwLThreeTwoGen{}\pcmarkr{\NEpNwMTwoZeroVsLThreeTwoGenLo}{\NEpNwMTwoZeroVsLThreeTwoGenHi} & \NFOneTwoQEighthLThreeTwoGen{}\pcmarkr{\NFOneTwoQEighthMTwoZeroVsLThreeTwoGenLo}{\NFOneTwoQEighthMTwoZeroVsLThreeTwoGenHi} & \NFOneTwoQOneFourhLThreeTwoGen{}\pcmarkr{\NFOneTwoQOneFourhMTwoZeroVsLThreeTwoGenLo}{\NFOneTwoQOneFourhMTwoZeroVsLThreeTwoGenHi} & \NFOneTwoROnehLThreeTwoGen{}\pcmarkr{\NFOneTwoROnehMTwoZeroVsLThreeTwoGenLo}{\NFOneTwoROnehMTwoZeroVsLThreeTwoGenHi} & \NFOneTwoNhLThreeTwoGen{}\pcmarkr{\NFOneTwoNhMTwoZeroVsLThreeTwoGenLo}{\NFOneTwoNhMTwoZeroVsLThreeTwoGenHi} \\
DARIC (32/chunk) & \NDarQEightrDAThreeTwoGen{}\pcmark{\NDarQEightrDAThreeTwoVsMTwoZeroGenLo}{\NDarQEightrDAThreeTwoVsMTwoZeroGenHi} & \NDarQOneFourrDAThreeTwoGen{}\pcmark{\NDarQOneFourrDAThreeTwoVsMTwoZeroGenLo}{\NDarQOneFourrDAThreeTwoVsMTwoZeroGenHi} & \NDarROnerDAThreeTwoGen{}\pcmark{\NDarROnerDAThreeTwoVsMTwoZeroGenLo}{\NDarROnerDAThreeTwoVsMTwoZeroGenHi} & \NDarNrDAThreeTwoGen{}\pcmark{\NDarNrDAThreeTwoVsMTwoZeroGenLo}{\NDarNrDAThreeTwoVsMTwoZeroGenHi} & \NDarQEightwDAThreeTwoGen{}\pcmark{\NDarQEightwDAThreeTwoVsMTwoZeroGenLo}{\NDarQEightwDAThreeTwoVsMTwoZeroGenHi} & \NDarQOneFourwDAThreeTwoGen{}\pcmark{\NDarQOneFourwDAThreeTwoVsMTwoZeroGenLo}{\NDarQOneFourwDAThreeTwoVsMTwoZeroGenHi} & \NDarROnewDAThreeTwoGen{}\pcmark{\NDarROnewDAThreeTwoVsMTwoZeroGenLo}{\NDarROnewDAThreeTwoVsMTwoZeroGenHi} & \NDarNwDAThreeTwoGen{}\pcmark{\NDarNwDAThreeTwoVsMTwoZeroGenLo}{\NDarNwDAThreeTwoVsMTwoZeroGenHi} & \NDarQEighthDAThreeTwoGen{}\pcmark{\NDarQEighthDAThreeTwoVsMTwoZeroGenLo}{\NDarQEighthDAThreeTwoVsMTwoZeroGenHi} & \NDarQOneFourhDAThreeTwoGen{}\pcmark{\NDarQOneFourhDAThreeTwoVsMTwoZeroGenLo}{\NDarQOneFourhDAThreeTwoVsMTwoZeroGenHi} & \NDarROnehDAThreeTwoGen{}\pcmark{\NDarROnehDAThreeTwoVsMTwoZeroGenLo}{\NDarROnehDAThreeTwoVsMTwoZeroGenHi} & \NDarNhDAThreeTwoGen{}\pcmark{\NDarNhDAThreeTwoVsMTwoZeroGenLo}{\NDarNhDAThreeTwoVsMTwoZeroGenHi} \\
AgentKVShift & \NPubQEightrAKGen{}\pcmarkr{\NPubQEightrMTwoZeroVsAKGenLo}{\NPubQEightrMTwoZeroVsAKGenHi} & \NFOneTwoQOneFourrAKGen{}\pcmarkr{\NFOneTwoQOneFourrMTwoZeroVsAKGenLo}{\NFOneTwoQOneFourrMTwoZeroVsAKGenHi} & \NFOneTwoROnerAKGen{}\pcmarkr{\NFOneTwoROnerMTwoZeroVsAKGenLo}{\NFOneTwoROnerMTwoZeroVsAKGenHi} & \NFOneTwoNrAKGen{}\pcmarkr{\NFOneTwoNrMTwoZeroVsAKGenLo}{\NFOneTwoNrMTwoZeroVsAKGenHi} & \NFOneTwoQEightwAKGen{}\pcmarkr{\NFOneTwoQEightwMTwoZeroVsAKGenLo}{\NFOneTwoQEightwMTwoZeroVsAKGenHi} & \NFOneTwoQOneFourwAKGen{}\pcmarkr{\NFOneTwoQOneFourwMTwoZeroVsAKGenLo}{\NFOneTwoQOneFourwMTwoZeroVsAKGenHi} & \NFOneTwoROnewAKGen{}\pcmarkr{\NFOneTwoROnewMTwoZeroVsAKGenLo}{\NFOneTwoROnewMTwoZeroVsAKGenHi} & \NPubNwAKGen{}\pcmarkr{\NPubNwMTwoZeroVsAKGenLo}{\NPubNwMTwoZeroVsAKGenHi} & \NFOneTwoQEighthAKGen{}\pcmarkr{\NFOneTwoQEighthMTwoZeroVsAKGenLo}{\NFOneTwoQEighthMTwoZeroVsAKGenHi} & \NFOneTwoQOneFourhAKGen{}\pcmarkr{\NFOneTwoQOneFourhMTwoZeroVsAKGenLo}{\NFOneTwoQOneFourhMTwoZeroVsAKGenHi} & \NFOneTwoROnehAKGen{}\pcmarkr{\NFOneTwoROnehMTwoZeroVsAKGenLo}{\NFOneTwoROnehMTwoZeroVsAKGenHi} & \NFOneTwoNhAKGen{}\pcmarkr{\NFOneTwoNhMTwoZeroVsAKGenLo}{\NFOneTwoNhMTwoZeroVsAKGenHi} \\
FusionRAG 15\% & \NFcQEightrFRGen{}\pcmarkr{\NFcQEightrMTwoZeroVsFRGenLo}{\NFcQEightrMTwoZeroVsFRGenHi} & \NFcQOneFourrFRGen{}\pcmarkr{\NFcQOneFourrMTwoZeroVsFRGenLo}{\NFcQOneFourrMTwoZeroVsFRGenHi} & \NFcROnerFRGen{}\pcmarkr{\NFcROnerMTwoZeroVsFRGenLo}{\NFcROnerMTwoZeroVsFRGenHi} & \NGrNrFRGen{}\pcmarkr{\NGrNrMTwoZeroVsFRGenLo}{\NGrNrMTwoZeroVsFRGenHi} & \NFcQEightwFRGen{}\pcmarkr{\NFcQEightwMTwoZeroVsFRGenLo}{\NFcQEightwMTwoZeroVsFRGenHi} & \NFcQOneFourwFRGen{}\pcmarkr{\NFcQOneFourwMTwoZeroVsFRGenLo}{\NFcQOneFourwMTwoZeroVsFRGenHi} & \NGrROnewFRGen{}\pcmarkr{\NGrROnewMTwoZeroVsFRGenLo}{\NGrROnewMTwoZeroVsFRGenHi} & \NFcNwFRGen{}\pcmarkr{\NFcNwMTwoZeroVsFRGenLo}{\NFcNwMTwoZeroVsFRGenHi} & \NFcQEighthFRGen{}\pcmarkr{\NFcQEighthMTwoZeroVsFRGenLo}{\NFcQEighthMTwoZeroVsFRGenHi} & \NFcQOneFourhFRGen{}\pcmarkr{\NFcQOneFourhMTwoZeroVsFRGenLo}{\NFcQOneFourhMTwoZeroVsFRGenHi} & \NGrROnehFRGen{}\pcmarkr{\NGrROnehMTwoZeroVsFRGenLo}{\NGrROnehMTwoZeroVsFRGenHi} & \NFcNhFRGen{}\pcmarkr{\NFcNhMTwoZeroVsFRGenLo}{\NFcNhMTwoZeroVsFRGenHi} \\
QCFuse 50\% & \NQzQEightrQZFiveZeroGen{}\pcmarkr{\NQzQEightrMTwoZeroVsQZFiveZeroGenLo}{\NQzQEightrMTwoZeroVsQZFiveZeroGenHi} & \NFOneTwoQOneFourrQZFiveZeroGen{}\pcmarkr{\NFOneTwoQOneFourrMTwoZeroVsQZFiveZeroGenLo}{\NFOneTwoQOneFourrMTwoZeroVsQZFiveZeroGenHi} & \NFOneTwoROnerQZFiveZeroGen{}\pcmarkr{\NFOneTwoROnerMTwoZeroVsQZFiveZeroGenLo}{\NFOneTwoROnerMTwoZeroVsQZFiveZeroGenHi} & \NFOneTwoNrQZFiveZeroGen{}\pcmarkr{\NFOneTwoNrMTwoZeroVsQZFiveZeroGenLo}{\NFOneTwoNrMTwoZeroVsQZFiveZeroGenHi} & \NFOneTwoQEightwQZFiveZeroGen{}\pcmarkr{\NFOneTwoQEightwMTwoZeroVsQZFiveZeroGenLo}{\NFOneTwoQEightwMTwoZeroVsQZFiveZeroGenHi} & \NFOneTwoQOneFourwQZFiveZeroGen{}\pcmarkr{\NFOneTwoQOneFourwMTwoZeroVsQZFiveZeroGenLo}{\NFOneTwoQOneFourwMTwoZeroVsQZFiveZeroGenHi} & \NFOneTwoROnewQZFiveZeroGen{}\pcmarkr{\NFOneTwoROnewMTwoZeroVsQZFiveZeroGenLo}{\NFOneTwoROnewMTwoZeroVsQZFiveZeroGenHi} & \NQzNwQZFiveZeroGen{}\pcmarkr{\NQzNwMTwoZeroVsQZFiveZeroGenLo}{\NQzNwMTwoZeroVsQZFiveZeroGenHi} & \NFOneTwoQEighthQZFiveZeroGen{}\pcmarkr{\NFOneTwoQEighthMTwoZeroVsQZFiveZeroGenLo}{\NFOneTwoQEighthMTwoZeroVsQZFiveZeroGenHi} & \NFOneTwoQOneFourhQZFiveZeroGen{}\pcmarkr{\NFOneTwoQOneFourhMTwoZeroVsQZFiveZeroGenLo}{\NFOneTwoQOneFourhMTwoZeroVsQZFiveZeroGenHi} & \NFOneTwoROnehQZFiveZeroGen{}\pcmarkr{\NFOneTwoROnehMTwoZeroVsQZFiveZeroGenLo}{\NFOneTwoROnehMTwoZeroVsQZFiveZeroGenHi} & \NFOneTwoNhQZFiveZeroGen{}\pcmarkr{\NFOneTwoNhMTwoZeroVsQZFiveZeroGenLo}{\NFOneTwoNhMTwoZeroVsQZFiveZeroGenHi} \\
\midrule
SurroKV 10\% & \NSkvQEightrSKOneZeroGen{}\pcmark{\NSkvQEightrSKOneZeroVsMTwoZeroGenLo}{\NSkvQEightrSKOneZeroVsMTwoZeroGenHi} & \NSkvQOneFourrSKOneZeroGen{}\pcmark{\NSkvQOneFourrSKOneZeroVsMTwoZeroGenLo}{\NSkvQOneFourrSKOneZeroVsMTwoZeroGenHi} & \NSkvROnerSKOneZeroGen{}\pcmark{\NSkvROnerSKOneZeroVsMTwoZeroGenLo}{\NSkvROnerSKOneZeroVsMTwoZeroGenHi} & \NSkvNrSKOneZeroGen{}\pcmark{\NSkvNrSKOneZeroVsMTwoZeroGenLo}{\NSkvNrSKOneZeroVsMTwoZeroGenHi} & \NSkvQEightwSKOneZeroGen{}\pcmark{\NSkvQEightwSKOneZeroVsMTwoZeroGenLo}{\NSkvQEightwSKOneZeroVsMTwoZeroGenHi} & \NSkvQOneFourwSKOneZeroGen{}\pcmark{\NSkvQOneFourwSKOneZeroVsMTwoZeroGenLo}{\NSkvQOneFourwSKOneZeroVsMTwoZeroGenHi} & \NSkvROnewSKOneZeroGen{}\pcmark{\NSkvROnewSKOneZeroVsMTwoZeroGenLo}{\NSkvROnewSKOneZeroVsMTwoZeroGenHi} & \NSkvNwSKOneZeroGen{}\pcmark{\NSkvNwSKOneZeroVsMTwoZeroGenLo}{\NSkvNwSKOneZeroVsMTwoZeroGenHi} & \NSkvQEighthSKOneZeroGen{}\pcmark{\NSkvQEighthSKOneZeroVsMTwoZeroGenLo}{\NSkvQEighthSKOneZeroVsMTwoZeroGenHi} & \NSkvQOneFourhSKOneZeroGen{}\pcmark{\NSkvQOneFourhSKOneZeroVsMTwoZeroGenLo}{\NSkvQOneFourhSKOneZeroVsMTwoZeroGenHi} & \NSkvROnehSKOneZeroGen{}\pcmark{\NSkvROnehSKOneZeroVsMTwoZeroGenLo}{\NSkvROnehSKOneZeroVsMTwoZeroGenHi} & \NSkvNhSKOneZeroGen{}\pcmark{\NSkvNhSKOneZeroVsMTwoZeroGenLo}{\NSkvNhSKOneZeroVsMTwoZeroGenHi} \\
SurroKV 20\% & \NSkvQEightrSKTwoZeroGen{}\pcmark{\NSkvQEightrSKTwoZeroVsMTwoZeroGenLo}{\NSkvQEightrSKTwoZeroVsMTwoZeroGenHi} & \NSkvQOneFourrSKTwoZeroGen{}\pcmark{\NSkvQOneFourrSKTwoZeroVsMTwoZeroGenLo}{\NSkvQOneFourrSKTwoZeroVsMTwoZeroGenHi} & \NSkvROnerSKTwoZeroGen{}\pcmark{\NSkvROnerSKTwoZeroVsMTwoZeroGenLo}{\NSkvROnerSKTwoZeroVsMTwoZeroGenHi} & \NSkvNrSKTwoZeroGen{}\pcmark{\NSkvNrSKTwoZeroVsMTwoZeroGenLo}{\NSkvNrSKTwoZeroVsMTwoZeroGenHi} & \NSkvQEightwSKTwoZeroGen{}\pcmark{\NSkvQEightwSKTwoZeroVsMTwoZeroGenLo}{\NSkvQEightwSKTwoZeroVsMTwoZeroGenHi} & \NSkvQOneFourwSKTwoZeroGen{}\pcmark{\NSkvQOneFourwSKTwoZeroVsMTwoZeroGenLo}{\NSkvQOneFourwSKTwoZeroVsMTwoZeroGenHi} & \NSkvROnewSKTwoZeroGen{}\pcmark{\NSkvROnewSKTwoZeroVsMTwoZeroGenLo}{\NSkvROnewSKTwoZeroVsMTwoZeroGenHi} & \NSkvNwSKTwoZeroGen{}\pcmark{\NSkvNwSKTwoZeroVsMTwoZeroGenLo}{\NSkvNwSKTwoZeroVsMTwoZeroGenHi} & \NSkvQEighthSKTwoZeroGen{}\pcmark{\NSkvQEighthSKTwoZeroVsMTwoZeroGenLo}{\NSkvQEighthSKTwoZeroVsMTwoZeroGenHi} & \NSkvQOneFourhSKTwoZeroGen{}\pcmark{\NSkvQOneFourhSKTwoZeroVsMTwoZeroGenLo}{\NSkvQOneFourhSKTwoZeroVsMTwoZeroGenHi} & \NSkvROnehSKTwoZeroGen{}\pcmark{\NSkvROnehSKTwoZeroVsMTwoZeroGenLo}{\NSkvROnehSKTwoZeroVsMTwoZeroGenHi} & \NSkvNhSKTwoZeroGen{}\pcmark{\NSkvNhSKTwoZeroVsMTwoZeroGenLo}{\NSkvNhSKTwoZeroVsMTwoZeroGenHi} \\
No fixed text 20\% & \NFcQEightrPKpTwoZeroGen{}\pcmarkr{\NFcQEightrMTwoZeroVsPKpTwoZeroGenLo}{\NFcQEightrMTwoZeroVsPKpTwoZeroGenHi} & \NFcQOneFourrPKpTwoZeroGen{}\pcmarkr{\NFcQOneFourrMTwoZeroVsPKpTwoZeroGenLo}{\NFcQOneFourrMTwoZeroVsPKpTwoZeroGenHi} & \NFcROnerPKpTwoZeroGen{}\pcmarkr{\NFcROnerMTwoZeroVsPKpTwoZeroGenLo}{\NFcROnerMTwoZeroVsPKpTwoZeroGenHi} & \NGrNrPKpTwoZeroGen{}\pcmarkr{\NGrNrMTwoZeroVsPKpTwoZeroGenLo}{\NGrNrMTwoZeroVsPKpTwoZeroGenHi} & \NFcQEightwPKpTwoZeroGen{}\pcmarkr{\NFcQEightwMTwoZeroVsPKpTwoZeroGenLo}{\NFcQEightwMTwoZeroVsPKpTwoZeroGenHi} & \NFcQOneFourwPKpTwoZeroGen{}\pcmarkr{\NFcQOneFourwMTwoZeroVsPKpTwoZeroGenLo}{\NFcQOneFourwMTwoZeroVsPKpTwoZeroGenHi} & \NGrROnewPKpTwoZeroGen{}\pcmarkr{\NGrROnewMTwoZeroVsPKpTwoZeroGenLo}{\NGrROnewMTwoZeroVsPKpTwoZeroGenHi} & \NFcNwPKpTwoZeroGen{}\pcmarkr{\NFcNwMTwoZeroVsPKpTwoZeroGenLo}{\NFcNwMTwoZeroVsPKpTwoZeroGenHi} & \NFcQEighthPKpTwoZeroGen{}\pcmarkr{\NFcQEighthMTwoZeroVsPKpTwoZeroGenLo}{\NFcQEighthMTwoZeroVsPKpTwoZeroGenHi} & \NFcQOneFourhPKpTwoZeroGen{}\pcmarkr{\NFcQOneFourhMTwoZeroVsPKpTwoZeroGenLo}{\NFcQOneFourhMTwoZeroVsPKpTwoZeroGenHi} & \NGrROnehPKpTwoZeroGen{}\pcmarkr{\NGrROnehMTwoZeroVsPKpTwoZeroGenLo}{\NGrROnehMTwoZeroVsPKpTwoZeroGenHi} & \NFcNhPKpTwoZeroGen{}\pcmarkr{\NFcNhMTwoZeroVsPKpTwoZeroGenLo}{\NFcNhMTwoZeroVsPKpTwoZeroGenHi} \\
\midrule
Recipe 20\% BF16 & \NFcQEightrMTwoZeroGen{}\pcmark{\NFcQEightrMTwoZeroVsFGenLo}{\NFcQEightrMTwoZeroVsFGenHi} & \NFcQOneFourrMTwoZeroGen{}\pcmark{\NFcQOneFourrMTwoZeroVsFGenLo}{\NFcQOneFourrMTwoZeroVsFGenHi} & \NFcROnerMTwoZeroGen{}\pcmark{\NFcROnerMTwoZeroVsFGenLo}{\NFcROnerMTwoZeroVsFGenHi} & \NGrNrMTwoZeroGen{}\pcmark{\NGrNrMTwoZeroVsFGenLo}{\NGrNrMTwoZeroVsFGenHi} & \NFcQEightwMTwoZeroGen{}\pcmark{\NFcQEightwMTwoZeroVsFGenLo}{\NFcQEightwMTwoZeroVsFGenHi} & \NFcQOneFourwMTwoZeroGen{}\pcmark{\NFcQOneFourwMTwoZeroVsFGenLo}{\NFcQOneFourwMTwoZeroVsFGenHi} & \NGrROnewMTwoZeroGen{}\pcmark{\NGrROnewMTwoZeroVsFGenLo}{\NGrROnewMTwoZeroVsFGenHi} & \NFcNwMTwoZeroGen{}\pcmark{\NFcNwMTwoZeroVsFGenLo}{\NFcNwMTwoZeroVsFGenHi} & \NFcQEighthMTwoZeroGen{}\pcmark{\NFcQEighthMTwoZeroVsFGenLo}{\NFcQEighthMTwoZeroVsFGenHi} & \NFcQOneFourhMTwoZeroGen{}\pcmark{\NFcQOneFourhMTwoZeroVsFGenLo}{\NFcQOneFourhMTwoZeroVsFGenHi} & \NGrROnehMTwoZeroGen{}\pcmark{\NGrROnehMTwoZeroVsFGenLo}{\NGrROnehMTwoZeroVsFGenHi} & \NFcNhMTwoZeroGen{}\pcmark{\NFcNhMTwoZeroVsFGenLo}{\NFcNhMTwoZeroVsFGenHi} \\
Recipe 20\% INT4 & \NFcQEightrMTwoZeroiGen{}\pcmark{\NFcQEightrMTwoZeroiVsFGenLo}{\NFcQEightrMTwoZeroiVsFGenHi} & \NFcQOneFourrMTwoZeroiGen{}\pcmark{\NFcQOneFourrMTwoZeroiVsFGenLo}{\NFcQOneFourrMTwoZeroiVsFGenHi} & \NFcROnerMTwoZeroiGen{}\pcmark{\NFcROnerMTwoZeroiVsFGenLo}{\NFcROnerMTwoZeroiVsFGenHi} & \NGrNrMTwoZeroiGen{}\pcmark{\NGrNrMTwoZeroiVsFGenLo}{\NGrNrMTwoZeroiVsFGenHi} & \NFcQEightwMTwoZeroiGen{}\pcmark{\NFcQEightwMTwoZeroiVsFGenLo}{\NFcQEightwMTwoZeroiVsFGenHi} & \NFcQOneFourwMTwoZeroiGen{}\pcmark{\NFcQOneFourwMTwoZeroiVsFGenLo}{\NFcQOneFourwMTwoZeroiVsFGenHi} & \NGrROnewMTwoZeroiGen{}\pcmark{\NGrROnewMTwoZeroiVsFGenLo}{\NGrROnewMTwoZeroiVsFGenHi} & \NFcNwMTwoZeroiGen{}\pcmark{\NFcNwMTwoZeroiVsFGenLo}{\NFcNwMTwoZeroiVsFGenHi} & \NFcQEighthMTwoZeroiGen{}\pcmark{\NFcQEighthMTwoZeroiVsFGenLo}{\NFcQEighthMTwoZeroiVsFGenHi} & \NFcQOneFourhMTwoZeroiGen{}\pcmark{\NFcQOneFourhMTwoZeroiVsFGenLo}{\NFcQOneFourhMTwoZeroiVsFGenHi} & \NGrROnehMTwoZeroiGen{}\pcmark{\NGrROnehMTwoZeroiVsFGenLo}{\NGrROnehMTwoZeroiVsFGenHi} & \NFcNhMTwoZeroiGen{}\pcmark{\NFcNhMTwoZeroiVsFGenLo}{\NFcNhMTwoZeroiVsFGenHi} \\
Recipe 40\% BF16 & \NPFourZeroQEightrMFourZeroGen{}\pcmark{\NPFourZeroQEightrMFourZeroVsFGenLo}{\NPFourZeroQEightrMFourZeroVsFGenHi} & \NFOneTwoQOneFourrMFourZeroGen{}\pcmark{\NFOneTwoQOneFourrMFourZeroVsFGenLo}{\NFOneTwoQOneFourrMFourZeroVsFGenHi} & \NPFourZeroROnerMFourZeroGen{}\pcmark{\NPFourZeroROnerMFourZeroVsFGenLo}{\NPFourZeroROnerMFourZeroVsFGenHi} & \NFOneTwoNrMFourZeroGen{}\pcmark{\NFOneTwoNrMFourZeroVsFGenLo}{\NFOneTwoNrMFourZeroVsFGenHi} & \NPFourZeroQEightwMFourZeroGen{}\pcmark{\NPFourZeroQEightwMFourZeroVsFGenLo}{\NPFourZeroQEightwMFourZeroVsFGenHi} & \NPFourZeroQOneFourwMFourZeroGen{}\pcmark{\NPFourZeroQOneFourwMFourZeroVsFGenLo}{\NPFourZeroQOneFourwMFourZeroVsFGenHi} & \NFOneTwoROnewMFourZeroGen{}\pcmark{\NFOneTwoROnewMFourZeroVsFGenLo}{\NFOneTwoROnewMFourZeroVsFGenHi} & \NPFourZeroNwMFourZeroGen{}\pcmark{\NPFourZeroNwMFourZeroVsFGenLo}{\NPFourZeroNwMFourZeroVsFGenHi} & \NFOneTwoQEighthMFourZeroGen{}\pcmark{\NFOneTwoQEighthMFourZeroVsFGenLo}{\NFOneTwoQEighthMFourZeroVsFGenHi} & \NFOneTwoQOneFourhMFourZeroGen{}\pcmark{\NFOneTwoQOneFourhMFourZeroVsFGenLo}{\NFOneTwoQOneFourhMFourZeroVsFGenHi} & \NFOneTwoROnehMFourZeroGen{}\pcmark{\NFOneTwoROnehMFourZeroVsFGenLo}{\NFOneTwoROnehMFourZeroVsFGenHi} & \NPFourZeroNhMFourZeroGen{}\pcmark{\NPFourZeroNhMFourZeroVsFGenLo}{\NPFourZeroNhMFourZeroVsFGenHi} \\
Recipe 40\% INT4 & \NPFourZeroQEightrMFourZeroiGen{}\pcmark{\NPFourZeroQEightrMFourZeroiVsFGenLo}{\NPFourZeroQEightrMFourZeroiVsFGenHi} & \NFOneTwoQOneFourrMFourZeroiGen{}\pcmark{\NFOneTwoQOneFourrMFourZeroiVsFGenLo}{\NFOneTwoQOneFourrMFourZeroiVsFGenHi} & \NPFourZeroROnerMFourZeroiGen{}\pcmark{\NPFourZeroROnerMFourZeroiVsFGenLo}{\NPFourZeroROnerMFourZeroiVsFGenHi} & \NFOneTwoNrMFourZeroiGen{}\pcmark{\NFOneTwoNrMFourZeroiVsFGenLo}{\NFOneTwoNrMFourZeroiVsFGenHi} & \NPFourZeroQEightwMFourZeroiGen{}\pcmark{\NPFourZeroQEightwMFourZeroiVsFGenLo}{\NPFourZeroQEightwMFourZeroiVsFGenHi} & \NPFourZeroQOneFourwMFourZeroiGen{}\pcmark{\NPFourZeroQOneFourwMFourZeroiVsFGenLo}{\NPFourZeroQOneFourwMFourZeroiVsFGenHi} & \NFOneTwoROnewMFourZeroiGen{}\pcmark{\NFOneTwoROnewMFourZeroiVsFGenLo}{\NFOneTwoROnewMFourZeroiVsFGenHi} & \NFOneTwoNwMFourZeroiGen{}\pcmark{\NFOneTwoNwMFourZeroiVsFGenLo}{\NFOneTwoNwMFourZeroiVsFGenHi} & \NFOneTwoQEighthMFourZeroiGen{}\pcmark{\NFOneTwoQEighthMFourZeroiVsFGenLo}{\NFOneTwoQEighthMFourZeroiVsFGenHi} & \NFOneTwoQOneFourhMFourZeroiGen{}\pcmark{\NFOneTwoQOneFourhMFourZeroiVsFGenLo}{\NFOneTwoQOneFourhMFourZeroiVsFGenHi} & \NFOneTwoROnehMFourZeroiGen{}\pcmark{\NFOneTwoROnehMFourZeroiVsFGenLo}{\NFOneTwoROnehMFourZeroiVsFGenHi} & \NPFourZeroNhMFourZeroiGen{}\pcmark{\NPFourZeroNhMFourZeroiVsFGenLo}{\NPFourZeroNhMFourZeroiVsFGenHi} \\
\bottomrule
\end{tabular}
\end{table}

\clearpage

%% file: sections/A-appendix.tex
\section{Results Beyond the 12 Cells}
\label{app:breadth}

\begin{figure}[h]
\centering
\includegraphics[width=\textwidth]{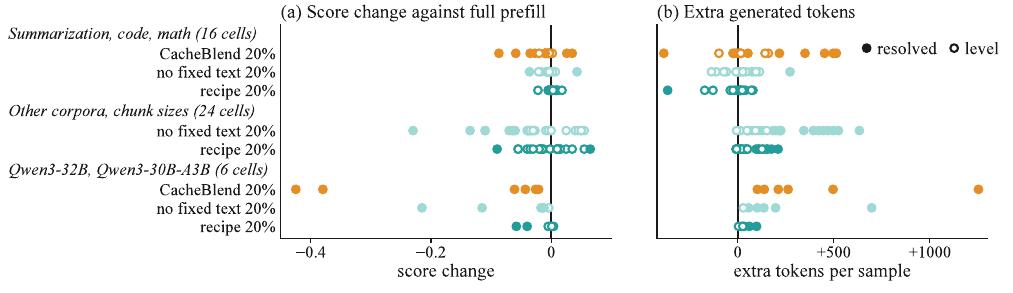}
\caption{Score change and extra generated tokens against full prefill beyond the 12 cells: summarization, code and math; other corpora and larger chunks; Qwen3-32B and Qwen3-30B-A3B. One mark per cell, filled where resolved.}
\label{fig:breadth}
\end{figure}

Section~\ref{sec:breadth} reports the summarization, code and math cells and the larger models; on MATH-500, each problem's chunks are its 16 most similar solved MATH training problems. %
On QASPER papers \citep{dasigi-dataset-2021}, on MultiHop-RAG news \citep{tang-multihoprag-2024}, and on RULER 32K's passages ungrouped and grouped into chunks of up to 512, 1,024 or 2,048 tokens, the recipe's online steps without the fixed text (Section~\ref{sec:ablations}) cost against full prefill in \NPOnethreeExtensionsCost{} of these \NPOnethreeExtensionsCells{} cells, and the recipe gains over them in \NPOnethreeExtensionsGainWhereCost{} of the cells where they cost and falls behind them in none. %
Among resolved differences, a lower score or more generated tokens counts as a cost, and a higher score or fewer tokens counts as a gain.